\documentclass[manuscript, screen, nonacm]{acmart}
\usepackage{subcaption}
\usepackage{tabularx}
\usepackage{overpic}
\usepackage{tikz}
\usetikzlibrary{positioning}
\usepackage{natbib}
\begin{document}

\title{Chronocooked: A Benchmark Environment for Interval
Timing in Reinforcement Learning Agents}


\author{Amrapali Pednekar}
\authornote{Corresponding Author.}
\orcid{0009-0005-6194-3955}
\email{Amrapali.Pednekar@UGent.be}
\affiliation{%
  \institution{Ghent University - imec}
  \city{Ghent}
  \country{Belgium}
}

\author{Álvaro Garrido-Pérez}
\orcid{0009-0003-5481-8166}
\email{Alvaro.GarridoPerez@UGent.be}
\affiliation{%
  \institution{Ghent University - imec}
  \city{Ghent}
  \country{Belgium}
} 

\author{Yara Khaluf}
\orcid{0000-0002-5590-9321}
\email{y.khaluf@vu.nl}
\affiliation{%
  \institution{Vrije Universiteit Amsterdam}
  \city{Amsterdam}
  \country{The Netherlands}
}

\author{Pieter Simoens}
\orcid{0000-0002-9569-9373}
\email{Pieter.Simoens@UGent.be}
\affiliation{%
  \institution{Ghent University - imec}
  \city{Ghent}
  \country{Belgium}
}



\begin{abstract}
    {\bf Background:} 
    Interval timing is extensively studied as an important aspect of human behaviour. As artificial agents are increasingly designed to function alongside humans, their interval timing abilities also needs to be studied. However, research in this area remains limited and scattered. 

    {\bf Objectives:}
    This paper presents Chronocooked, a reinforcement learning (RL) benchmark environment that enables a systematic study of interval timing abilities in RL agents.
    
    {\bf Methods:}
    Inspired by Overcooked, the suite comprises cooking scenarios involving interval timing tasks drawn from the psychology literature. The tasks and reward functions are designed such that temporal information is unobserved but critical for optimal performance. The environment is intentionally kept simple to enable controlled experiments and support biologically plausible models. Each task is accompanied by evaluation metrics to study different aspects of interval timing in RL agents, namely, task performance, human-like timing and scalability. We report baselines using a non-recurrent architecture (CNN), a recurrent architecture (LSTM), and a biologically inspired recurrent architecture (CTRNN).

    {\bf Results:}
    Recurrent RL agents with sufficient memory successfully performed the timing tasks, exhibited temporal control in the Fixed Interval (FI) task, and qualitatively reproduced human-like psychometric curves in the Temporal Bisection task. However, no quantitative evidence of human-like timing behaviour was found. In contrast, the non-recurrent CNN model showed limited ability to perform the basic timing tasks. Overall, the two recurrent architectures showed comparable performance, but the biologically plausible CTRNN architecture was found to be more scalable to new timing requirements than the more powerful LSTM architecture.


    {\bf Conclusions:}
    Although RL agents can successfully perform time-dependent tasks, they do not necessarily process and perceive time in the same way as humans. Understanding these differences is important for anticipating their impact on human–robot interactions (HRI).

\end{abstract}



\maketitle

\noindent\textbf{Code Repository:} \url{https://github.com/decide-ugent/chronocooked} 

\section{Introduction}


Interval timing forms a core component of human cognition that shapes how humans perceive, coordinate, and make decisions in everyday life. It is defined as the ability to perceive, estimate and track durations in the seconds-to-minutes range \citep{zimbardo2014putting, ariely2001timely, thones2019standard, block2014time}. It is an  underlying element for everyday tasks such as pacing a speech, judging when to cross a street, or knowing when its time to check the soup on the stove \cite{ng2021ecological}. Interestingly, humans can perform this kind of timing without external clocks or explicit environmental cues, relying instead on an internal sense of elapsed time. For instance, even in a dark room, we have a rough sense of how much time has passed since we entered.


Given its ubiquity in human behaviour \cite{marshall2015everywhere}, interval timing should be a fundamental characteristic of AI agents that are expected to function alongside humans on shared tasks. Many existing AI systems already perform well on tasks that implicitly require temporal processing \cite{vinyals2019grandmaster, jaderberg2019human, badia2020agent57}. However, these systems are typically evaluated only on final task performance. It remains unclear whether they develop an internal sense of time during training or merely rely on task-specific heuristics and environmental cues. For example, in human-robot interaction (HRI), interval timing tasks like synchronization and coordination with human partners are often achieved by relying on observable cues from the environment or the human partner \cite{ajoudani2018progress,hoffman2019evaluating,nikolaidis2015efficient}, rather than on the robot's own temporal processing. As a result, whether artificial agents develop a genuine sense of time when solving such tasks, or merely exploit available cues, remains largely unexamined. 

Temporal processing, in general, has received limited attention as a distinct dimension in artificial agents, unlike visual, auditory, and other sensory inputs. While temporal processing is a multifaceted dimension, this study focuses on a specific aspect of temporal processing, namely, interval timing in artificial agents. A field that would directly benefit from studying interval timing capabilities in artificial agents is HRI.
This is because many HRI tasks like synchronising actions, anticipating a partner's actions, and selecting appropriate times to act or communicate, impose temporal demands that can require interval timing. The inherent role of interval timing in such tasks is evidenced by the human–human versions of these tasks \cite{wolf2018joint,jung2011sharing,miyata2021vocal}. Hence, studying interval timing in artificial agents could provide a complementary perspective on improving performance and efficiency in HRI.

Interval timing has been extensively studied in psychology using interval timing tasks that deliberately isolate temporal processing from other strategies such as counting or reliance on visual, auditory or haptic cues. These paradigms allow researchers to investigate how humans encode and perceive time, and make decisions based on it \cite{matell2000neuropsychological,merchant2014introduction, buhusi2005makes, hass2014neurocomputational,jazayeri2010temporal}. 

Building on this approach, we introduce Chronocooked, a reinforcement learning (RL) benchmark environment for studying interval timing in RL agents. It uses cooking-themed scenarios that operationalize well-established psychology paradigms within an RL environment. A RL framework is chosen because it treats interval timing as a closed-loop problem, in which agents interact with the environment and learn policies through their actions and rewards \cite{sutton1998reinforcement}. It thus enables the study of temporal characteristics that emerge in the RL agents as a result of solving the interval timing task. Each task is accompanied by a set of evaluation metrics and baseline results across three model architectures. We summarize our contributions as follows:


\begin{itemize}
    \item \textbf{An RL environment for interval timing:} Chronocooked introduces tasks inspired from psychology that isolate temporal processing from environmental cues. The reward functions and environment states do not carry any explicit information about time. Hence,  time is an unobserved variable that must be inferred for optimal task performance. The environment is kept simple to provide a fair testing ground for all types of models, including biologically inspired models that may capture timing well but have limited computational resources. 

    \item \textbf{Evaluation metrics:} The study translates established evaluation metrics from psychology and introduces a few task-specific metrics that together serve three goals. (i) Task performance: evaluates whether RL agents correctly infer and use temporal information to solve the task (ii) Human-like timing: evaluates the extent to which RL agents' temporal processing and timing behaviour replicates that of humans. A closer replication could mean better prediction of human responses to timing events and a smoother HRI experience. (iii) Scalability: evaluates how well RL agents adapt to new timing requirements. These metrics aim to isolate flexible timing strategies from the effects of memory capacity or additional training. A flexible timing strategy suggests that an agent has learned the underlying timing task and can adapt to new timing requirements with less training. 

    \item \textbf{Curriculum-learning (CL) extensions:} The benchmark also introduces extended versions of the psychology-inspired tasks. These tasks can be trained using curriculum learning \cite{bengio2009curriculum}, and thus enable introducing temporal complexities incrementally. They provide a controlled way to study how agents adapt as new time requirements are progressively introduced in an existing task.

    \item \textbf{Benchmarking of representative models:} Three types of models are evaluated using Chronocooked. A non-recurrent model (Convolutional neural network (CNN)) \cite{lecun1989handwritten}, a recurrent model (Long short-term memory (LSTM)) \cite{hochreiter1997long}, and a biologically plausible recurrent model (Continuos time recurrent neural network (CTRNN)) \cite{yu2017continuous,beer1995dynamics,funahashi1993approximation}, establishing a comparative foundation for future work on temporal processing in RL agents.

\end{itemize}


\section{Prior Work}

Research on the timing abilities of AI models has seen some advances in recent years. Several studies have investigated temporal behaviour of deep learning models in different settings \cite{deverett2019interval,roseboom2019activity,labash2023emergence,safaie2020turning}. These studies aim to show some level of behavioural similarity with biological systems either through a qualitative replication of human or animal timing data, or through some degree of conformity to well-established timing biases. Across these studies, timing ability is supported either through reward functions that explicitly encode temporal structure, or through environmental cues correlated with elapsed time. In contrast, our benchmark provides no explicit temporal information in either the reward function or environment states. This allows us to test whether the RL agents can still infer time and achieve optimal task performance. Furthermore, Chronocooked provides evaluation metrics to quantify similarities between human and artificial agents’ temporal behaviours. To the best of our knowledge, this is the first systematic framework for comparing established human timing biases with those of artificial agents.



A second line of work focuses on neural similarities. Research in this area examines whether biologically plausible neural activities such as time cells, ramping activity, and neural oscillations, emerge in artificial agents as time-keeping substrates \cite{lin2023temporal, pednekar2025emergent, goudar2018encoding, contextdependent_mante_2013, maniadakis2012experiencing, maniadakis2015artificial}. These studies largely address how agents solve the timing tasks. However, the findings vary considerably depending on task design and reward structure, without a common benchmark to compare across studies. Our work complements this line of research by providing a standardized environment and reward structure in which such neural mechanisms could be examined across a consistent set of tasks. 

A third line of work explicitly engineers timing mechanisms into AI models. These studies aim to improve task performance by incorporating concepts from neuroscience and psychology into AI models \cite{kabir2025deep, howlong_maniadakis_2016, teaching_loureno_2020, predictive_fountas_2022, timeorder_komosiski_2015, latent_cimolino_2021, cominelli2019influence, gers2002learning, wang2022neural}. For example, \citet{kabir2025deep} introduced scale-invariant temporal memory, showing improved task performance and learning efficiency across different temporal scales of the environment. \citet{teaching_loureno_2020, predictive_fountas_2022} used theoretical knowledge of the relationship between environmental stimulus statistics and temporal processing to induce timing abilities in artificial agents. \citet{maniadakis2012experiencing} introduced a neuro-evolutionary framework to add a sources of temporal processing in artificial agents to enable them to simultaneously measure event durations and determine when events occurred. \citet{cominelli2019influence} introduced a cognitively inspired module for social robotics that incorporated the influence of emotion on time perception. These studies demonstrate how theoretical knowledge from psychology and neuroscience can inform time-aware agent design. Our benchmark offers a common framework in which such engineered models could be tested for a wider range of timing abilities such as human-like timing, beyond the specific task each was originally designed for.



A related and emerging research direction involves studying the impact of using temporal information as an additional input to artificial agents \cite{jia2025time,watson2026let}. These studies show that providing information such as time left until the end of episode or wait time for task completion, can improve model performance, robustness, and efficiency. These studies complement our work by highlighting the value of temporal information as an additional dimension. ChronoCooked motivates the study of timing abilities of RL agents beyond explicitly providing task-specific temporal information.



Timing models have also been independently developed in neuroscience and psychology to explain temporal behaviour in humans and animals \cite{karmarkar2007timing,hass2014neurocomputational,hass2016time,basgol2021time}. These models prioritize biological plausibility over task performance, but could serve as useful components in AI systems seeking to incorporate a sense of time. Additionally, RL has been linked to interval timing due to the temporal difference learning algorithm that has been successfully used to implement models of the basal ganglia \cite{gershman2014time,petter2018integrating,daw2006representation}. Exploring this connection further may require integrating biologically plausible models into RL settings. The computational complexity of existing RL environments, however, can be an obstacle for such models. 

Across all these research threads, a shared limitation is the absence of a simple, standardized benchmark. The closest prior work to our benchmark is PsychLab \cite{leibo2018psychlab}, an RL environment for simulating psychology experiments, which includes some timing tasks \cite{deverett2019interval}. We take this work one step further by introducing a benchmark environment solely focused on interval timing. In contrast to PsychLab, Chronocooked includes additional timing tasks, evaluation metrics to quantify different timing abilities and baseline results on three types of model architectures. 


\section{Chronocooked Environment}

\begin{figure}[ht]
    \centering
    \begin{subfigure}{0.49\textwidth}
        \centering
        \includegraphics[width=0.55\linewidth]{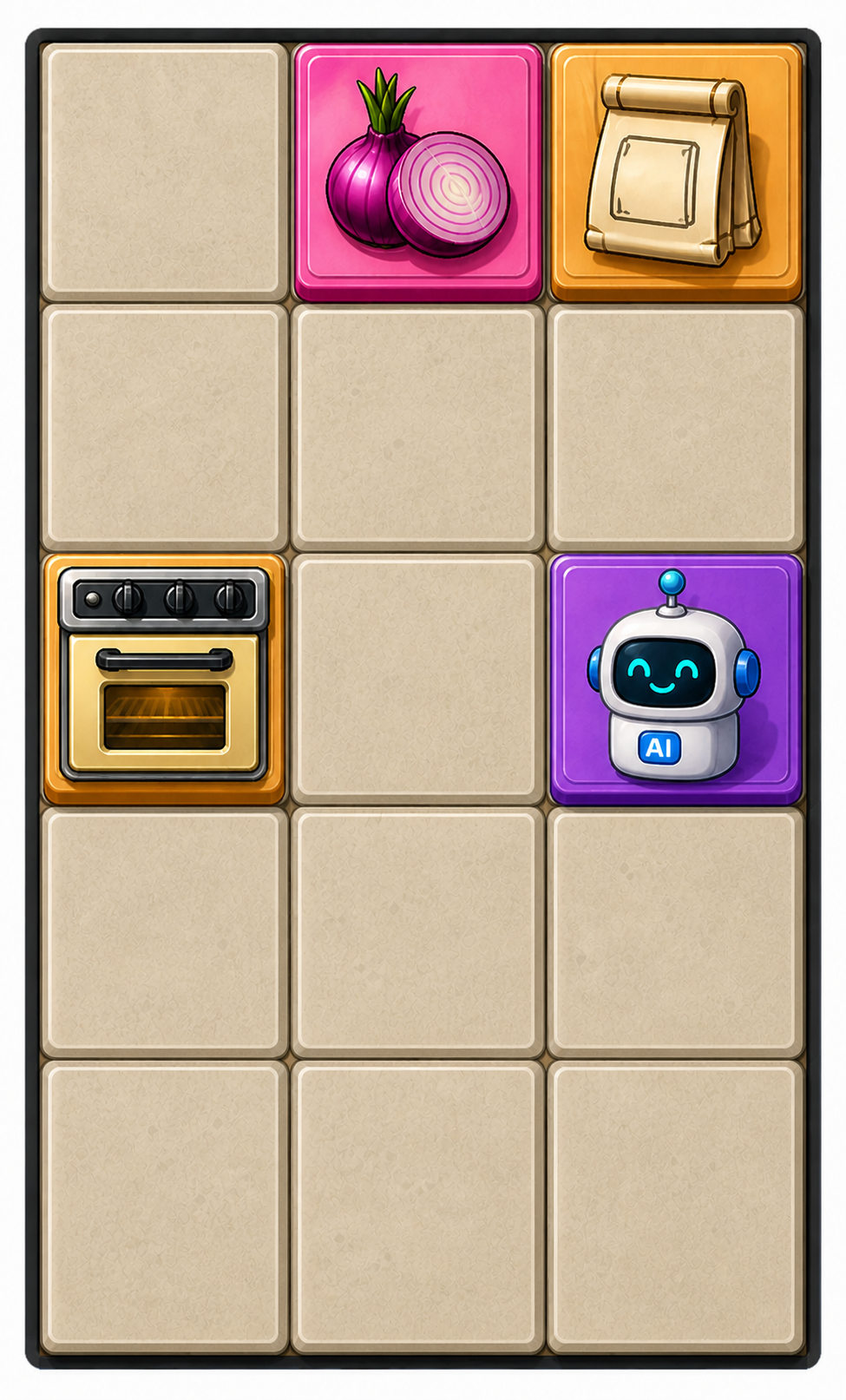}
        \caption{Fixed Interval (FI) timing 
                               }
        \label{fig:task_fi}
    \end{subfigure}
    \begin{subfigure}{0.49\textwidth}
        \centering
        \includegraphics[width=0.6\linewidth]{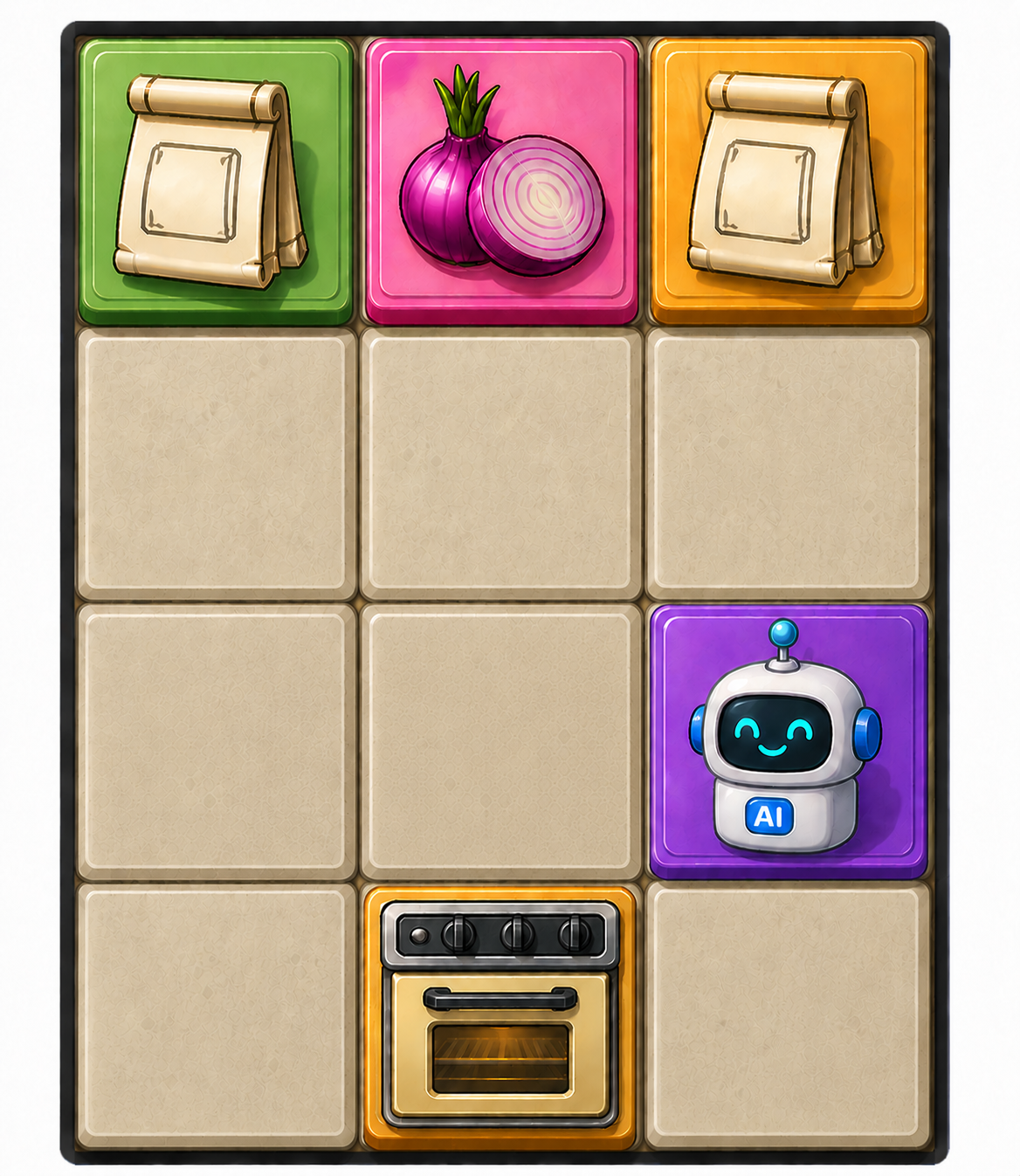}
        \caption{Temporal Bisection task 
                               }
        \label{fig:task_bisection}
    \end{subfigure}

    \caption{Chronocooked grid layout for the two core tasks. All other tasks discussed in this study are modifications of these two tasks to evaluate different temporal processing abilities of an RL agent.}
     \Description{The Chronocooked grid layout consisting of an onion dispenser(in pink), an oven(in yellow), a delivery counter (in orange or green) for the FI timing task and the temporal bisection task}
    \label{fig:tasks}
\end{figure}

The gymnasium \cite{towers2024gymnasium} based environment is a simplified version of the Overcooked environment (Overcooked-AI \cite{carroll2019utility}). The modified version consists of a single agent scenario with a 5X3 or 4X3 grid world. It features three types of counters, namely, an onion dispenser, an oven and a delivery counter, and two items, namely onions and soup (Figure \ref{fig:tasks}). The number and placement of the different counters differs by tasks and is described in the task-specific sections. 

The agent's goal is to prepare and deliver soup to the delivery counter. To do this, it has to pick up an onion from the onion dispenser, place it in the oven, and wait for the soup to cook (‘oven on’ or timing phase). The oven starts an internal (invisible) timer tied to a predefined target duration (TD), upon receiving the onion. The timing phase can vary slightly depending on the specific task. After TD or in some cases once the oven indicates a `ready' state, the agent can take the soup and deliver it at the delivery counter to complete the episode. All episodes end either after soup delivery or after a maximum of 200 time steps has been reached. The basic reward structure is that the agent receives a reward only at the end of the episode (+1 for successful task completion and 0 otherwise), while all intermediate actions receive a reward of 0. This structure may vary slightly across tasks, as described in the task-specific sections.

Each task has around eight to twelve seeds that correspond to the different start positions of the agent.  The agent can perform six actions, five for navigation (`wait', `up', `down', `left' and `right') and one `interact' action to engage with an adjacent counter. There are no obstructions for the agent's movements, except that it cannot move through the counters and walls. The observation space include information about the grid layout, the agent's position, the state of each oven (`on', `off', and, in some cases, `ready'), and the agent's state (e.g., carrying an onion or carrying a soup).  More details can be found in the repository’s documentation. 


The tasks are inspired by two established paradigms in psychology, the fixed interval (FI) timing task (Section \ref{FI_timing}) \cite{freestone2018temporal, watson2011fixed,lejeune1991comparative,fox2015timing,zeiler1994temporal} and the temporal bisection task (Section \ref{temporal_bisection}) \cite{penney2018duration, allan1991human, gibbon1981form}.  These form the two core tasks of Chronocooked. Each task includes metrics for evaluating the different temporal capabilities (Sections 5.1, 5.2, 7.1 and 7.2) and quantifying human-like timing (Sections 5.3, 7.3 and 7.4). Chronocooked also includes extensions of the two core tasks, namely, Multi-Timer (Section \ref{Multi-timer} - FI timing extension), Multi-Anchor Categorization (Section \ref{multianchor} - Temporal Bisection extension) and Asynchronous Multi-Oven (Section \ref{asyn_multioven} -  Temporal Bisection extension). In addition to evaluating the timing capabilities needed to solve the tasks, the extensions assess scalability to new timing requirements (Section 6.1). 

\section{Chronocooked Baselines}
We report the results for three types of model architectures, CNN-MLP (non-recurrent model), CNN-LSTM-MLP (recurrent model) and CNN-CTRNN-MLP (biologically inspired recurrent model). All models have been trained using the Proximal Policy Optimization (PPO) \cite{schulman2017proximal} algorithm, implemented using stable baselines3 \cite{raffin2021stable}. All recurrent architectures share the actor and critic network.  For the CNN-CTRNN-MLP model some modifications were made to the original policy implementation of recurrent PPO. The implementation code of all models including the modified CTRNN policy implementation, is available in the repository. 

The recurrent models were trained with three hidden-state sizes (also referred as memory sizes): 8,125,256. The corresponding models are referred as $CTRNN8$, $CTRNN125$, $CTRNN256$ and $LSTM8$, $LSTM125$, $LSTM256$ respectively. For brevity, we use "model" and "RL agent" interchangeably to refer to a specific architecture–memory size configuration (e.g., $LSTM256$). For a given task, all models have the same training hyper-parameter settings (see Appendix \ref{baseline_models_appendix} for more details). The number of parameters vary across recurrent models as a function of the hidden-state size and architecture. For example, $LSTM256$ has the maximum parameter capacity with 329,728 parameters in the recurrent-layer, whereas $CTRNN8$ has the least with 592 parameters (Number of recurrent-layer parameters for other models are $LSTM8$-2368 , $CTRNN125$- 23875, $CTRNN256$ - 82432 and $LSTM125$ - 95500). 


The models were trained for 900k training steps for the FI task, 960k training steps for the bisection task and 1750k training steps for all core task extensions. For each task, four runs of each model were conducted to account for different parameter initialization settings. Each run consists of different seeds corresponding to different start positions of the RL agent on the grid. Additionally, each task was tested for different environment configurations like target durations and anchor durations. The aggregation of data for the different seeds, runs and environment configuration to get model-level or architecture-level performance is discussed separately in each task.   


\section{Fixed Interval (FI) Timing}
\label{FI_timing}
The FI timing task was originally developed to study how schedules of reinforcement shape animal behaviour \cite{ferster1957schedules}. Analysis of response patterns suggested that animals tracked elapsed time rather than simply reacting to reinforcement \cite{catania1970reinforcement,gibbon1977scalar}. This led FI timing to become a standard paradigm for studying interval timing in both humans and animals. The task requires participants to respond after a fixed interval has elapsed. Responses made before the interval result in no reinforcement, while responses made after the interval are rewarded. Thus the FI task involves various components like tracking elapsed time, associating the elapsed time with the moment of reinforcement, and acting based on this association. The distribution of responses can be analysed to assess timing accuracy and within-interval temporal control \cite{freestone2018temporal, watson2011fixed, fox2015timing,lejeune1991comparative,zeiler1994temporal, wearden1988interval,gibbon1977scalar, wearden2008scalar,lejeune2006scalar}. A real-world example of FI timing is checking on bread while it bakes, without a timer. Checking too early yields no reward (bread nowhere close to ready), whereas checking after an appropriate interval increases the likelihood of finding the bread ready or nearly finished.


%

In Chronocooked, the FI task is implemented using the oven duration as the fixed interval to be timed (see Figure \ref{fig:task_fi}). After the agent places the onion in the oven, an internal (invisible) timer associated with a target duration (TD) (fixed interval) begins. Interacting with the oven before TD has no effect. The agent can collect the soup only at or any time after TD. After collecting the soup from the oven, the agent must deliver it to the delivery counter. It receives a reward of +1 for successful soup delivery, irrespective of when the agent collects the soup from the oven. All other cases and intermediate actions yield a reward of 0. The reward function is designed such that no explicit reward is given for accurate timing. Additionally, the environment does not provide any cues for the time elapsed or when the target interval has been reached. There is however, an explicit lower bound on response time, since the agent cannot take the soup before TD. This constraint is intended to replicate the original FI timing experiment, in which responding before the fixed interval has elapsed, yields no reinforcement. There is also an implicit upper bound due to the discounting factor which incentivizes the agent to deliver the soup as soon as possible. However, because discounting is an integral component of temporal-difference learning in RL, we retain this mechanism.


This task evaluates an agent's ability to acquire temporal information without explicit rewards for timing. Evaluation is performed by analysing the agent's behaviour in the timing phase. For example, an agent may repeatedly check the oven until the soup is ready, which constitutes an optimal policy because checking does not incur a penalty. However, such behaviour does not require an internal estimate of the target duration. In contrast, an agent that waits near the oven for approximately the target duration before collecting the soup provides behavioural evidence of an internal estimate of the target duration. Furthermore, the distribution of wait times can be analysed using established timing laws and biases, enabling comparisons with human timing data. 


\subsection{First oven check (Evaluation metric)}

The first oven check (FOC) is introduced in Chronocooked to quantify the agent's timing behaviour in the FI timing task. It represents the oven timer value at which the agent initiates the first of a consecutive series of `Interact' actions. If the oven timer is more than or equal to the target duration, agent can take the soup out and the corresponding oven time is recorded as the FOC for that trial. In contrast, if the oven timer is less than the target duration, the agent cannot take the soup out. In this case, if the agent consecutively continues to `Interact' with the oven until it can take the soup out (i.e., until TD), the oven timer corresponding to the first `Interact' action is recorded as the the FOC. However, if at least one of the consecutive actions is not `Interact' the oven timer is not recorded. The intuition is that the agent checks the oven because it considers the target duration to be reached and continues to check it until it can get the soup out. 

A FOC of zero means the agent begins interacting with the oven immediately upon activation and continues until a state change occurs. Such a policy is not only correct with respect to the final goal of soup delivery, it also enables faster convergence and robustness to any TD. However, it reflects a lack of time-keeping behaviour. Conversely, an FOC closer to the TD shows implicit time-keeping behaviour because it shows evidence that the agent has captured some information about the TD without any explicit reward for doing so. Thus, FOC allows us to compare the FI timing behaviour across different RL agents.

FOC is conceptually analogous to the breakpoint measure used in animal timing studies, which marks the point at which behaviour shifts from waiting to responding \cite{gibbon1977scalar,lejeune2006scalar}. Like breakpoint, FOC captures the moment an agent transitions from inaction to sustained responding with respect to oven checks. This transition is taken as evidence that it believes the target duration has elapsed. This analogy can be verified by checking whether FOC, increases linearly with the TD and does not exhibit any systematic deviations from this trend. This linear scaling is a well-established regularity in both human and animal timing literature \cite{gibbon1977scalar,lejeune2006scalar,wearden2008scalar}. If observed, it would support FOC as a quantitative behavioural measure of elapsed time, enabling FI timing behaviour to be compared not only across RL agents but also with human and animal data.


Thus, FOC can provide a measure of the degree of temporal control of an agent. This has direct relevance for HRI, for example, temporal control can help agents avoid unnecessary oven checks. Such restraint is valuable in settings where unnecessary actions can incur costs and lead to a less smooth interaction.

\subsection{Number of timing policies (Evaluation metric)}

FOC alone cannot distinguish between two qualitatively different ways an agent might time the interval: an embodied strategy, where the agent uses a fixed sequence of self-generated actions as a proxy for elapsed time (analogous to counting or repeated finger tapping), and an implicit strategy, where timing is represented internally without relying on a stereotyped action sequence. 

To capture this distinction, the number of distinct action sequences the agent exhibits in the timing phase (i.e., the timing policies) of different seeds are averaged across runs. For example, an RL agent may move `up' and `down' consecutively before interacting with the oven at TD for one seed and it may move `left' and `right' consecutively before interacting with the oven at TD for another seed. Both these would be considered two distinct action sequences. A low number of timing action sequences, close to one, indicates that the agent has learned a fixed action sequence in the timing phase to track the elapsed time across all seeds. In contrast, a high number of distinct sequences indicates greater variability in the timing policy and suggests that the agent may be relying on implicit time-keeping.

\begin{figure}[ht]
  \centering
  \includegraphics[width=0.9\linewidth]{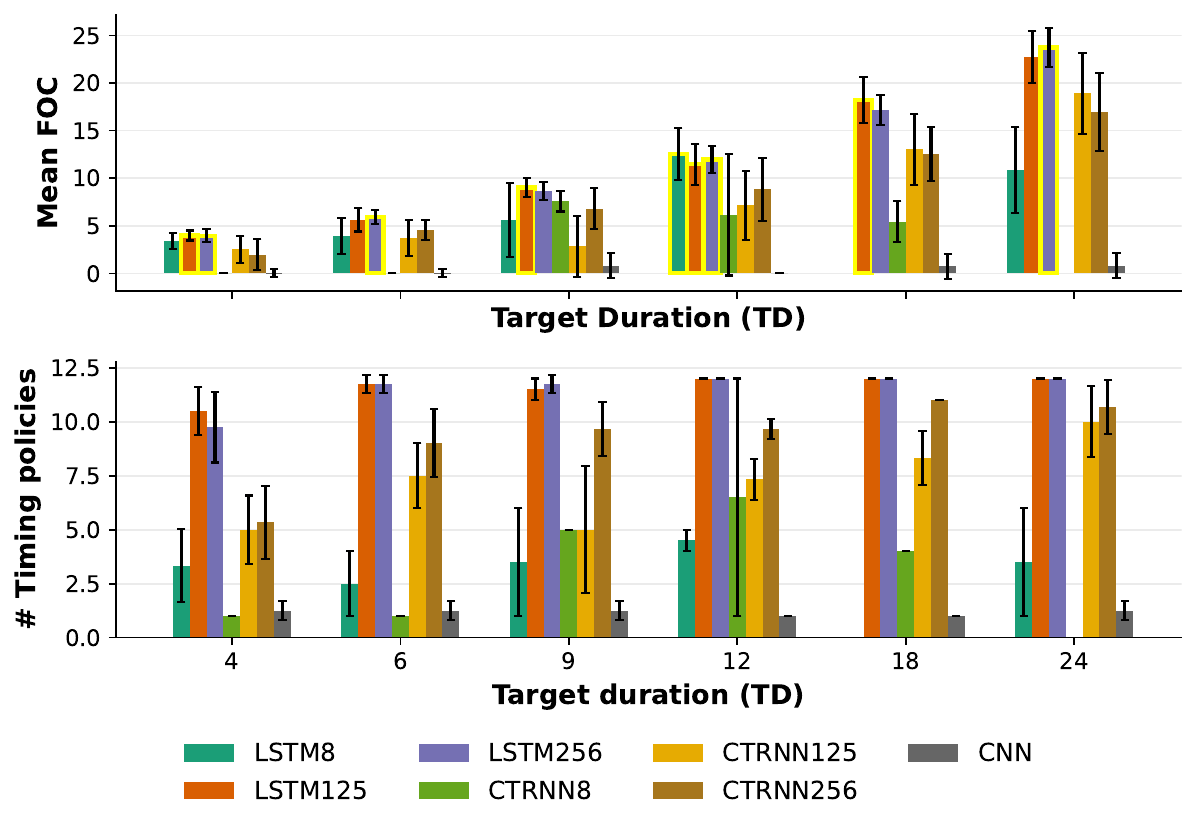}
  \caption{FI timing (task performance): First oven check (FOC) and number of timing policies (top and bottom respectively) across model types (shown in different colours) and TD (x-axis). The standard deviation is shown as error bars on corresponding barplots. Missing barplots indicate that models did not converge in the soup delivery task in any of the runs. In the top plot, the barplots with a yellow border indicate cases where mean FOC was not significantly different from the corresponding TD. }
  \label{fig:fi_foc_num_act_seq}
  \Description{The mean FOC of the high memory recurrent models especially LSTM models closely tracks the target duration. These models also exhibit variable action sequences in the timing phase. Together these results indicate a tendency of implicit time-keeping in high memory recurrent models}
\end{figure}


\subsection{Scalar timing (Evaluation metric)}
\begin{figure}[ht]
  \centering
  \includegraphics[width=\linewidth]{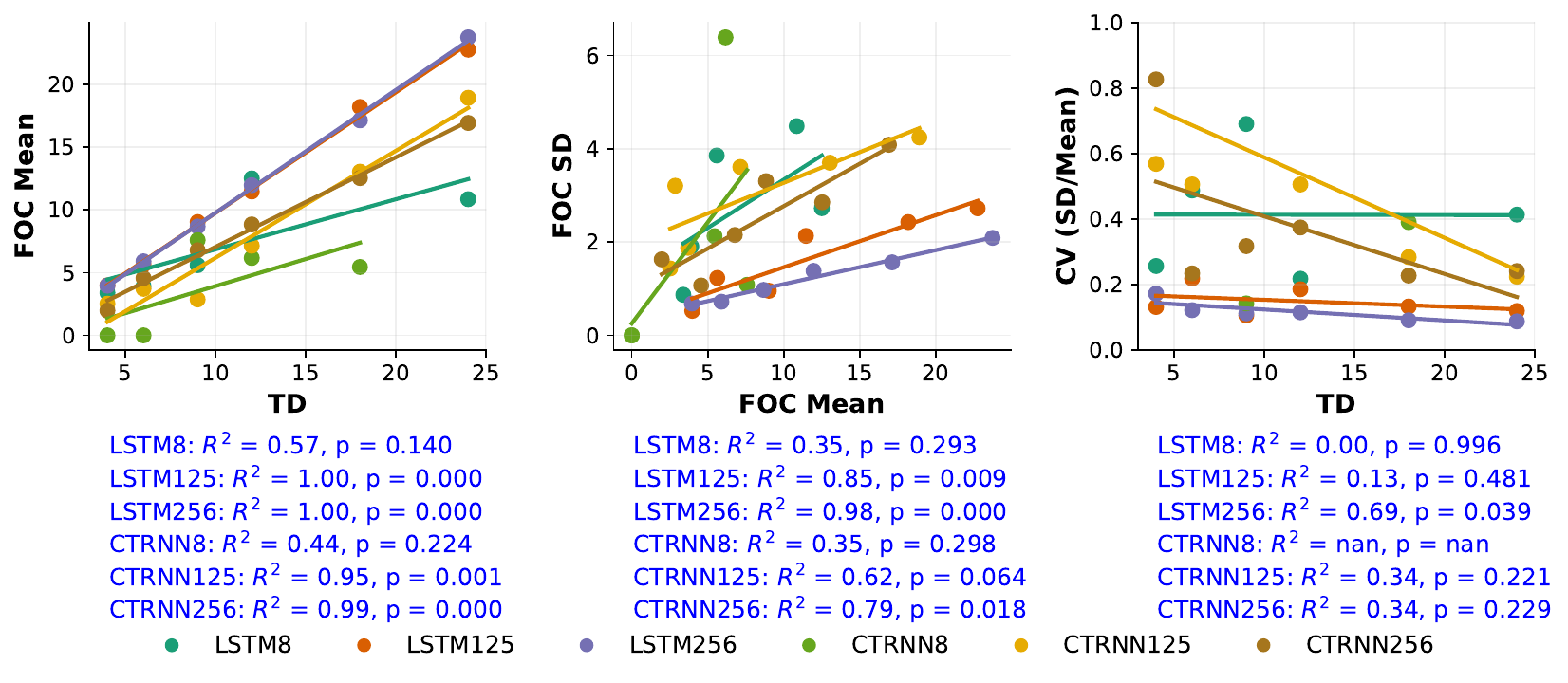}
  \caption{FI timing (human-like timing): The plots show results for the three regularities of scalar property. The leftmost plot also shows evidence for FOC validity because the mean FOC increases linearly with TDs. However, none of the models follow all scalar property regularities. The corresponding regression statistics for each plot are shown in blue at the bottom. }
  \label{fig:fi_scalar}
  \Description{LSTM125, LSTM256 and CTRNN256
showed strong positive correlation between the SD and mean FOC. The CV (SD/mean) showed no significant
relationship with TD for all models except LSTM256, consistent with the constant CV regularity expected under
scalar timing. Taken together, while some models showed partial conformity to individual regularities of the
scalar property, no model satisfied all three regularities simultaneously.}
\end{figure}

The scalar property is a well-established timing law that states that, both the mean and standard deviation of temporal responses scale linearly with the target duration \cite{gibbon1977scalar, wearden2008scalar, lejeune2006scalar}. An intuitive example of this property is when timing 10 seconds vs 10 minutes. There might be a 1 second error while timing 10 seconds but this error will be much higher while timing 10 minutes. According to this property, the timing error scales linearly with the target duration \cite{shouval2014does}. 

The scalar property comprises of three regularities \cite{wearden2008scalar,lejeune2006scalar}:
\begin{itemize}
    \item The mean of the behavioural measure increases linearly and near-accurately with TD
    \item The standard deviation (SD) of the behavioural measure increases linearly with its mean
    \item The coefficient of variation (CV), defined as SD/mean, remains constant across TDs.
\end{itemize}

To evaluate conformity to the scalar property in Chronocooked, agents are trained over a range of TDs for the FI timing task. Conformity is then assessed by examining the mean and SD of the behavioural measure (e.g., FOC) to check whether the three regularities of the scalar property described above hold. This is done by fitting linear regression models for mean-vs-TD, SD-vs-mean, and CV-vs-TD, and quantifying the strength of these linear relationships using the corresponding $R^{2}$ and p-values. 

This analysis enables comparison of agent timing behaviour with human and animal data. It is particularly relevant for HRI, as conformity to scalar timing implies that such agents may better predict human timing behaviour in the absence of other environmental cues, supporting smoother human-AI interaction. As future work, this analysis should be complemented by human studies on the same Chronocooked task (see Section 10, Future Work) for a more direct comparison between RL agent and human timing behaviour.

\subsection{Baseline results}
The different models from the three types of architectures were trained on a range of TDs (4,6,9,12,18 and 24 time steps). At least one run of all models converged and successfully performed the soup delivery task. There was however, a difference in the timing behaviour of the different models. 

\subsubsection{Implicit time-keeping} 
As shown in Figure \ref{fig:fi_foc_num_act_seq}, most models from the recurrent architectures, LSTM and CTRNN, showed evidence of implicit time-keeping, reflected in FOC values greater than zero and a high number of timing action sequences. At an architectural level, LSTM models generally showed a stronger tendency towards implicit time-keeping than their CTRNN counterparts because of both high FOC values closer to TD and greater number of timing action sequences. For both architectures, the higher memory sizes ($125$ and $256$) consistently exhibited high FOC and number of timing polices across TDs. In contrast, the non-recurrent CNN model exhibited FOC values close to zero across all TDs, along with a small number of distinct timing action sequences. Thus, it showed a lack of time-keeping behaviour. Together, these results suggest that implicit time-keeping in an FI task emerges when recurrence and sufficient memory capacity are present.


\subsubsection{FOC validity} For the models that consistently exhibited implicit time-keeping behaviour due to high FOC values and high number of timing policies across TDs ($LSTM125$, $LSTM256$, $CTRNN125$, $CTRNN256$), a strong positive linear correlation was observed between the mean FOC and TDs (see Figure \ref{fig:fi_scalar} leftmost plot). This means that there is a point in the timing phase at which the agent resorts to continuously checking the oven (FOC), analogous to the breakpoint measure in animal studies \cite{gibbon1977scalar,lejeune2006scalar}. Moreover, this point increases with TD, indicating that the agent delays continuous checking as the TD increases. Additionally, even for the smaller memory sizes ($LSTM8$ and $CTRNN8$), although no evidence of a strong positive correlation was found, the FOC tends to increase with the TD. These results support FOC as a valid quantitative behavioural measure of elapsed time in RL agents. 

\subsubsection{Scalar timing} At an architectural level, LSTM exhibited more evidence of scalar timing than its biologically inspired counterpart CTRNN.  The following observations about the regularities of the scalar property (as discussed in Section 5.3) were found: (i) Although FOC showed a strong positive linear relationship with TD (Figure \ref{fig:fi_scalar} leftmost plot) for all sufficiently large memory models ($LSTM125$, $LSTM256$, $CTRNN125$, $CTRNN256$), this trend did not always translate to FOC values statistically matching TD itself. For both CTRNN models, there was consistent deviation from TD. In the LSTM architecture as well, neither model consistently tracked TD (see Figure \ref{fig:fi_foc_num_act_seq} top plot and Appendix Table \ref{tab:FI_FOC_significance_supp} for significance tests details).  Nevertheless, considering the LSTM architecture as a whole, FOC tracked TD more closely than in CTRNN models. (ii) $LSTM125$, $LSTM256$, $CTRNN256$ showed strong positive correlation between the SD and mean FOC (Figure \ref{fig:fi_scalar} centre plot). (iii) The CV (SD/mean) showed no significant relationship with TD for all models except $LSTM256$, consistent with the constant CV regularity expected under scalar timing (Figure \ref{fig:fi_scalar} rightmost plot). Taken together, while some models showed partial conformity to individual regularities of the scalar property, no model satisfied all three regularities simultaneously, indicating that fully human-like scalar timing (as characterized theoretically in the timing literature) did not emerge in any of the models tested.

\section{Multi-Timer (FI Timing extension)}
\label{Multi-timer}

Many real-world timing tasks require actively producing a target duration with some accuracy. Using the bread baking example as before, checking the bread too early may delay or interrupt the baking process, whereas checking too late may result in an overcooked bread. Hence, there is a penalty involved for checking too early or too late. Humans are good at producing different intervals without environmental cues. The time production in humans is analogous to the FI interval timing task, where produced time accurately follows the target interval with a constant CV \cite{wearden1988interval}. Additionally, humans are also good at producing multiple temporal intervals signalled by different environmental cues \cite{labliuk2015determinants, guilhardi2010effect}. However, research on time production with multiple intervals is limited and mostly focused on animals \cite{buhusi2009interval,caetano2012stimulus,rice2014pigeons, nepomoceno2020temporal}. Taking inspirations from the methodology in these studies we introduce the multi-timer task in Chronocooked to study time production and multi-timing abilities in RL agents.  


In Chronocooked, the multi-timer task which is an extension of the FI timing task, involves  producing time with some accuracy. The basic grid layout of the FI timing task remains the same. However, the explicit temporal lower bound from the FI timing task is removed and the agent can take soup out of the oven at any time. Consequently, there is a corresponding reduced reward for taking the soup out too early or too late. Hence, instead of allowing a flexible timing policy, the task introduces a timing buffer. The agent must retrieve the soup within a window of $TD \pm$ buffer duration to receive the full reward of +1 after successful soup delivery (a successful timing and soup delivery task). Retrieving the soup too early or too late results in a reduced reward in spite of successful soup delivery (unsuccessful timing task but successful soup delivery task). All intermediate actions and unsuccessful soup deliveries result in 0 reward. 

The task also supports multiple TDs indicated by different oven states for each TD (analogous to different coloured indicator lights). The different oven states provide a cue to the agent about which TD must be timed. To study how the RL agent adapts to increasing number of TDs, the task is trained using curriculum learning (CL). In each CL stage, after a predefined performance threshold is reached, a new TD is introduced. Thus, in later CL stages, the RL agent must associate different oven states with their corresponding TDs and produce the appropriate duration to receive a reward of +1 for successful soup delivery. 

The task evaluates if an RL agent can handle time production task with or without a buffer duration. The buffers can be used to control the task difficulty. For example, buffer $=$ 0 requires the agent to time the TD exactly, enabling the study of exact timing abilities in the different RL agents. The task additionally evaluates multi-timing capabilities in RL agents and provides a controlled way to study how agents adapt to multiple timers. Agents that can produce a target duration in response to different environmental cues, and that can readily learn to associate new cues with new target durations, would be valuable for real-world tasks. For example, waiting for 10 seconds after the soup starts boiling, but waiting 30 seconds after the oven goes off.



\subsection{Scalability (Evaluation metric)}
Chronocooked introduces two scalability metrics that enable a controlled analysis of how RL agents adapt to new timing requirements. While additional memory or training may enable an agent to accommodate new timing requirements, our focus is on scalability arising from a flexible timing strategy.

\subsubsection{N-capacity} The first metric is N-capacity. It is defined as the maximum CL stage reached by a model before the maximum training budget. If multiple models reach the same final CL stage, the percentage of runs that converged (reached a performance of 0.8) in that stage is used for comparison. Combined with task performance in the final CL stage, this metric indicates a model’s capacity to incorporate multiple timing requirements in a fixed training budget. In case of the multi-timer task, the timing requirements are new TDs and this metric is referred as N-timer capacity.  

\subsubsection{CL-efficiency} The second metric is CL efficiency. It is quantified using CL-efficiency ratio, which is the ratio of the training steps required for a CL stage to those required for the preceding CL stage. This measure allows comparison of how training requirements change as new timing requirements (in this case, new TDs) are introduced. An RL agent with a flexible timing strategy must be able to incorporate each new TD with same or progressively fewer training steps. The assumption is that, the agent learns supplementary tasks, such as navigation and soup delivery, in the first CL stage. From the second stage onward, only the timing requirement changes, the other tasks and goals remain unchanged. Hence, a CL-efficiency ratio <=1 indicates a flexible timing strategy. This metric also enables comparison between the different RL agents to identify RL agents with more efficient CL than others. 


Humans and animals are able to rapidly recalibrate to new timing demands with minimal additional experience \cite{aggadi2025rats}. Flexible adaptation to new timing requirements indicates that the underlying timing task has been learned rather than a specific timing requirement being memorised. It is therefore an important aspect of temporal processing and may also be valuable in HRI, where such requirements commonly arise in everyday tasks. The scalability metrics introduced in Chronocooked provide a basic indication about flexible timing strategies based on CL training data. Future work can focus on the underlying neural representations to determine whether the same temporal representation is flexibly used for each new requirement.


\subsection{Baseline results}

All models were trained on the Multi-timer FI task using CL. The results for the initial CL stages are used to study time production abilities of all the models and the later stages are used to study multi-timing abilities in the recurrent models. The task includes two variants: a categorical variant (with timing buffer) and an exact-timer variant (without buffer). Since most models showed more reliable multi-timing performance in the categorical variant, the multi-timing results presented in this section focus on this variant while the time production results consider both variants. The Appendix \ref{multi-timer_appendix} provides more details about the task variants (including results of all models for each variant), task specifications like TD configurations, details on CL training and an extended version of the baseline results. The main findings relevant to the current study are summarised below. 

\begin{figure}[ht]
  \centering
  \includegraphics[width=0.65\linewidth]{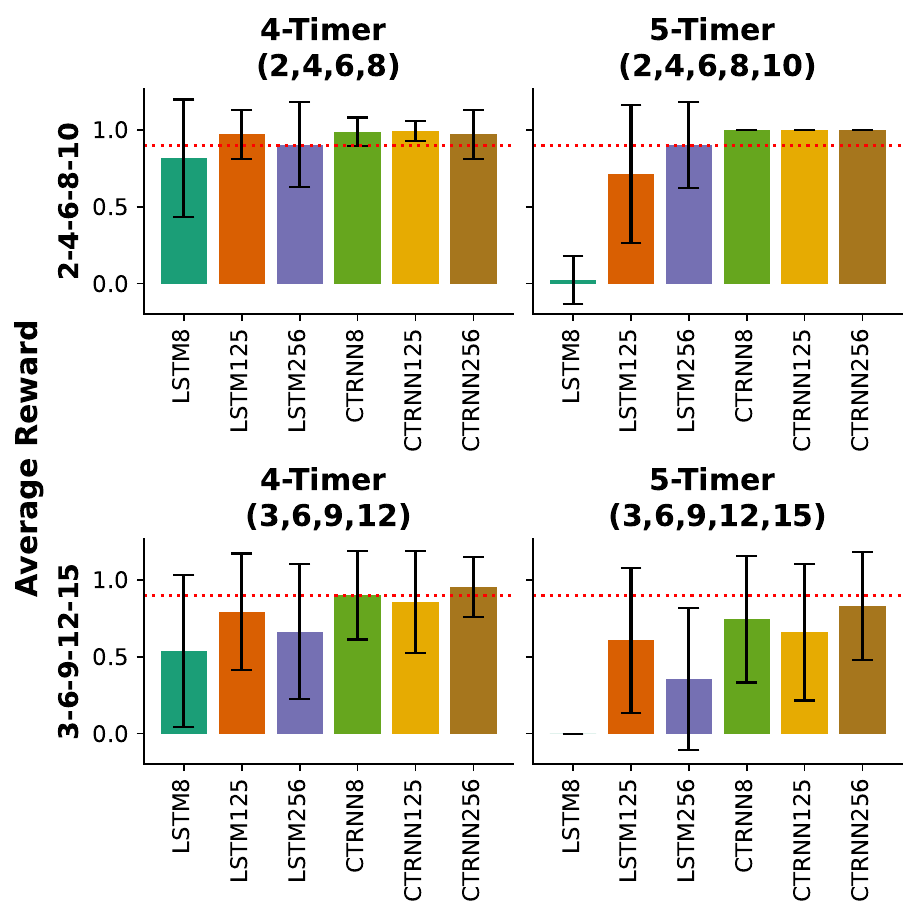}
  \caption{Multi-timer categorical variant (task performance): The plots show the average reward of the different models for two TD configurations (row titles) in the last two CL stages, namely, 4-timer and 5-timer setting (column titles) for the multi-timer task with buffer. The error bars depict the SD across the different model runs and seeds. The red dotted line indicates the CL performance threshold. }
  \label{fig:multitimer_taskperf_cat}
  \Description{All CTRNN models perform better than the LSTM models in both TD configurations. LSTM8 shows the worst performance in the 5-timer setting.}
\end{figure}

\subsubsection{Time production performance} The models showed varying capabilities for the time production task. Models from both recurrent architectures (LSTM and CTRNN) successfully performed the time production task with a buffer (see Appendix figure \ref{fig:multitimer_scalability_cat_supp}). However, the CTRNN models consistently failed to perform the task without a buffer (exact-timing variant), even with just 1 timer. In contrast, all the LSTM models successfully performed the exact-timing task with at least 1 timer (see Appendix figure \ref{fig:multitimer_scalability_exact_supp}). The CNN model could perform the time production task both with and without a buffer, although only one run was successful in the latter condition. These observations indicate an architectural shortcoming in CTRNN for exact-timing that needs further investigation. 

\subsubsection{Multi-timing performance}
While models from both recurrent architectures showed multi-timing capabilities with a buffer (categorical variant), CTRNN models showed a better overall task performance than LSTM models in the last CL stage, namely the 5-timer setting (see Figure \ref{fig:multitimer_taskperf_cat}). Even the low memory $CTRNN8$ model on avergae performed better than the high memory LSTM models across both TD configurations. Hence, within a fixed training budget (1750k training steps), CTRNN models were in general better than the LSTM models at associating multiple oven states with the corresponding TD .

\subsubsection{N-timer capacity} At an architectural level, CTRNN showed better capacity than LSTM in the categorical variant. While models from both architectures reached the 5-timer setting (last CL stage), 83\% of CTRNN runs converged (i.e. reached a performance threshold of 0.8)  in the 5-timer setting before reaching the maximum training budget (1750k training steps), compared to only 42\% in LSTM (see Appendix figure \ref{fig:multitimer_scalability_cat_supp} for training steps data of categorical variant).

\subsubsection{CL efficiency} An architecture-level analysis of the CL-efficiency ratio in the categorical variant showed that, CTRNN exhibited slower learning than LSTM in CL stages associated with the acquisition of a new timing capability. However, it reached a comparable CL-efficiency ratio in the later CL stages. The following observations support this claim :
\begin{itemize}
    \item  \textbf{CL stage 0}: CTRNN (N=23, M=19.23, SD=20.84, median=10.17) took significantly longer compared to LSTM (N=24, M=5.04, SD=2.58, median=4.74) to converge in the first CL stage (Mann-Whitney U=473.00, p<0.001). 
    \item \textbf{CL stage 2}: This stage corresponded to the first introduction of the multi-timing scenario in the categorical variant. The CL-efficiency ratio of CTRNN (M = 2.39, SD = 2.06, median = 1.18, Wilcoxon W = 168.00, p = 0.035) was significantly more than 1 but for LSTM (M = 0.84, SD = 1.24, median = 0.51, Wilcoxon W = 52.00, p = 0.997) it was not. The CL-efficiency ratio of CTRNN was also significantly greater than LSTM (Mann-Whitney U, stat=369.00, p=0.001). 
    
    \item \textbf{Intermediate CL stages}: For the CL stages, where no new task element other than a new TD was introduced, such as 3 or 4-timer settings, both architectures showed CL-efficiency ratios that were not significantly above 1. There was also no significant difference between the two architectures.
    \item  \textbf{Last CL stage}: This stage corresponded to the 5-timer setting. As the task complexity accumulated, both architectures required comparable additional training steps with CL-efficiency ratio significantly above 1.
\end{itemize}

Hence, while both recurrent architectures show multi-timing capabilities in the categorical variant, CTRNN's better task performance in the final CL stages and higher N-timer capacity suggest that CTRNN exhibited a more scalable multi-timing strategy overall. Additionally, although CTRNN required significantly more training steps than LSTM in the initial CL stages, it was able to catch up in later stages with LSTM, which exhibits a CL-efficiency ratio consistently below 1 (except in the final CL stage).

\section{Temporal Bisection}
\label{temporal_bisection}

The temporal bisection task is a widely used paradigm in timing research to study binary classification of time intervals. The task has been used to infer important information about temporal encoding and subjective timing in humans and animals \cite{gibbon1981form, allan1991human, wearden1991human, kopec2010human,penney2018duration, droit2001temporal}. It involves two anchor durations, short ($S$) and long ($L$). Participants are trained with 50\% $S$ and 50\% $L$ trials to identify each anchor duration as $S$ or $L$. During test, probe durations, which include the two anchor durations and durations in between, are introduced and participants have to categorize each duration as closer to $S$ or $L$. Hence, the temporal bisection task involves multiple processes like tracking elapsed time for each anchor duration,  storing these anchor durations in memory, tracking elapsed time for the probe durations, comparing the probe duration with the stored anchor durations in memory, and making a decision of $S$ or $L$ based on this comparison. Additionally, the probe durations between the two anchor durations have no objectively correct category for $S$ or $L$. Hence, these probe durations can be used to study subjective time of the participants. Consequently, the distribution of responses for this task carry important information about a participant's temporal encoding and subjective timing. 



This task is implemented in the Chronocooked environment by using two delivery counters, one for $S$ and one for $L$ (see Figure \ref{fig:task_bisection}). The anchor durations ($S$ and $L$) are fixed at the start of the training phase. The agent's goal is to delivery the soup to the correct delivery counter depending on how  long the oven has been on. In the training phase, after the agent places the onion in the oven, the oven runs for a duration corresponding to either $S$ or $L$ duration with 50\% probability. After the specific duration has elapsed, the oven transitions to a `ready' state. The agent must then take the soup and deliver it to the correct delivery counter. Interacting with the oven before it is ready has no effect. The agent receives a reward of +1 for correct delivery, irrespective of the time it takes to deliver the soup or the time it takes to collect the soup from the oven. All intermediate actions and incorrect deliveries yield a reward of 0. In the test phase, along with the anchor durations, the agent is evaluated on all intermediate durations as well as durations slightly beyond the range of the anchor intervals.

The Chronocooked bisection task evaluates whether an RL agent can associate the time elapsed while the oven was on, with the correct delivery counter. Additionally, since the intermediate probe durations are, by definition, out-of-distribution (OOD) for the agent, the task also allows us to examine how the agent behaves when faced with them. The agent may fail to perform the task altogether, (unsuccessful soup delivery), it may still deliver the soup successfully but assign the delivery counter randomly (a successful soup delivery task but a failed timing task), or it may assign the delivery counter based on the closeness of the probe duration to the training phase anchors (reflecting a graded timing-based categorization). The agents' responses to the intermediate probe durations can also be compared to different regularities observed in temporal bisection data from human experiments to quantify human-like timing.  



\subsection{Psychometric curve (Evaluation metric)}
A Psychometric curve is the standard behavioural measure in psychology for temporal bisection experiments. The curve plots the probability of responding $L$, $P(\mathrm{long})$, as a function of the probe durations. Because the intermediate probe durations have no objectively correct response, the psychometric curve captures not only overall task performance but also how subjective time changes across the range of probe durations. Human psychometric curves typically exhibit a sigmoidal shape, with the probability of a $L$ response increasing monotonically from approximately 0\% at the short anchor to 100\% at the long anchor. The shape and location of the curve form the basis for several quantitative measures, such as the bisection point and Weber fraction, that characterize human time perception \cite{gibbon1981form, allan1991human, wearden1991human, kopec2010human, penney2018duration, bausenhart2018assessing}.

The Chronocooked temporal bisection task, similarly uses a psychometric curve to study the task performance and temporal behaviour of an RL agent. The percentage of $L$ responses, $P(\mathrm{long})$, for a given probe duration is calculated as the proportion of deliveries to the delivery counter corresponding to the $L$ anchor among seeds for which the agent successfully delivered soup to one of the counters. Thus, failed soup delivery cases, in which the agent does not deliver soup to any delivery counter, are excluded from the calculation of $P(\mathrm{long})$, while failed timing tasks, in which the agent delivers the soup to the wrong counter, are retained. Additionally, if more than 50\% of the seeds exhibit a failed soup delivery at a given probe duration, that duration is excluded from the calculation. The resulting P(long) is plotted against different probe durations ranging from 1 to $L$ + 4 time steps. Thus including OOD durations along with the anchor durations.  

The psychometric curve is obtained by fitting a logistic curve to the $P(\mathrm{long})$ data:
\begin{equation}
f(x) = \frac{1}{1 + \exp\left(-\frac{x-x_0}{k}\right)}
\label{eq:psychometric_log_fit}
\end{equation}
where $x_0$ represents the function's midpoint, and $k$ controls the slope of the curve. 
The response distribution can then be analysed for the success rate, temporal encoding and OOD temporal bisection in RL agents. Furthermore, it can be compared with human psychometric curves to assess similarities in temporal bisection behaviour

\begin{figure}[ht]
  \centering
  \includegraphics[width=\linewidth]{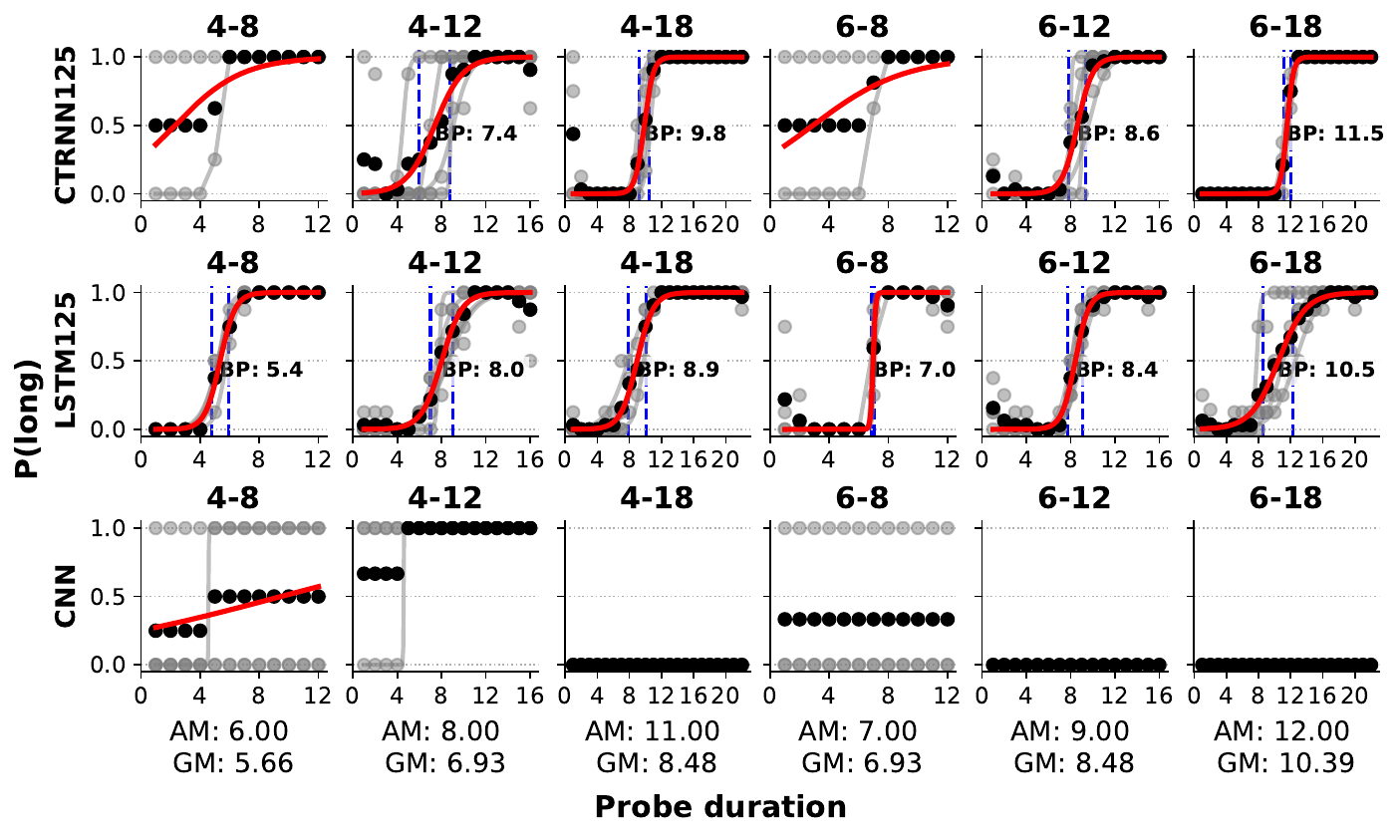}
  \caption{Temporal bisection task (task performance): Psychometric curves of the three model architectures (rows) for the different anchors (columns). Each plot shows P(long) plotted against the different probe durations. The arithmetic mean (AM) and geometric mean (GM) of each anchor are shown at the bottom of each column. The grey dots represent p(long) of each run and the solid grey line shows the corresponding logistic fit. Similarly, the black dots represent the average P(long) across all runs and the red line shows the logistic fit on the average P(long). The vertical blue dotted lines corresponds to durations with P(long)$=$25\% and P(long)$=$75\% respectively. The region between the two blue dotted lines is called the Just Noticeable Difference (JND). Smaller JND corresponds to higher discriminability and vice versa. The bisection points (BPs) (duration at which P(long)$=$50\%) are shown in the respective plots.}
  \label{fig:psychometric_curves}
  \Description{ The RL agents with sufficient memory and a recurrent architecture can both converge on the soup
delivery task and perform the bisection task including OOD. They also exhibit an s-shaped response curve which is qualitatively similar to human and animal data.}
\end{figure}

\begin{table}
  \caption{Temporal bisection task (task performance)}
  \label{tab:bisection_task_performance}
  \begin{tabular}{@{}ccl@{}}
    \toprule
     \textbf{Model} & \textbf{Soup delivery (\%)} & \textbf{Bisection success (\%)} \\
    \midrule
    CNN & 100.0 & 8.3 \\ 
    CTRNN-125 & 85.4 & 75.0 \\ 
    CTRNN-256 & 97.9 & 93.8 \\
    LSTM-256 & 100.0 & 97.9 \\
    LSTM-125 & 100.0 & 100.0 \\
    CTRNN-8 & 16.7 & 100 \\ 
    LSTM-8 & 22.2 & 100 \\ 
 
  \bottomrule
\end{tabular}
\end{table}

\subsection{Task performance (Evaluation metric)}


We define a successful temporal bisection as one in which the RL agent's $P(\mathrm{long})$ monotonically increases with the probe duration. Hence, a run is considered successful if the resulting $P(\mathrm{long})$ fits equation \ref{eq:psychometric_log_fit}. Additionally, for data that does not fit the logistic function, we extend the definition of a successful bisection task to include cases where $P(\mathrm{long})$ is less than 0.2 for probe durations between 0 and $S$ and greater than 0.8 for probe durations between $L$ and $L+4$. This criterion allows RL agents that reliably distinguish between the two anchor extremes to be classified as successfully performing the timing task.

The anchor configurations, as in psychology literature \cite{kopec2010human}, are used to vary the difficulty level of the temporal bisection task. The difficulty level is defined by the $spread$ of the anchor durations. It is calculated as: 
\begin{equation}
 \mathrm{Spread} = \frac{L}{S}
 \label{eq:spread}
\end{equation}
where $L$ and $S$ denote the long and short anchor durations, respectively. $spread$ is inversely proportional to the bisection task difficulty. 

We differentiate between a failed soup delivery and a failed temporal bisection task. An RL agent may fail the soup task altogether due to failure to converge. However, an RL agent that successfully performs the soup delivery task may still fail to perform the temporal bisection task (soup delivered to the wrong delivery counter). The latter case isolates failures specific to temporal bisection. Hence, we restrict all analysis to only successful soup deliveries. 

\subsection{Bisection point (Evaluation metric)}

The psychometric curve from human bisection data exhibits a monotonic increase in $P(\mathrm{long})$  from the $S$ to $L$ anchor. The duration at which a participant is equally likely to respond $S$ or $L$ is called the bisection point (BP) or point of subjective equality. The BP point carries important insights about temporal processing and time perception \cite{penney2018duration, kopec2010human}. The location of the BP can provide insights into the underlying temporal encoding mechanism. Theoretically, a BP at the geometric mean (GM) of the anchor durations has been associated with logarithmic encoding of time \cite{gibbon1981form,allan1991human,allan2002location}, whereas a BP at the arithmetic mean (AM) has been associated with linear encoding \cite{wearden1991human}. In practice however, a number of variations are observed in the location of BP. A number of studies have attempted to explain and rationalize these variations in BP data \cite{kopec2010human,levy2015memory}. 

Kopec and colleagues \cite{kopec2010human} consolidated findings from 148 human timing experiments and identified three regularities in BP as a function of the $spread$ ($L/S$).  
\begin{itemize}
    \item The BP transitions from being below the GM at smaller spreads, to being closer to the GM, and eventually to being below the AM as the spread increases
    \item When normalized by the AM of the anchor durations, the BP is sub-arithmetic and decreases only slightly (up to approximately 90\% of AM) as the spread increases
    \item When normalized by the GM, the BP is sub-geometric at smaller spreads and becomes supra-geometric as the spread increases.
\end{itemize}

In Chronocooked, the BP is calculated using the mean psychometric curve (shown in red in  Figure \ref{fig:psychometric_curves}). Specifically, it is the duration at which the inverse logistic function fitted to the psychometric curve obtained after averaging $P(\mathrm{long})$ across all runs, reaches a probability of $0.5$. The BP is recorded for different $spreads$ and anchor duration values to check for the three regularities mentioned above using the same methods as in \citet{kopec2010human}.  This method not only helps to gain more insights on temporal encoding of RL agents but also enables comparison with human data.

Similarity in the characteristics of the BP can provide evidence for similarity in temporal encoding which is useful for HRI settings. For example, an RL agent with a monotonically increasing response curve for the temporal bisection task, and a BP that tracks human data, would better predict which delivery counter a participant is likely to approach when presented with a probe duration between the two anchors. This is particularly relevant for HRI, as it allows agents to better anticipate human responses in ambiguous timing scenarios, ultimately enabling more natural and effective human-robot interaction. 


\subsection{Weber's ratio (Evaluation metric)}

Weber's law is a well-established psychophysical law that states that the amount of change needed to notice a change in a stimulus increases proportionally with the stimulus magnitude \cite{laming2009weber}. An intuitive example of this property would be a faulty timer. A 10-second deviation in a timer set for 20 seconds is more noticeable than the same 10-second deviation in a timer set for 10 minutes. Thus, replication of Weber's law would imply that an RL agent perceives changes in time in a manner similar to humans.

In the Chronocooked temporal bisection task, Weber's law is quantified using the Weber ratio, similar to psychology literature \cite{gibbon1981form,allan1991human, wearden1991human, kopec2010human}. It is defined as, 
\begin{equation}
\mathrm{Weber\ ratio} = \frac{t_{75} - t_{25}}{BP}
\label{eq:weber_ratio}
\end{equation}
where $t_{25}$ and $t_{75}$ are durations corresponding to 25\% and 75\% and of $L$ responses respectively (indicated with blue dotted lines in Figure~\ref{fig:psychometric_curves}). The difference between these two durations is called the just-noticeable difference (JND). It represents the amount by which a duration must change for the difference to become noticeable. A smaller JND indicates greater temporal discrimination ability, whereas a larger JND indicates weaker discrimination ability.

The data consolidated by Kopec and colleagues \cite{kopec2010human} across 148 experiments revealed two  regularities in the Weber ratio.
\begin{itemize}
    \item The discrimination ability (which is inversely proportional to weber ratio)  decreases with the spread between the anchor durations
    \item When restricted to the same spread, the Weber ratio shows no correlation with absolute duration and thus shows evidence for weber's law. 
\end{itemize}

These regularities provide benchmarks for assessing whether the Weber ratio of an RL agent exhibits patterns similar to those observed in human timing data. As with BP, this analysis must be complemented by direct comparisons of Weber ratios between humans and artificial agents using Chronocooked as a common experimental framework. 



\subsection{Baseline results}

The models were tested across 14 anchor configurations ($S$-$L$): 4–12, 6–8, 2–24, 2–32, 3–24, 3–32, 4–8, 4–18, 6–12, 6–18, 7–14, 7–21, 3–6, and 3–9. These anchor configurations correspond to varying levels of difficulty, defined by the $spread$ (Equation \ref{eq:spread}). Figure \ref{fig:psychometric_curves} shows the response distribution (grey dots) and the psychometric curves (solid grey lines) for the three types of architectures (row titles) across different anchor durations (column titles). For the CTRNN and LSTM models, memory size 125 is shown in the main text, while the corresponding psychometric curves for memory size 256 and 8 are provided in Appendix \ref{bisection_appendix} Figure \ref{fig:psychometric_curves_supp}. 


\begin{figure}[ht]
  \centering

  \begin{subfigure}{\linewidth}
    \centering
    \includegraphics[width=0.64\linewidth]{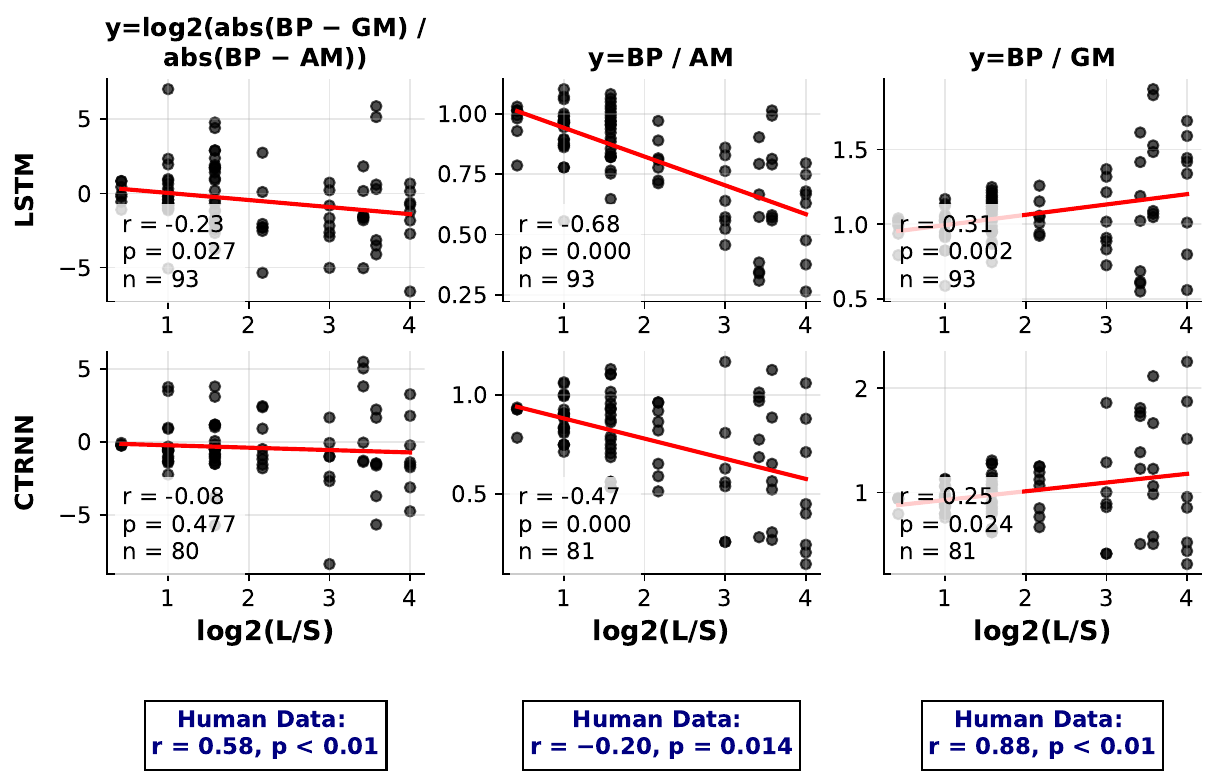}
    \caption{Bisection point analysis}
    \label{fig:bp_plots}
  \end{subfigure}

  \begin{subfigure}{\linewidth}
    \centering
    \includegraphics[width=0.64\linewidth]{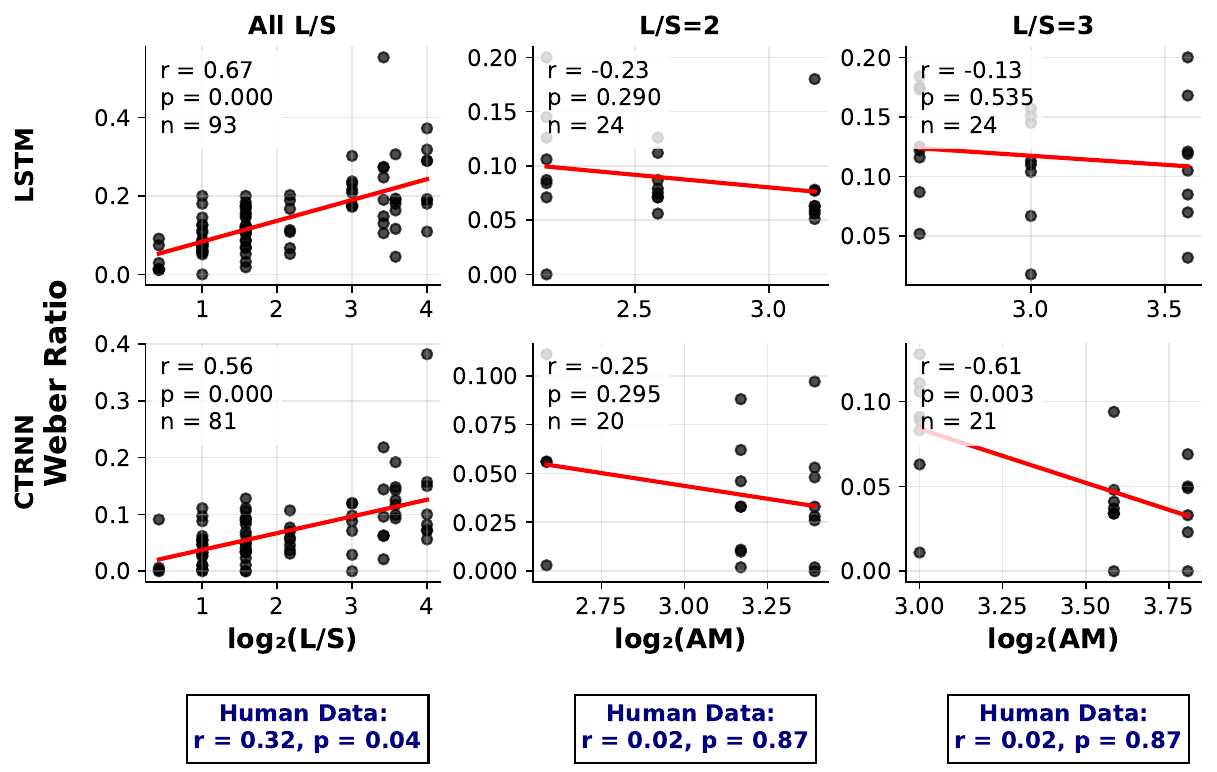}
    \caption{Weber's ratio analysis}
    \label{fig:wf_plots}
  \end{subfigure}

  \caption{Temporal bisection task (Human-like timing): (a) The plots show three BP regularities (y-axis)  as a function of spread (x-axis). Both architectures show that BP tracks GM and stays around 1.2xGM even at larger spreads, unlike humans, where BP track the AM. (b) The plots show Weber ratio against spread (first column) and against AM for spreads of 2 and 3 (second and third column, respectively). Both architectures show decreasing discriminablity with spread, consistent with human data. For a specific spread, both architectures show a stronger positive trend in Weber ratio than human data (r=0.02) and CTRNN violates Weber’s law at spread 3. (a-b) Each plot shows the corresponding statistics of the regression fits,namely, the Pearson correlation coefficient (r), p-value (p) and number of samples (n). The results for corresponding human data taken from \citet{kopec2010human} are shown at the bottom of each columns.}
  \label{fig:temporal_analysis}
  \Description{The BP plots show that for both architectures, BP tracks GM and stays around 1.2xGM even at larger spreads, unlike humans, where BP track the AM. The weber ratio plots show that, both architectures exhibit decreasing discriminablity with spread, consistent with human data. For a specific spread, both architectures show a stronger positive trend in Weber ratio than human data (r=0.02) and CTRNN violates Weber’s law at spread 3 }
\end{figure}

\begin{table}
\caption{Temporal bisection task (Human-like timing): Summary of BP and Weber ratio regularities relative to human data.}
\label{tab:bisection_human_data_comparison}
\centering
\begin{tabularx}{\linewidth}{@{}lccX@{}}
\toprule
\textbf{Effect} & \textbf{LSTM} & \textbf{CTRNN} & \textbf{Comments (contrast from human data)}\\
\midrule
BP: sub-geometric $\rightarrow$ sub-arithmetic
& $\times$ & --
& Opposite direction (both models). \\

BP: BP/AM vs log2(L/S) (BP track AM)
  & $\times$ & $\times$
  & Same trend as humans, but with a steeper decrease (both models)\\

BP: sub-arithmetic for small spreads
  & $\times$ & $\checkmark$
  & However, CTRNN declines with spread rather than remaining at a constant fraction of AM. \\
  
BP: BP/GM vs log2(L/S)
  & $\times$ & $\times$
  & Same trend as humans, but with a less steep increase (both models) \\
  
BP: supra-geometric for high spreads
  & $\times$ & $\times$
  & BP remains below 1.2 times GM (both models) \\
  
Weber ratio (discriminability)
  & $\checkmark$ & $\checkmark$
  & Same trend as humans, but with a steeper increase (both models) \\

Weber's law
  & -- & $\times$
  & CTRNN violates Weber's law; no evidence for LSTM. \\
\bottomrule

\end{tabularx}
\end{table}

\subsubsection{Task performance} Most models across the three architectures, successfully converge for the soup delivery task. However, there is a difference in the performance on the temporal bisection task. Table \ref{tab:bisection_task_performance} shows the percentage of successful soup-delivery runs, and the percentage of successful bisection tasks among runs that successfully performed the soup-delivery task (as defined in Section 7.2 and shown by the grey solid lines in Figure \ref{fig:psychometric_curves}). For the recurrent models, sufficient memory size appears necessary to reliably converge and learn the soup-delivery task. LSTM models with memory sizes of 125 and 256 achieved 100\% success for soup delivery task, whereas the CTRNN models reached a comparable success rate only at a memory size of 256. This difference may reflect the greater capacity of the LSTM architecture to retain information over time. LSTM models also generally perform the temporal bisection task with a higher success rate than their CTRNN counterparts. These results show that RL agents with sufficient memory and a recurrent architecture can both converge on the soup delivery task and perform the temporal bisection task well. 

\subsubsection{Temporal-difference learning not sufficient for temporal bisection}
Interestingly, the CNN based RL agent, despite having no explicit memory mechanism, successfully learned the soup-delivery task and achieved a small degree of success (8.3\%) on the bisection task. Given the reliance of the bisection task on temporal information, this is noteworthy. A Previous study \cite{deverett2019interval} had shown that a feedforward model could successfully perform the time-reproduction task. This capability was attributed to temporal-difference learning  in RL, which can provide a source of temporal information without requiring an explicit memory mechanism. However, in the present study temporal-difference learning alone does not appear sufficient to achieve robust performance on the temporal bisection task, as reflected by the substantially lower success rate of the CNN compared with the recurrent architectures.

\subsubsection{Psychometric curve}Interestingly, the high-memory recurrent models, namely, $LSTM125$, $LSTM256$, $CTRNN125$, and $CTRNN256$, that successfully perform the temporal bisection task exhibit a sigmoid-like psychometric curve (Figure \ref{fig:psychometric_curves}). This shows that the responses of these RL agents qualitatively match human bisection response data. In contrast, the CNN model, which has a low success rate on the bisection task, exhibits a step-function-like response distribution with a sharp shift from  $S$ to $L$. It therefore does not qualitatively match human bisection data. Nevertheless, the psychometric curve analysis shows that, irrespective of architecture and memory size, all RL agents that successfully learn the bisection task also successfully bisect OOD probe durations without any extra training. This provides preliminary evidence of temporal bisection capabilities in RL agents.

\subsubsection{Human-like timing}
In order to quantitatively evaluate the similarity to human psychometric curve we analyse the regularities in the BP and Weber ratio as described in Section 7.3 and 7.4 for the high-memory recurrent models, namely, $LSTM125$, $LSTM256$, $CTRNN125$, and $CTRNN256$. BP and Weber ratio did not significantly differ between memory sizes for either LSTM (BP: f=2.65, p=0.107, Weber ratio: f=0.56, p=0.456) or CTRNN (BP: f=0.03, p=0.856, Weber ratio: f=0.18,p=0.671), after accounting for anchor configuration. Hence, to simplify the analysis we perform the BP analysis at the architectural level. Figure \ref{fig:bp_plots} and \ref{fig:wf_plots} shows the results after following the analysis methods in \citet{kopec2010human}. Each plot contains the corresponding Pearson correlation coefficient (r), p-value (p) and number of samples (n) for the linear regression fit. The results for corresponding human data taken from \citet{kopec2010human} are shown at the bottom of each column. Table \ref{tab:bisection_human_data_comparison} summarizes the comparison of BP and weber ratio for the two architectures against human data from \citet{kopec2010human} (Refer Appendix \ref{bisection_appendix} for a detailed discussion of each plot in Figure \ref{fig:bp_plots} and \ref{fig:wf_plots}). The following significant observations were found: 

\begin{itemize}
    \item \textbf{Bisection point (BP):} The relationship between BP and AM or GM was reversed for the RL agents compared to humans (Figure \ref{fig:bp_plots}). In human data, BP tracks AM and remains at or below approximately 0.9xAM across spreads, whereas in the RL agents, BP tracks GM and remains at or below approximately 1.2xGM. At smaller spreads, AM and GM are closer and the two groups show similar behaviour. As spread increases, AM moves further above GM, revealing that BP is generally lower in the RL agents than in humans. This indicates a greater propensity for the RL agents to respond $L$. Further investigation of neural activity may help uncover the mechanisms underlying this behaviour.

    \item \textbf{Discrimination ability:} Discrimination ability (as defined by the Weber ratio) decreased with increasing $spread$ (Figure \ref{fig:wf_plots}, first column). This was consistent with human data. However, the decrease was steeper in the RL agents (LSTM:r=0.67, p<0.05, CTRNN: r=0.56, p<0.05) in comparison to human data (r=0.32, p=0.04), which indicates a greater reduction in discrimination ability as difficulty (defined by the $spread$) increases. Additionally, between the two recurrent architectures the decrease in LSTM was significantly steeper than that in CTRNN (t=2.97, p=0.03).

    \item \textbf{Weber's law:}   For spread=2, neither architecture showed a significant relationship between Weber ratio and AM (see Figure \ref{fig:wf_plots} second column), consistent with the observation in human data. However, both architectures showed a stronger negative trend (LSTM: r=-0.23, p=0.290; CTRNN: r=-0.25, p=0.295) than the small positive trend in human data (r=0.02, p=0.87). Additionally, CTRNN showed evidence of a violation of Weber's law for spread 3 (r=0.64, p<0.05; Figure \ref{fig:wf_plots}, third column), whereas no evidence for or against Weber's law was found for LSTM.

    \item \textbf{Architecture similarity:} Both architectures showed similar trends across the BP and Weber ratio analyses. This suggests similarities in temporal encoding between the two types of RL agents. Further neural analysis may help identify the mechanisms underlying these similarities.
\end{itemize}

Hence, the combined analysis of the psychometric curve, BP, and Weber ratio reveals that while the recurrent models with sufficient memory qualitatively resemble human temporal bisection data, there is not enough quantitative evidence to show that the models emergently exhibit human-like timing. As mentioned earlier, a more direct comparison using human experiments in the Chronocooked framework can provide more conclusive evidence in this regard. 

\section{Multi-Anchor Categorization (Temporal Bisection Extension)}
\label{multianchor}

\begin{figure}[ht]
  \centering
  \includegraphics[width=0.3\linewidth]{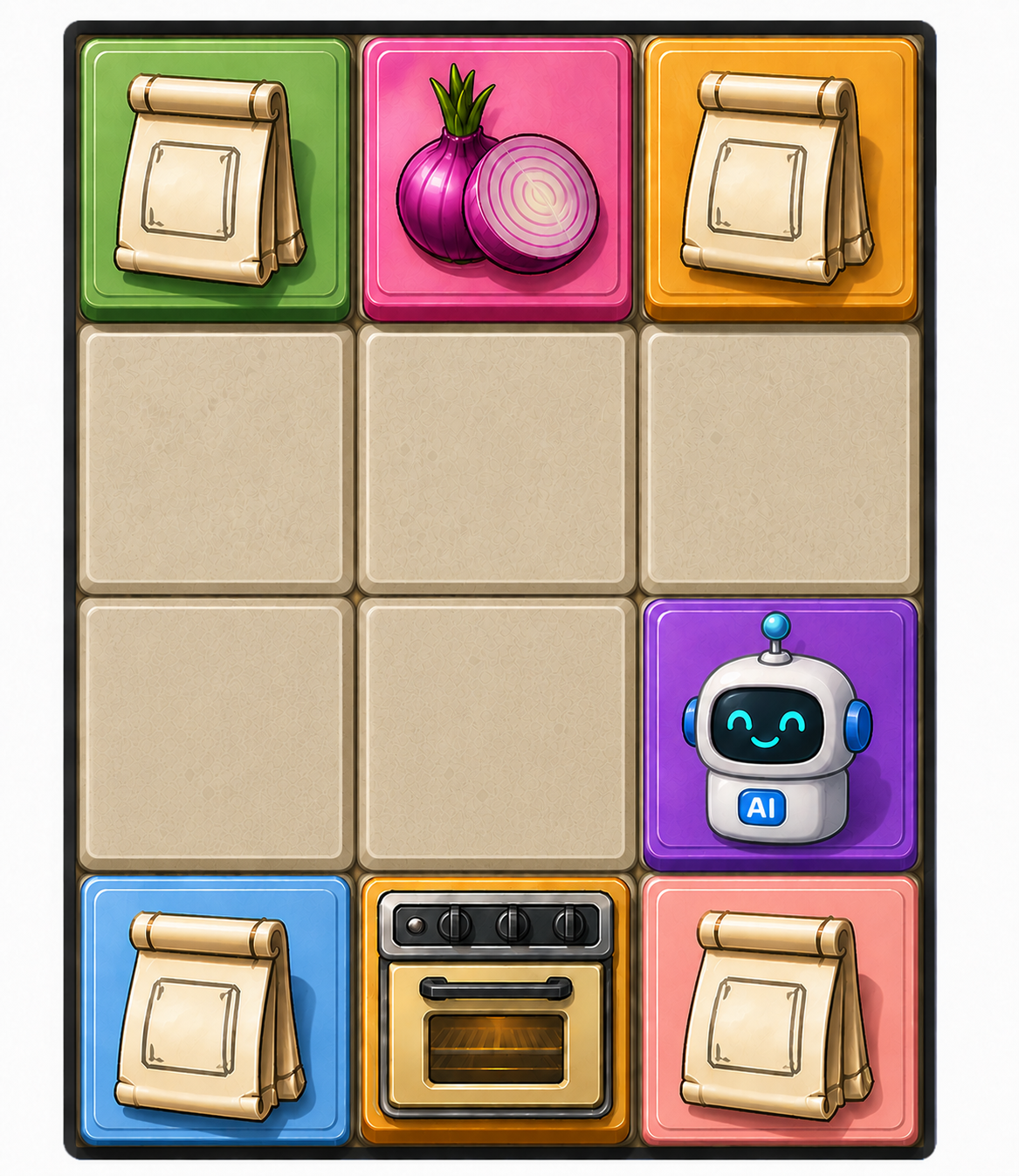}
  \caption{Multi-Anchor categorization Task in Chronocooked. The example shows 4 delivery counters (in different colours) each for a different anchor duration. }
  \label{fig:multianchor_task_grid}
  \Description{The grid layout for Multi-Anchor categorization Task in Chronocooked. It is similar to the temporal bisection task layout except that it contains more delivery counters. The example shows 4 delivery counters (in different colours) each for a different anchor duration.}
\end{figure}

While binary temporal decisions provide a foundation for studying timing behaviour \cite{kopec2010human, levy2015memory}, many real-world situations require more than two temporal categories. Recent work has begun extending timing paradigms beyond binary scenarios \cite{aufschnaiter2022investigating}. This is particularly relevant for HRI, where agents must operate in temporally complex environments alongside humans. For instance, a household robot managing a cooking task must judge whether a dish has been cooking for a "short" (undercooked), "medium" (almost done) or "long" (overcooked) duration, rather than only distinguishing between two extremes.

Chronocooked introduces the multi-anchor categorization task that evaluates whether an RL agent's temporal categorization ability can scale beyond binary categorization  by increasing the number of learned temporal anchors. It is implemented using multiple delivery counters, each associated with a different anchor duration. As in the bisection task, during training, the agent places the onion in the oven, which cooks for one of several target anchor durations selected at random. Once the target anchor duration has elapsed, the oven enters the `ready' state, indicating that the soup can be collected. The agent must identify the elapsed cooking duration and deliver the soup to the corresponding delivery counter. The agent receives a reward of +1 for correct delivery, while incorrect deliveries and all intermediate actions receive a reward of 0. During the test phase (similar to the bisection task), the probe durations include the anchor durations and all durations lying between and beyond the different anchors durations. 


This task is compatible with CL to enable progressively increasing the number of anchor durations. Training can begin with two anchors and two delivery counters (2-anchor setting) as in the temporal bisection task, additional anchors and delivery counters can be introduced once a predefined performance threshold is achieved. In the test phase, similar to the bisection task, probe durations including the different anchor durations and OOD intermediate durations are used to test the agent's temporal categorization ability. 


The multi-anchor categorization task of Chronocooked, evaluates the temporal categorization abilities of RL agents beyond binary decisions. It also enables a systematic study of how an RL agent adapts to increasing number of anchor durations (scalability). Scalability follows the same definition as in Section 6.1, where each new timing requirement corresponds to a new anchor duration. An agent with a scalable timing strategy is expected to reliably handle a larger number of temporal categories (corresponding to the different anchor durations) within a fixed training budget, and to require progressively fewer training steps as new anchors are introduced.


\subsection{Task performance (Evaluation metric)}

\begin{figure}[ht]
  \centering
  \includegraphics[width=\linewidth]{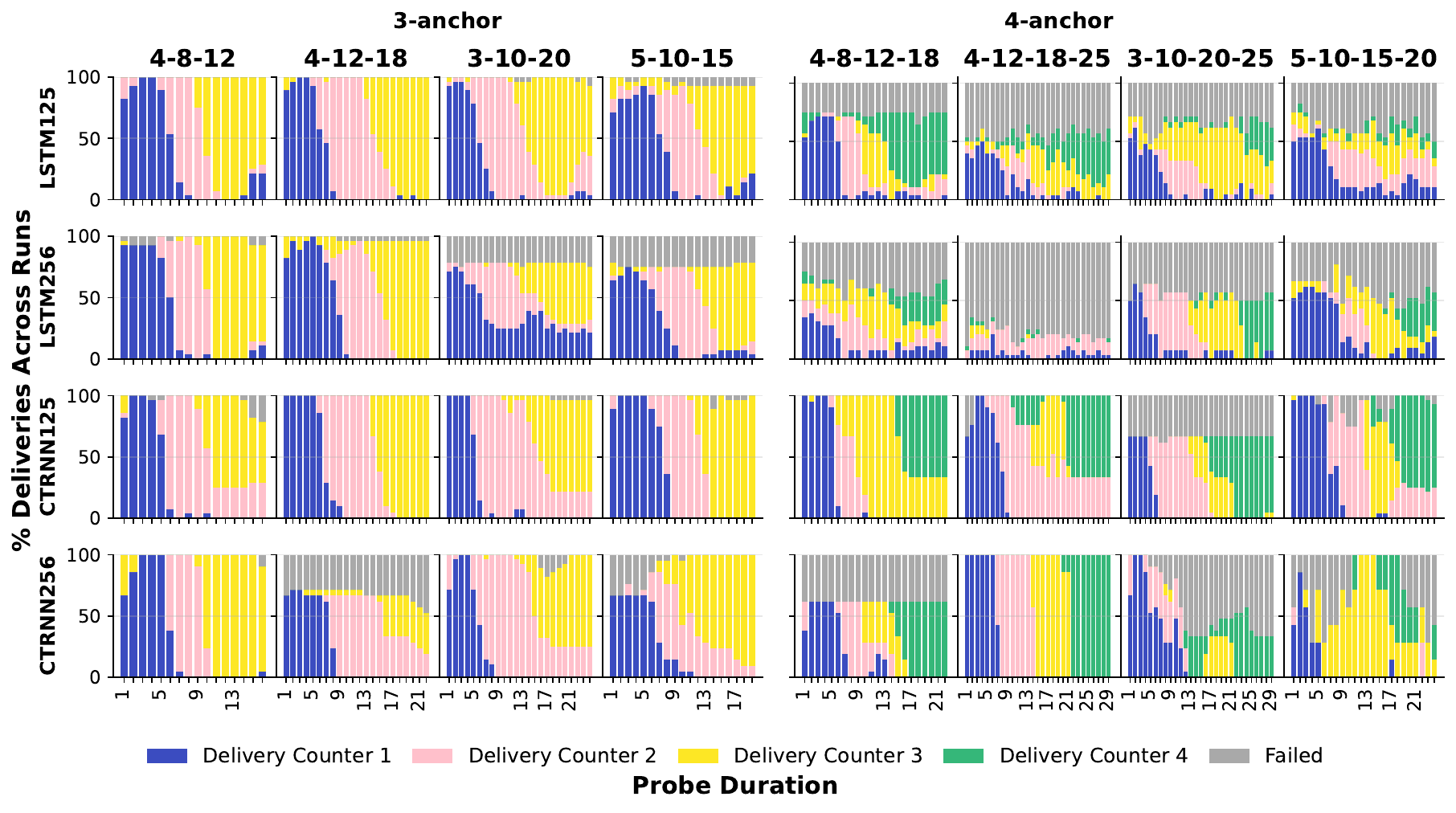}
  \caption{Multi-anchor (Temporal categorization ability): Visualization of 3-anchor (left) and 4-anchor(right) setting. Each subplot shows a stacked bar plot indicating the percentage of deliveries in each delivery counter (indicated by a different colour), aggregated across four runs for a given RL agent (rows) and anchor configuration (columns). Delivery counter 1 corresponds to the smallest anchor duration, delivery counter 2 to the second smallest, and so on. A smooth transition from delivery counter 1 to delivery counter 2 and then 3 and 4 indicates a good performing model. Grey bars denote failed cases, in which the agent did not deliver soup to any counter. }
  \label{fig:multianchor_task_visualization}
  \Description{ The figure shows a clear transition from delivery counter 1 to 2, then 3, and then 4, as probe durations move from one extreme to the other. Importantly, this includes OOD probe durations. CTRNN models exhibit better performance as compared to LSTM models in the 4-anchor setting (last CL stage).}
\end{figure}

For the multi-anchor categorization task, the psychometric curve is replaced with the response distribution across delivery counters. The responses are visualized as a stacked bar plot of percentage of deliveries at each delivery counter for a given probe duration (Figure \ref{fig:multianchor_task_visualization}). A smooth transition between the delivery counters of two consecutive anchors, as the probe duration moves from the smaller to the larger anchor, indicates successful temporal categorization. Unlike the psychometric curve, this visualization is not derived from the psychology literature but is introduced in Chronocooked to evaluate performance on the multi-anchor categorization task.

\begin{figure}[ht]
    \centering
    
    \begin{subfigure}{0.49\linewidth}
        \centering
            \includegraphics[width=\linewidth]{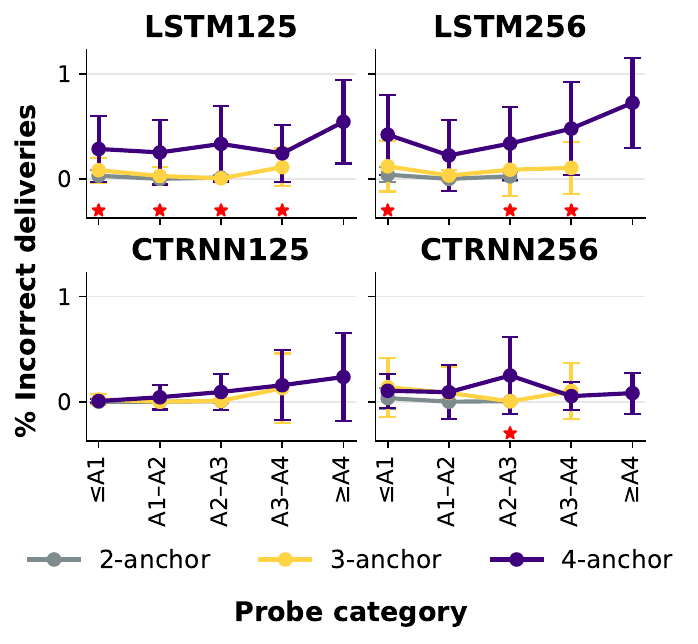}
            \caption{Model-level performance 
            }

    \end{subfigure}
    \hfill
    \begin{subfigure}{0.49\linewidth}
        \centering
            \includegraphics[width=0.8\linewidth]{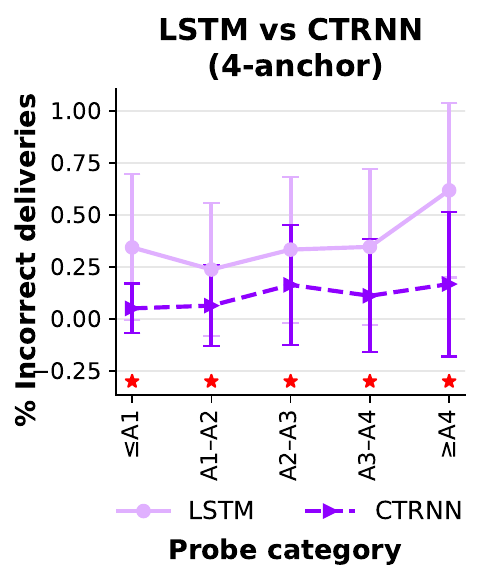}
            \caption{Architecture-level performance}
            
    \end{subfigure}

    \caption{
        Multi-anchor (Task performance): The left panel shows the model-level task performance (as described in Section 8.1) of each recurrent model. Specifically, the plots show the percentage of incorrect deliveries aggregated across runs and main anchor configurations for the different n-anchor settings (indicated by different colours). The red stars on the x-axis denote probe duration categories where \% of incorrect deliveries in the 4-anchor setting was significantly worse than the corresponding 3-anchor setting. The Wilcoxon signed-rank test was used for the comparison. The right panel shows the architecture-level comparison for the 4-anchor setting (last CL stage) task performance. Data across the different main anchor configurations and the two memory sizes within each architecture is combined. The red stars on x-axis in this case denote the probe duration category where LSTM task performance was significantly worse than CTRNN. A Mann-Whitney U test was used for the comparison. For both plots, the error bars show the SD and the p-values are corrected using the Holm–Bonferroni method. Refer Appendix \ref{multianchor_appendix} for details on all significance tests.
    }
    \label{fig:multianchor_task_performance}
    \Description{CTRNN models showed a better task performance than LSTM models both for the model-level and architecture-level analysis.}
\end{figure}

The task performance is quantified using the percentage of incorrect deliveries. We first determine the number of incorrect delivery counters selected by the agent across the different seeds (different starting positions) for each probe duration for a specific CL stage or n-anchor setting. To do this, the probe durations are divided into categories based on the closest anchor durations. For example, for the main anchor configuration 4–8–12–18, probe duration categories are <=4, 4–8, 8–12, 12–18, and >=18. For a given probe duration, a delivery counter is considered incorrect if it does not correspond to one of the anchors of the corresponding probe duration category. For example, for the main anchor configuration 4–8–12–18, probe durations less than or equal to 4 (probe duration category: <= 4) allow delivery counter 1, while probe durations between 4 and 8 (probe duration category: 4-8) allow delivery counters 1 and 2, all other counters are considered incorrect. For each combination of run, model, main anchor configuration, and n-anchor setting, the percentage of incorrect deliveries within each probe duration category are computed by pooling across seeds and the individual probe durations that fall within that category. Failed cases in which soup delivery fails, are excluded to distinguish failures of the soup delivery task from incorrect temporal categorization. For each n-anchor setting and probe duration category, the percentage of incorrect deliveries is averaged across runs, for a given model and main anchor configuration (see Appendix Figure \ref{fig:multianchor_task_performance_supp}). To simplify the analysis, these values are further averaged across the different main anchor configurations, for each model and n-anchor setting (Figure \ref{fig:multianchor_task_performance}, left plot). In this aggregated view, probe duration categories are denoted generically as <=A1 for the smallest anchor, A1–A2 for the two smallest anchors, and so on. This enables comparison of overall task performance in different n-anchor settings for a given model, irrespective of the main anchor configurations.

For a model-level performance, the \% incorrect deliveries in the final CL stage (or n-anchor setting, e.g. 4-anchor setting) is compared with the preceding CL stage (e.g. 3-anchor setting) using a paired parametric or non-parametric test. This comparison is performed separately for each probe duration category. Hence, the resulting p-values must be corrected for multiple tests. The task performance quantification using \% of incorrect deliveries and a paired test of the two last n-anchor settings, enables comparison between the different models. For example, a model where more probe duration categories have  significantly more \% incorrect deliveries in the 4-anchor setting than in the 3-anchor setting can be considered to have poorer overall task performance than a model with fewer such cases.

\subsection{Baseline Results}
The RL agents that reliably performed the temporal bisection task ($LSTM125$, $LSTM256$, $CTRNN125$, $CTRNN256$) were tested for their multi-anchor categorization ability. The main results are summarized below (see Appendix \ref{multianchor_appendix} for an extended version of the result analysis and also for details on task specifications like anchor configurations and CL training).

\subsubsection{Temporal categorization ability} All the four RL agents exhibited temporal categorization ability beyond binary decisions. Figure \ref{fig:multianchor_task_visualization} shows that although there are some cases of non-convergence (shown in grey) in the 4-anchor setting, there is a clear transition from delivery counter 1 to 2, then 3, and then 4, as probe durations move from one extreme to the other. Importantly, this includes OOD probe durations. Hence, the RL agents not only exhibit multi-anchor temporal categorization ability, but are also able to generalize beyond the trained anchor durations.

\subsubsection{Task performance} CTRNN models showed a better overall task performance in comparison to the LSTM models. This suggests that within a fixed training budget (1750k training steps), CTRNN models can better incorporate multiple anchors than LSTM models. The following observations support this claim. 
\begin{itemize}
    \item \textbf{Model-level performance:} Unlike the CTRNN models, the performance of LSTM models ($LSTM125$ and $LSTM256$), was significantly worse (significantly more incorrect deliveries) in the the 4-anchor setting (last CL stage) than the corresponding 3-anchor setting for most probe duration categories (see Figure \ref{fig:multianchor_task_performance} left plot)
    \item \textbf{Architecture-level performance:} An architecture-level comparison of the last CL stage (4-anchor setting) showed that LSTM yielded significantly more incorrect deliveries than CTRNN across all five probe duration categories (see Figure \ref{fig:multianchor_task_performance} right plot). 
    
\end{itemize}

\subsubsection{N-anchor capacity} At an architectural level,  CTRNN showcased a greater capacity to incorporate 4 anchors than LSTM. This is because, within a fixed training budget (1750k training steps), 44\% of CTRNN runs (50\% from CTRNN125 and 38\% from CTRNN256) converged in the 4-anchor setting (i.e. reached a performance threshold of 0.8) before the maximum training budget, compared to only 19\% of LSTM runs (25\% from LSTM125 and 13\% from LSTM256). See Appendix Figure \ref{fig:Scalability_multianchor_supp} for training time steps data. 

\subsubsection{CL efficiency}  There was no significant difference in the CL efficiency of the two architectures in any of the CL stages. Additionally, no evidence of a scalable time representation was found in either architectures. This is because, the CL-efficiency ratio for the LSTM models was significantly more than 1 in both 3-anchor setting (N=30, M=2.22, SD=2.27, Median = 1.55, Wilcoxon W=353.00, p=0.006) and 4-anchor setting (N=6, M=3.33, Median=3.42, SD=2.16, Wilcoxon W=2.643, p=0.023). Whereas, for the CTRNN models it was significantly more than 1 in the 4-anchor setting (N=14, M=3.96, Median=1.60, SD=4.93, Wilcoxon W=75.00, p=0.020) but not in the 3-anchor setting (N=26, M=1.26, SD=1.17, Median =0.85, Wilcoxon W=183.00, p=0.430). Hence, neither architecture showed conclusive evidence of a flexible timing strategy.

Nevertheless, overall the CTRNN architecture exhibited better performance than LSTM in the multi-anchor categorization task, due to a higher model-level and architecture-level task performance and higher convergence rate in the final CL stage (4-anchor setting) within a  fixed training budget.

\section{Asynchronous Multi-Oven (Temporal Bisection Extension)}
\label{asyn_multioven}

\begin{figure}[ht]
  \centering
  \includegraphics[width=0.3\linewidth]{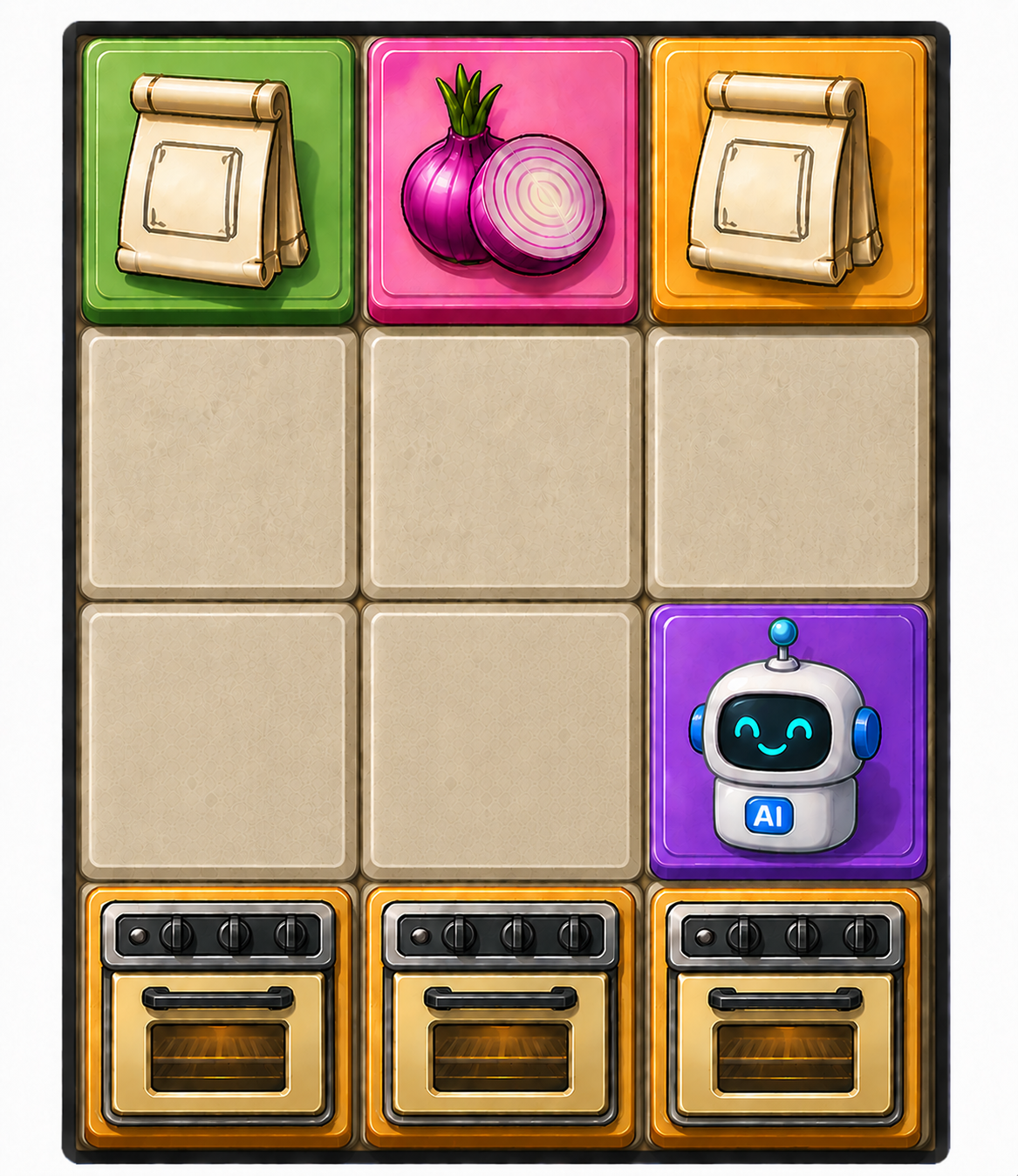}
  \caption{Asynchronous Multi-Oven Bisection Task in Chronocooked. The example shows 3 oven each with a different start time and different TD.  }
  \label{fig:multioven_task_grid}
  \Description{The example shows the grid layout for multi-oven bisection task. The layout is similar to the temporal bisection task, except that there are 3 ovens in the last row instead of just 1 oven.}
\end{figure}

Humans routinely monitor multiple timing events simultaneously, even when they begin at different times. For example, while one dish is cooking in the oven, another may be placed on the stove several minutes later, requiring attention at different points in time. Even in the absence of external cues such as smell, texture or timer, humans can approximately estimate the elapsed time of each cooking process and know when the oven should have finished or when the soup should have reached a particular stage of cooking. There has been quite some work in animal timing studies on concurrent asynchronous timing research \cite{kleinman2016two,leak1995simultaneous,meck1984simultaneous}. However, this area of timing research lacks a single, widely adopted paradigm. 

The asynchronous multi-oven task in Chronocooked provides a simplified simulation of asynchronous concurrent timing task that requires an agent to simultaneously perform multiple temporal bisection tasks with asynchronous start times. This task extends the temporal bisection task by introducing multiple ovens. It retains the same binary decision setting with two anchor durations, $S$ and $L$ and two delivery counters, one for $S$ and one for $L$. The anchors remain same across all ovens. Only one oven (referred as the primary oven) can be activated by the agents' action of placing the onion in it. The other ovens are activated by the environment at random times, constrained between the start of an episode to $6 + 0.5 \times TD_{primary-oven}$, to ensure there is enough time to delivery soup from all ovens before the maximum episode length (200 time steps). Hence, each oven has an independent start time and is randomly assigned either the $S$ or $L$ target duration with 50\% probability. Once an oven reaches its target duration, it enters the `ready' state, and the agent can collect the soup and deliver it to the appropriate delivery counter. The agent receives a reward of 1/N for each correct delivery, where N is the total number of ovens. All intermediate actions and incorrect soup delivery yield a reward of 0. No explicit penalty is assigned for late collection of the soup from the oven. Hence, soups do not need to be delivered immediately after becoming ready, it should only be delivered at the correct delivery counter. Although all soups need to be delivered before the maximum of 200 time steps have been reached to get all the rewards ($N*1/N$). In addition to the asynchronous setting, the task includes a synchronous variant in which all ovens are activated when the agent starts the primary oven. This variant isolates the challenge of performing multiple temporal bisection tasks while removing the additional complexity of tracking asynchronous start times.  It is important to note that, irrespective of the ovens being synchronised (synchronous variant) or asynchronised (asynchronous variant), the target duration ($S$ or $L$) is randomly selected with 50\% probability for each oven separately. 

The task is compatible with CL to study how the RL agents adapt to new timing requirements. The training can start with a single oven (1-oven setting) which is same as the temporal bisection task. Once the agent reaches a predefined performance threshold, training can advance to the second CL stage where an additional oven is introduced (2-oven setting). This process can be repeated until the desired number of ovens is reached. During the test phase, each oven is tested on probe durations that include not only the trained anchor durations but also all intermediate durations between the anchors, following the same protocol as the temporal bisection task.

The task enables evaluation of an agent's ability to handle multiple synchronous or asynchronous timing events. It also enables studying an agent's ability to handle self-triggered versus environment-triggered timing events. Moreover, using the CL training setup, makes it possible to evaluate an agent's ability to scale to progressively added asynchronous (or synchronous) timing tasks. Scalability follows the same definition as in Section 6.1, where each new timing requirement corresponds to a new bisection task. An agent with a scalable timing strategy is expected to reliably perform a larger number of timing tasks within a fixed training budget, and to require progressively fewer training steps as new ovens are introduced. 

Hence, this task brings forward different timing scenarios such as asynchronous vs synchronous multi-timing and self-triggered vs environment-trigged timing events. Future work could investigate each of these scenarios in more detail. For example, studying the underlying neural dynamics could help understand the timing mechanisms used by the RL agents for the different scenarios.


\subsection{Baseline Results}
The RL agents that reliably performed the temporal bisection task ($LSTM125$, $LSTM256$, $CTRNN125$, $CTRNN256$) were tested for their asynchronous multi-oven bisection capabilities. See Appendix \ref{async_multioven_appendix} for more details on CL training, anchor configurations and an extended version of the baseline results including comparison with synchronous multi-oven bisection task and self-triggered vs environment triggered timing events.

\begin{figure}[ht]
    \centering
        \includegraphics[width=0.9\linewidth]{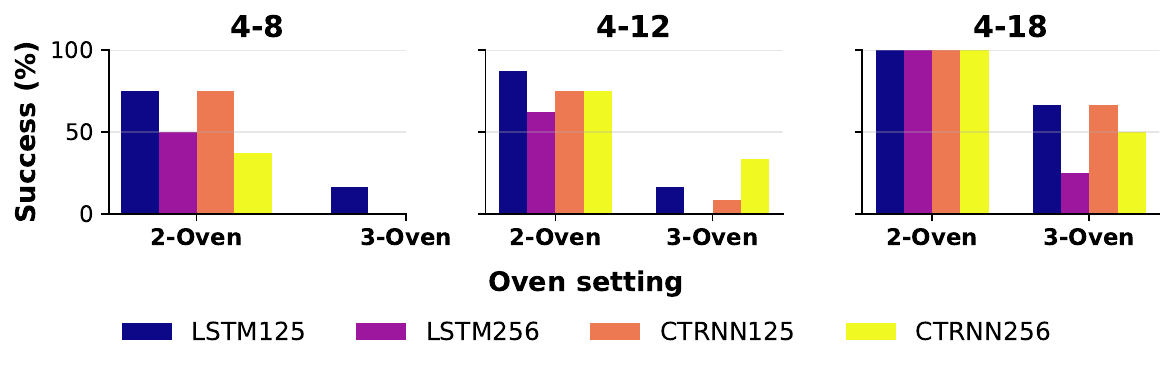}
        \caption{Asynchronous Multi-Oven Bisection (task performance): The plots show the success rate (successful bisection tasks) across all ovens, probe durations (including OOD durations) and runs, for each model (shown in different colours) and anchor configuration (columns), in 2-oven and 3-oven setting (x-axis). While the asynchronous multi-oven task was in general difficult, performance improved as distance between $S$ and $L$ anchor increased.}
        \label{fig:multioven_async_task_performance}

    \Description{In the asynchronous variant, performance improved as distance between $S$ and $L$ increased. In the 3-oven setting, models achieved at most a 50\% success rate for the 4–18 anchor configuration (maximum distance between $S$ and $L$), while performance was lower for all other anchor configurations. }
\end{figure}

\subsubsection{Asynchronous multi-timing ability}
The RL agents were able to perform multiple concurrent bisection tasks that started at different times. However, all agents showed limited capabilities, with performance declining substantially beyond the 2-oven setting (Figure \ref{fig:multioven_async_task_performance}). Additionally, although the RL agents showed the ability to keep track of environment-triggered events (the asynchronous ovens), such events were more difficult to track than events triggered by the agent's own actions (the primary oven). See Appendix Figures \ref{fig:multioven_asyn_psychometric}, \ref{fig:multioven_psychometric_4_12} and \ref{fig:multioven_psychometric_4_8} for more details. 

\begin{figure}[ht]
  \centering
  \includegraphics[width=0.9\linewidth]{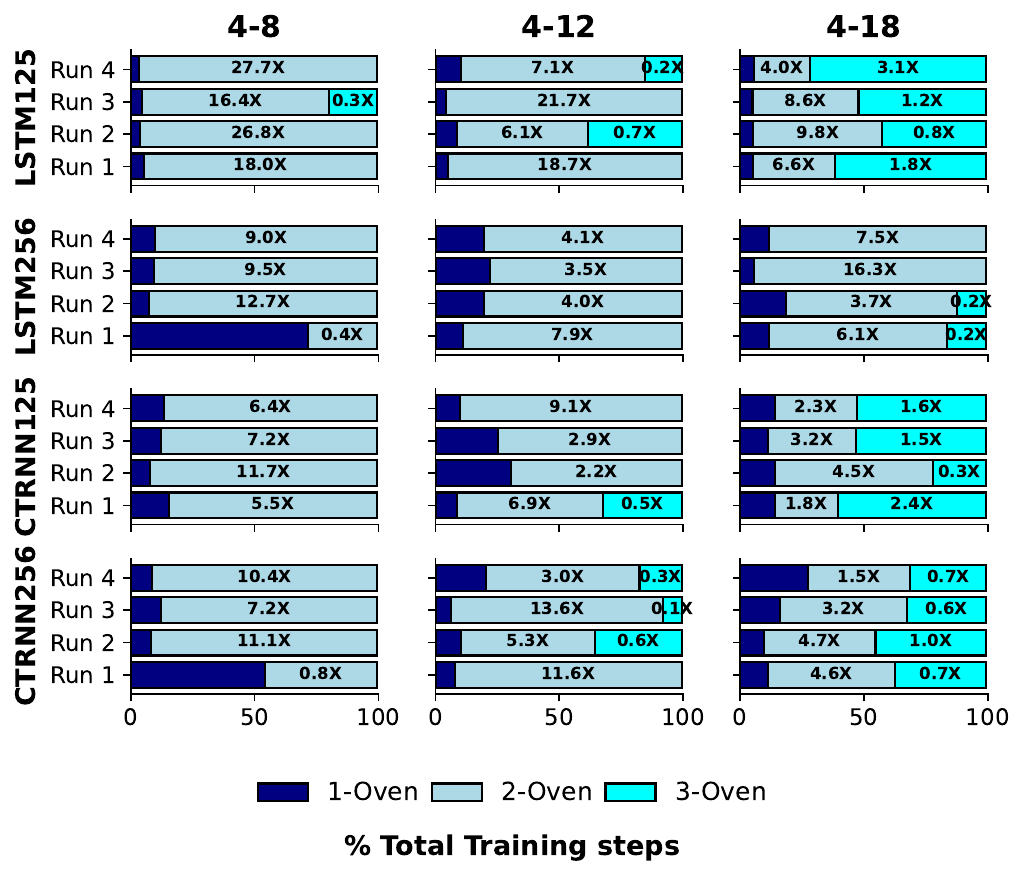}
  \caption{Asynchronous multi-oven bisection (Scalability): The plot shows the training time steps required for each CL stage (n-oven setting) shown in different colours for each model (rows) and anchor configurations (columns). The numbers in the bars represent the corresponding CL-efficiency ratios. All models needed significantly more than 1x the time steps in 1-oven setting to converge in the 2-oven setting (CL-efficiency ratio >1). }
  \label{fig:scalability_multioven_async_cl_supp}
  \Description{All models reach the 3-oven setting only for anchor configuration 4-18 (largest spread). Only one run from $LSTM125$ reaches 3-oven setting for anchor configuration 4-8 (smallest spread). All models need significantly more than 1x time steps of previous CL stage to complete the current CL stage.}
\end{figure} 

\subsubsection{Task performance}

Task performance is quantified using percentage of successful bisection tasks across the all ovens and runs. A successful bisection task is determined using the same techniques from temporal bisection (section 7.2). It is important to note that the evaluation also includes OOD probe durations. The following main observation were made about the task performance of the different models. 
\begin{itemize} 

    \item At the architecture level, both architectures showed similar performance. For the 2-oven setting the average performance of CTRNN and LSTM models across all ovens, runs and anchor configurations was 79\%. In the 3-oven setting, it was 26\% for CTRNN and 25\% for LSTM. The results come with a caveat that the $LSTM256$, which has the maximum number of parameters showed consistently poor performance overall. Further investigation is needed to identify the cause of this observation (See Appendix \ref{async_multioven_appendix} for more details).
    \item The distance between the two anchors affected task difficulty. As the distance between $S$ and $L$ increased the performance of all models improved (Figure \ref{fig:multioven_async_task_performance}).

\end{itemize}

\subsubsection{N-oven capacity} At the architectural level, none of the runs from either architectures converged in the 3-oven setting. In contrast, for the 2-oven setting, which corresponded to the first introduction of an asynchronous oven, 50\% of CTRNN runs and 38\% of LSTM runs converged. Thus, while both architectures showed the ability to incorporate one additional asynchronous oven, this ability was slightly greater for CTRNN.

\subsubsection{CL efficiency} An architecture-level analysis of the CL efficiency in the asynchronous variant shows that CTRNN exhibited slower learning than LSTM in the first CL stage (which corresponded to learning all the sub-tasks in addition to the timing task). This is because, it needed significantly more training steps than LSTM in the first CL stage (CTRNN: N=24, M=15.37, Median=12.13, SD=10.40; LSTM: N=24, M=11.87, Median=7.95, SD=13.86; Mann-Whitney U=415.00, p=0.004). In the next CL stage that corresponded to the introduction of an additional asynchronous bisection task, there was no significant difference in the CL-efficiency ratio of the two architectures (CTRNN: M=4.56, SD=3.24, Median=3.86, n=12; LSTM:M=6.51, SD=2.08, Median=6.37, n=8 ;Mann-Whitney U=21.00, p=0.984). Additionally, the CL-efficiency ratio in this stage for both architectures was significantly more than 1 (CTRNN: Wilcoxon W=78.00, p<0.001; LSTM: t=7.50, p<0.001). Hence no evidence of scalability  was found in either architectures. 

Hence, although the CTRNN models needed significantly more training steps than LSTM in the first CL stage, in the later CL stages it exhibited a comparable CL-efficiency ratio, task performance and slightly better convergence rate than the LSTM models within a fixed training budget (1750k training steps). This shows that the CTRNN models are able to catch up with the LSTM models in the later CL stage in spite of slow initial learning. This could hint towards the emergence of a flexible timing strategy not caught by the simple definition of CL efficiency established in this study.

\section{ (Future Work) Human experiments}
One of the goals of Chronocooked is to analyse human-like timing in artificial agents. To achieve this, human experiments under comparable conditions are needed for direct comparison. We envision extending the tasks to online versions that can be used with human participants. In this study, we limit our analysis to well-established findings in human timing research from the psychology literature and compare it to the artificial agent's timing behaviour. An important next step is to extend the framework for human studies and enable direct comparisons between artificial agents and human participants.


\section{Discussion}

In this study, we introduced Chronocooked, an RL environment for studying interval timing in RL agents. The environment includes tasks inspired from the psychology literature, designed such that time is the only missing piece of information. These tasks enabled evaluation of whether RL agents could achieve optimal performance by inferring time in different settings such as time production and categorization. They also allowed a controlled analysis of the resulting temporal characteristics such as human-like timing, implicit or embodied timing, OOD generalization, and asynchronous timing. Moreover, through CL-compatible extensions, the tasks supported a systematic study of how agents adapt to new timing requirements.

All models, including the non-recurrent CNN, successfully performed both core timing tasks (FI timing and temporal bisection), although with varying capacity. For both core tasks, higher memory recurrent models generally appeared to exhibit better performance. In the FI task, CNN relied on embodied timing strategies, while higher memory LSTM and CTRNN models exhibited implicit time-keeping. In the bisection task, CNN and low-memory recurrent models showed limited capacity, while higher memory recurrent models were more robust and exhibited a s-shaped psychometric response curve. Nevertheless, all models generalized successfully to OOD probe durations. Across the different core task extensions, models from both recurrent architectures were able to successfully perform the multi-timer, multi-anchor categorization, and asynchronous multi-oven tasks, again with varying capability.



Having established that most models solve the timing tasks, we next evaluated human-like timing. Such timing ability is important in the context of HRI. We hypothesize that models exhibiting more human-like timing would better predict human responses in time-dependent tasks, improving the overall interaction experience. None of the models showed emergent human-like timing properties. This was expected since the models were not optimized for this. Many existing timing models from psychology and neuroscience literature also have such properties explicitly engineered in them, rather than arising emergently.  This is further complicated by inconsistencies in the human timing literature regarding specific timing biases, although we attempted to consolidate the most consistently reported ones for our metrics. Additionally, most of the metrics can only be tested for non-conformity (p>0.05) and not for conformity. For example, rejecting the null hypothesis that the slope of Weber ratio linearly increases with AM, does not imply a constant one, and is not equivalent to confirming human-like timing \cite{penney2018duration}. As noted earlier, direct human experiments using Chronocooked as a common framework would help make these metrics more concrete. Interestingly, for both core tasks (FI and bisection), the two recurrent architectures exhibited similar trends for the different human-like timing metrics. This indicates an underlying similarity in temporal processing of the two types of artificial agents.  


A second metric evaluated once a timing task is learned is scalability, that is, how well an RL agent adapts to new timing requirements. At an architectural level, CTRNN showed better scalability than LSTM. Despite slower initial learning across all tasks, it consistently caught up and showed better performance at the final CL stage. This pattern held across all the three task extensions tested using CL. Note however, under our definition of CL efficiency (requiring a CL-efficiency ratio <= 1), even CTRNN did not fully satisfy this criterion. Further investigation into the underlying neural mechanisms is needed to examine how CTRNN's temporal representations differ from LSTM's across stages in a way that produces this consistent pattern. A neural analysis can also help to confirm whether flexible temporal representation are involved while incorporating new timing requirements. 


Studying neural similarities between artificial and biological timing is another promising future application of this environment. A substantial body of research compares artificial and biological neural networks \cite{artificial_yang_2021, interpretability_kar_2022, convolutional_lindsay_2021, task_yang_2019}. Since RL agents in Chronocooked emergently develop distinct timing strategies, this environment offers an opportunity to examine similarity to biological systems at the level of neural mechanisms, beyond behavioural comparisons alone. Since RL agents offer a more structured and less costly experimental setting than traditional human or animal studies, it could help the timing research community to generate new hypotheses about biological timing mechanisms. Central to this line of work is a broader question: when two systems, one biological and one artificial, independently solve the same task using similar underlying processes, this raises the question of whether that underlying mechanism which happens to converge across systems is necessary to solve the task \cite{task_yang_2019, ludvig2011primer,artificial_kanwisher_2023}.


This study has several limitations. First, our human-like timing metrics must be complemented with direct human experiments using Chronocooked as a common framework. This would help address the issues around the general lack of consensus on timing biases in the timing literature, as discussed above. Second, our timing is discretized into time steps, whereas human, animal, and real-world HRI timing is continuous, spanning sub-second to hour-long durations; we do not address this distinction here, focusing instead on whether an agent can solve a timing task defined in discrete time step units. Third, while Chronocooked aims to cover the important interval timing tasks, it is not an exhaustive list of interval timing tasks in psychology literature. Additionally, the task extensions introduced in this work do not correspond to a single, well-established paradigm in psychology. Rather, we adapted some concepts in the literature to construct extensions of the basic tasks. Nevertheless, human data from these extended tasks would be valuable for studying human capacity and efficiency on similar problems. Fourth, our baseline experiments focus on model-free RL, specifically the PPO algorithm with a shared actor-critic architecture. Future work can explore the same timing capabilities using different types of RL algorithms and model architectures. Fifth, a neural analysis of the different agents is missing and would be a valuable addition to the results in these studies. Finally, although this study focuses on HRI, it is a preliminary investigation about the temporal processing capabilities of RL agents. Real-world HRI introduces additional temporal constraints, such as processing speed and hardware limitations, which are beyond the scope of this work.

Nevertheless, temporal processing is an important dimension of many artificial and real-world tasks, alongside visual, auditory, and textual processing \cite{marshall2015everywhere}. While substantial progress has been made in these other domains, temporal processing in artificial agents remains comparatively under-explored. This study takes a small step towards exploring a specific aspect of temporal processing, namely interval timing in RL agents. The analysis conducted using the Chronocooked environment addresses two questions: (1) can RL agents process temporal information in the absence of other environmental cues, and (2) What timing abilities (specifically, human-like timing and scalability) emerge in RL agents while learning optimal polices for time-dependent tasks?
We hypothesize that studying interval timing in RL agents could provide an additional dimension for improving performance and efficiency in HRI tasks. More broadly, this work aims to connect largely separate areas of research, namely, reinforcement learning, HRI and timing research in psychology and neuroscience.

\section{Acknowledgements}
This work was supported by European Union’s Horizon 2020 FET research program under grant agreement
No. 964464 (ChronoPilot).

\bibliography{references}

@article{block2014time,
  title={Time perception, attention, and memory: A selective review},
  author={Block, Richard A and Gruber, Ronald P},
  journal={Acta psychologica},
  volume={149},
  pages={129--133},
  year={2014},
  publisher={Elsevier}
}

@article{thones2019standard,
  title={A standard conceptual framework for the study of subjective time},
  author={Th{\"o}nes, Sven and Stocker, Kurt},
  journal={Consciousness and cognition},
  volume={71},
  pages={114--122},
  year={2019},
  publisher={Elsevier}
}

@inproceedings{badia2020agent57,
  title={Agent57: Outperforming the atari human benchmark},
  author={Badia, Adri{\`a} Puigdom{\`e}nech and Piot, Bilal and Kapturowski, Steven and Sprechmann, Pablo and Vitvitskyi, Alex and Guo, Zhaohan Daniel and Blundell, Charles},
  booktitle={International conference on machine learning},
  pages={507--517},
  year={2020},
  organization={PMLR}
}

@article{ariely2001timely,
  title={A timely account of the role of duration in decision making},
  author={Ariely, Dan and Zakay, Dan},
  journal={Acta psychologica},
  volume={108},
  number={2},
  pages={187--207},
  year={2001},
  publisher={Elsevier}
}

@incollection{zimbardo2014putting,
  title={Putting time in perspective: A valid, reliable individual-differences metric},
  author={Zimbardo, Philip G and Boyd, John N},
  booktitle={Time perspective theory; review, research and application: Essays in honor of Philip G. Zimbardo},
  pages={17--55},
  year={2014},
  publisher={Springer}
}

@inproceedings{kabir2025deep,
  title={Deep reinforcement learning with time-scale invariant memory},
  author={Kabir, Md Rysul and Mochizuki-Freeman, James and Tiganj, Zoran},
  booktitle={Proceedings of the AAAI Conference on Artificial Intelligence},
  volume={39},
  pages={1345--1354},
  year={2025}
}

@article{deverett2019interval,
  title={Interval timing in deep reinforcement learning agents},
  author={Deverett, Ben and Faulkner, Ryan and Fortunato, Meire and Wayne, Gregory and Leibo, Joel Z},
  journal={Advances in Neural Information Processing Systems},
  volume={32},
  year={2019}
}

@inproceedings{pednekar2025emergent,
  title={Emergent time-keeping mechanisms in a deep reinforcement learning agent performing an interval timing task},
  author={Pednekar, Amrapali and Garrido, Alvaro and Simoens, Pieter and Khaluf, Yara},
  booktitle={Artificial Life Conference Proceedings 37},
  volume={2025},
  pages={51},
  year={2025},
  organization={MIT Press One Rogers Street, Cambridge, MA 02142-1209, USA journals-info~…}
}

@article{roseboom2019activity,
  title={Activity in perceptual classification networks as a basis for human subjective time perception},
  author={Roseboom, Warrick and Fountas, Zafeirios and Nikiforou, Kyriacos and Bhowmik, David and Shanahan, Murray and Seth, Anil K},
  journal={Nature communications},
  volume={10},
  number={1},
  pages={267},
  year={2019},
  publisher={Nature Publishing Group UK London}
}

@article{lin2023temporal,
  title={Temporal encoding in deep reinforcement learning agents},
  author={Lin, Dongyan and Huang, Ann Zixiang and Richards, Blake Aaron},
  journal={Scientific Reports},
  volume={13},
  number={1},
  pages={22335},
  year={2023},
  publisher={Nature Publishing Group UK London}
}

@article{goudar2018encoding,
  title={Encoding sensory and motor patterns as time-invariant trajectories in recurrent neural networks},
  author={Goudar, Vishwa and Buonomano, Dean V},
  journal={Elife},
  volume={7},
  pages={e31134},
  year={2018},
  publisher={eLife Sciences Publications, Ltd}
}

@inproceedings{labash2023emergence,
  title={Emergence of adaptive circadian rhythms in deep reinforcement learning},
  author={Labash, Aqeel and Stelzer, Florian and Majoral, Daniel and Zafra, Raul Vicente},
  booktitle={International Conference on Machine Learning},
  pages={18153--18170},
  year={2023},
  organization={PMLR}
}

@article{wang2022neural,
  title={A neural network model for timing control with reinforcement},
  author={Wang, Jing and El-Jayyousi, Yousuf and Ozden, Ilker},
  journal={Frontiers in Computational Neuroscience},
  volume={16},
  pages={918031},
  year={2022},
  publisher={Frontiers Media SA}
}

@inproceedings{cominelli2019influence,
  title={The Influence of Emotions on Time Perception in a Cognitive System for Social Robotics},
  author={Cominelli, Lorenzo and Garofalo, Roberto and De Rossi, Danilo},
  booktitle={2019 28th IEEE International Conference on Robot and Human Interactive Communication (RO-MAN)},
  pages={1--6},
  year={2019},
  organization={IEEE}
}

@article{contextdependent_mante_2013,
	title = {Context-dependent computation by recurrent dynamics in prefrontal cortex},
	doi = {10.1038/NATURE12742},
	author = {Mante, Valerio and Sussillo, David and Shenoy, Krishna V. and Newsome, William T.},
	journal = {Nature},
	year = {2013},
	pubmedId = {https://pubmed.ncbi.nlm.nih.gov/24201281},
	researchRabbitId = {7ffec82b-433c-460a-bdaa-9761bca2038d}
}

@article{latent_cimolino_2021,
	title = {Latent Time-Adaptive Drift-Diffusion Model.},
	author = {Cimolino, Gabriele and Rivest, François},
	journal = {arXiv: Learning},
	year = {2021},
	researchRabbitId = {f6c670fe-663c-47b5-9a4d-d55c7b5302b9}
}

@inproceedings{maniadakis2012experiencing,
  title={Experiencing and processing time with neural networks},
  author={Maniadakis, Michail and Trahanias, Panos},
  booktitle={Proc. 4th Int. Conf. on Advanced Cogn. Tech. and App},
  year={2012}
}

@inproceedings{maniadakis2015artificial,
  title={Artificial agents perceiving and processing time},
  author={Maniadakis, Michail and Trahanias, Panos},
  booktitle={2015 International Joint Conference on Neural Networks (IJCNN)},
  pages={1--8},
  year={2015},
  organization={IEEE}
}

@article{howlong_maniadakis_2016,
	title = {When and How-Long: A Unified Approach for Time Perception},
	doi = {10.3389/FPSYG.2016.00466},
	author = {Maniadakis, Michail and Trahanias, Panos},
	journal = {Front. Psychol.},
	year = {2016},
	pubmedId = {https://pubmed.ncbi.nlm.nih.gov/27065930},
	researchRabbitId = {c8c4d64e-a340-4abd-9f0a-184177f00ff3}
}

@article{teaching_loureno_2020,
	title = {Teaching Robots to Perceive Time: A Twofold Learning Approach},
	doi = {10.1109/ICDL-EPIROB48136.2020.9278033},
	author = {Lourenço, Inês and Ventura, Rodrigo and Wahlberg, Bo},
	journal = {Joint IEEE International Conference on Development and Learning and on Epigenetic Robotics},
	year = {2020},
	researchRabbitId = {b8a1cc96-983a-49f4-821b-5d6051edb497}
}

@article{predictive_fountas_2022,
	title = {A Predictive Processing Model of Episodic Memory and Time Perception},
	doi = {10.1162/NECO_A_01514},
	author = {Fountas, Zafeirios and Sylaidi, Anastasia and Nikiforou, Kyriacos and Seth, Anil K. and Shanahan, Murray and Roseboom, Warrick and Fountas, Zafeirios and Sylaidi, Anastasia and Nikiforou, Kyriacos and Seth, Anil K. and Shanahan, Murray and Roseboom, Warrick},
	journal = {Neural Computation},
	year = {2022},
	pubmedId = {https://pubmed.ncbi.nlm.nih.gov/35671462},
	researchRabbitId = {1f0a7d65-0d63-4756-86ee-e59a44780c89}
}

@article{safaie2020turning,
  title={Turning the body into a clock: Accurate timing is facilitated by simple stereotyped interactions with the environment},
  author={Safaie, Mostafa and Jurado-Parras, Maria-Teresa and Sarno, Stefania and Louis, Jordane and Karoutchi, Corane and Petit, Ludovic F and Pasquet, Matthieu O and Eloy, Christophe and Robbe, David},
  journal={Proceedings of the National Academy of Sciences},
  volume={117},
  number={23},
  pages={13084--13093},
  year={2020},
  publisher={National Academy of Sciences}
}

@article{timeorder_komosiski_2015,
	title = {Time-order error and scalar variance in a computational model of human timing: simulations and predictions},
	doi = {10.1186/S40469-015-0002-0},
	author = {Komosiński, Maciej and Kupś, Adam},
	journal = {Computational cognitive science},
	year = {2015},
	researchRabbitId = {cd869605-5c01-42a0-ae00-e44ea5b035c4}
}

@article{hass2014neurocomputational,
  title={Neurocomputational models of time perception},
  author={Hass, Joachim and Durstewitz, Daniel},
  journal={Neurobiology of interval timing},
  pages={49--71},
  year={2014},
  publisher={Springer}
}

@article{hass2016time,
  title={Time at the center, or time at the side? Assessing current models of time perception},
  author={Hass, Joachim and Durstewitz, Daniel},
  journal={Current Opinion in Behavioral Sciences},
  volume={8},
  pages={238--244},
  year={2016},
  publisher={Elsevier}
}

@article{basgol2021time,
  title={Time perception: A review on psychological, computational, and robotic models},
  author={Basgol, Hamit and Ayhan, Inci and Ugur, Emre},
  journal={IEEE Transactions on Cognitive and Developmental Systems},
  volume={14},
  number={2},
  pages={301--315},
  year={2021},
  publisher={IEEE}
}

@article{jazayeri2010temporal,
  title={Temporal context calibrates interval timing},
  author={Jazayeri, Mehrdad and Shadlen, Michael N},
  journal={Nature neuroscience},
  volume={13},
  number={8},
  pages={1020--1026},
  year={2010},
  publisher={Nature Publishing Group US New York}
}

@article{gers2002learning,
  title={Learning precise timing with LSTM recurrent networks},
  author={Gers, Felix A and Schraudolph, Nicol N and Schmidhuber, J{\"u}rgen},
  journal={Journal of machine learning research},
  volume={3},
  number={Aug},
  pages={115--143},
  year={2002}
}

@article{petter2018integrating,
  title={Integrating models of interval timing and reinforcement learning},
  author={Petter, Elijah A and Gershman, Samuel J and Meck, Warren H},
  journal={Trends in cognitive sciences},
  volume={22},
  number={10},
  pages={911--922},
  year={2018},
  publisher={Elsevier}
}

@article{leibo2018psychlab,
  title={Psychlab: a psychology laboratory for deep reinforcement learning agents},
  author={Leibo, Joel Z and d'Autume, Cyprien de Masson and Zoran, Daniel and Amos, David and Beattie, Charles and Anderson, Keith and Casta{\~n}eda, Antonio Garc{\'\i}a and Sanchez, Manuel and Green, Simon and Gruslys, Audrunas and others},
  journal={arXiv preprint arXiv:1801.08116},
  year={2018}
}

@article{gershman2014time,
  title={Time representation in reinforcement learning models of the basal ganglia},
  author={Gershman, Samuel J and Moustafa, Ahmed A and Ludvig, Elliot A},
  journal={Frontiers in computational neuroscience},
  volume={7},
  pages={194},
  year={2014},
  publisher={Frontiers Media SA}
}

@article{daw2006representation,
  title={Representation and timing in theories of the dopamine system},
  author={Daw, Nathaniel D and Courville, Aaron C and Touretzky, David S},
  journal={Neural computation},
  volume={18},
  number={7},
  pages={1637--1677},
  year={2006},
  publisher={MIT Press}
}

@article{karmarkar2007timing,
  title={Timing in the absence of clocks: encoding time in neural network states},
  author={Karmarkar, Uma R and Buonomano, Dean V},
  journal={Neuron},
  volume={53},
  number={3},
  pages={427--438},
  year={2007},
  publisher={Elsevier}
}

@article{carroll2019utility,
  title={On the utility of learning about humans for human-ai coordination},
  author={Carroll, Micah and Shah, Rohin and Ho, Mark K and Griffiths, Tom and Seshia, Sanjit and Abbeel, Pieter and Dragan, Anca},
  journal={Advances in neural information processing systems},
  volume={32},
  year={2019}
}

@article{allan1991human,
  title={Human bisection at the geometric mean},
  author={Allan, Lorraine G and Gibbon, John},
  journal={Learning and motivation},
  volume={22},
  number={1-2},
  pages={39--58},
  year={1991},
  publisher={Elsevier}
}

@article{allan2002location,
  title={The location and interpretation of the bisection point},
  author={Allan, Lorraine G},
  journal={The Quarterly Journal of Experimental Psychology: Section B},
  volume={55},
  number={1},
  pages={43--60},
  year={2002},
  publisher={Taylor \& Francis}
}

@article{gibbon1981form,
  title={On the form and location of the psychometric bisection function for time},
  author={Gibbon, John},
  journal={Journal of Mathematical Psychology},
  volume={24},
  number={1},
  pages={58--87},
  year={1981},
  publisher={Elsevier}
}

@incollection{penney2018duration,
  title={Duration bisection: a user’s guide},
  author={Penney, Trevor B and Cheng, Xiaoqin},
  booktitle={Timing and time perception: Procedures, measures, \& applications},
  pages={98--127},
  year={2018},
  publisher={Brill}
}

@article{wearden1988interval,
  title={Interval production as an analogue of the peak procedure: Evidence for similarity of human and animal timing processes},
  author={Wearden, John H and McShane, Bairbre},
  journal={The Quarterly Journal of Experimental Psychology},
  volume={40},
  number={4},
  pages={363--375},
  year={1988},
  publisher={Taylor \& Francis}
}

@article{wearden1991human,
  title={Human performance on an analogue of an interval bisection task},
  author={Wearden, JH},
  journal={The Quarterly Journal of Experimental Psychology Section B},
  volume={43},
  number={1b},
  pages={59--81},
  year={1991},
  publisher={SAGE Publications Sage UK: London, England}
}

@incollection{freestone2018temporal,
  title={Temporal Decision-making: Common Procedures and Contemporary Approaches},
  author={Freestone, David and Balc{\i}, Fuat},
  booktitle={Timing and Time Perception: Procedures, Measures, \& Applications},
  pages={128--148},
  year={2018},
  publisher={Brill}
}

@article{watson2011fixed,
  title={Fixed interval schedule},
  author={Watson, Steuart T and Griffes, Christine},
  journal={Encyclopedia of Child Behavior and Development},
  pages={659--660},
  year={2011},
  publisher={Springer US}
}

@article{ferster1957schedules,
  title={Schedules of reinforcement.},
  author={Ferster, Charles B and Skinner, Burrhus Frederic},
  year={1957},
  publisher={Appleton-Century-Crofts}
}

@article{lejeune1991comparative,
  title={The comparative psychology of fixed-interval responding: Some quantitative analyses},
  author={Lejeune, Helga and Wearden, JH},
  journal={Learning and Motivation},
  volume={22},
  number={1-2},
  pages={84--111},
  year={1991},
  publisher={Elsevier}
}

@article{hoffman2019evaluating,
  title={Evaluating fluency in human--robot collaboration},
  author={Hoffman, Guy},
  journal={IEEE Transactions on Human-Machine Systems},
  volume={49},
  number={3},
  pages={209--218},
  year={2019},
  publisher={IEEE}
}

@article{ajoudani2018progress,
  title={Progress and prospects of the human--robot collaboration},
  author={Ajoudani, Arash and Zanchettin, Andrea Maria and Ivaldi, Serena and Albu-Sch{\"a}ffer, Alin and Kosuge, Kazuhiro and Khatib, Oussama},
  journal={Autonomous robots},
  volume={42},
  number={5},
  pages={957--975},
  year={2018},
  publisher={Springer}
}

@inproceedings{nikolaidis2015efficient,
  title={Efficient model learning from joint-action demonstrations for human-robot collaborative tasks},
  author={Nikolaidis, Stefanos and Ramakrishnan, Ramya and Gu, Keren and Shah, Julie},
  booktitle={Proceedings of the tenth annual ACM/IEEE international conference on human-robot interaction},
  pages={189--196},
  year={2015}
}

@article{vinyals2019grandmaster,
  title={Grandmaster level in StarCraft II using multi-agent reinforcement learning},
  author={Vinyals, Oriol and Babuschkin, Igor and Czarnecki, Wojciech M and Mathieu, Micha{\"e}l and Dudzik, Andrew and Chung, Junyoung and Choi, David H and Powell, Richard and Ewalds, Timo and Georgiev, Petko and others},
  journal={nature},
  volume={575},
  number={7782},
  pages={350--354},
  year={2019},
  publisher={Nature Publishing Group UK London}
}

@article{jaderberg2019human,
  title={Human-level performance in 3D multiplayer games with population-based reinforcement learning},
  author={Jaderberg, Max and Czarnecki, Wojciech M and Dunning, Iain and Marris, Luke and Lever, Guy and Castaneda, Antonio Garcia and Beattie, Charles and Rabinowitz, Neil C and Morcos, Ari S and Ruderman, Avraham and others},
  journal={Science},
  volume={364},
  number={6443},
  pages={859--865},
  year={2019},
  publisher={American Association for the Advancement of Science}
}

@article{matell2000neuropsychological,
  title={Neuropsychological mechanisms of interval timing behavior},
  author={Matell, Matthew S and Meck, Warren H},
  journal={Bioessays},
  volume={22},
  number={1},
  pages={94--103},
  year={2000},
  publisher={Wiley Online Library}
}

@article{merchant2014introduction,
  title={Introduction to the neurobiology of interval timing},
  author={Merchant, Hugo and De Lafuente, Victor},
  journal={Neurobiology of interval timing},
  pages={1--13},
  year={2014},
  publisher={Springer}
}

@article{buhusi2005makes,
  title={What makes us tick? Functional and neural mechanisms of interval timing},
  author={Buhusi, Catalin V and Meck, Warren H},
  journal={Nature reviews neuroscience},
  volume={6},
  number={10},
  pages={755--765},
  year={2005},
  publisher={Nature Publishing Group UK London}
}

@article{wearden2008scalar,
  title={Scalar properties in human timing: Conformity and violations},
  author={Wearden, John H and Lejeune, Helga},
  journal={The Quarterly Journal of Experimental Psychology},
  volume={61},
  number={4},
  pages={569--587},
  year={2008},
  publisher={Taylor \& Francis}
}

@article{lejeune2006scalar,
  title={Scalar properties in animal timing: Conformity and violations},
  author={Lejeune, Helga and Wearden, JH},
  journal={The Quarterly journal of experimental psychology},
  volume={59},
  number={11},
  pages={1875--1908},
  year={2006},
  publisher={Taylor \& Francis}
}

@article{gibbon1977scalar,
  title={Scalar expectancy theory and Weber's law in animal timing.},
  author={Gibbon, John},
  journal={Psychological review},
  volume={84},
  number={3},
  pages={279},
  year={1977},
  publisher={American Psychological Association}
}

@article{laming2009weber,
  title={Weber’s law},
  author={Laming, Donald},
  journal={Inside Psychology: A science over 50 years},
  pages={179--191},
  year={2009},
  publisher={Oxford Univ. Press London, UK}
}

@article{fox2015timing,
  title={Timing in response-initiated fixed intervals},
  author={Fox, Adam E and Kyonka, Elizabeth GE},
  journal={Journal of the experimental analysis of behavior},
  volume={103},
  number={2},
  pages={375--392},
  year={2015},
  publisher={Wiley Online Library}
}

@article{zeiler1994temporal,
  title={Temporal control in fixed-interval schedules},
  author={Zeiler, Michael D and Powell, David G},
  journal={Journal of the Experimental Analysis of Behavior},
  volume={61},
  number={1},
  pages={1--9},
  year={1994},
  publisher={Wiley Online Library}
}

@article{catania1970reinforcement,
  title={Reinforcement schedules and psychophysical judgment: A study of some temporal properties of behavior},
  author={Catania, AC},
  journal={The theory of reinforcement schedules},
  year={1970},
  publisher={Appleton-Century-Crofts}
}

@article{raffin2021stable,
  title={Stable-baselines3: Reliable reinforcement learning implementations},
  author={Raffin, Antonin and Hill, Ashley and Gleave, Adam and Kanervisto, Anssi and Ernestus, Maximilian and Dormann, Noah},
  journal={Journal of machine learning research},
  volume={22},
  number={268},
  pages={1--8},
  year={2021}
}

@article{towers2024gymnasium,
  title={Gymnasium: A standard interface for reinforcement learning environments},
  author={Towers, Mark and Kwiatkowski, Ariel and Terry, Jordan and Balis, John U and De Cola, Gianluca and Deleu, Tristan and Goul{\~a}o, Manuel and Kallinteris, Andreas and Krimmel, Markus and KG, Arjun and others},
  journal={arXiv preprint arXiv:2407.17032},
  year={2024}
}

@article{schulman2017proximal,
  title={Proximal policy optimization algorithms},
  author={Schulman, John and Wolski, Filip and Dhariwal, Prafulla and Radford, Alec and Klimov, Oleg},
  journal={arXiv preprint arXiv:1707.06347},
  year={2017}
}

@article{kopec2010human,
  title={Human performance on the temporal bisection task},
  author={Kopec, Charles D and Brody, Carlos D},
  journal={Brain and cognition},
  volume={74},
  number={3},
  pages={262--272},
  year={2010},
  publisher={Elsevier}
}

@article{levy2015memory,
  title={Memory bias in the temporal bisection point},
  author={Levy, Joshua M and Namboodiri, Vijay MK and Hussain Shuler, Marshall G},
  journal={Frontiers in integrative neuroscience},
  volume={9},
  pages={44},
  year={2015},
  publisher={Frontiers Media SA}
}

@article{yu2017continuous,
  title={Continuous timescale long-short term memory neural network for human intent understanding},
  author={Yu, Zhibin and Moirangthem, Dennis S and Lee, Minho},
  journal={Frontiers in neurorobotics},
  volume={11},
  pages={42},
  year={2017},
  publisher={Frontiers Media SA}
}

@article{beer1995dynamics,
  title={On the dynamics of small continuous-time recurrent neural networks},
  author={Beer, Randall D},
  journal={Adaptive Behavior},
  volume={3},
  number={4},
  pages={469--509},
  year={1995},
  publisher={Sage Publications Sage CA: Thousand Oaks, CA}
}

@article{funahashi1993approximation,
  title={Approximation of dynamical systems by continuous time recurrent neural networks},
  author={Funahashi, Ken-ichi and Nakamura, Yuichi},
  journal={Neural networks},
  volume={6},
  number={6},
  pages={801--806},
  year={1993},
  publisher={Elsevier}
}

@article{hochreiter1997long,
  title={Long short-term memory},
  author={Hochreiter, Sepp and Schmidhuber, J{\"u}rgen},
  journal={Neural computation},
  volume={9},
  number={8},
  pages={1735--1780},
  year={1997},
  publisher={MIT press}
}

@article{lecun1989handwritten,
  title={Handwritten digit recognition with a back-propagation network},
  author={LeCun, Yann and Boser, Bernhard and Denker, John and Henderson, Donnie and Howard, Richard and Hubbard, Wayne and Jackel, Lawrence},
  journal={Advances in neural information processing systems},
  volume={2},
  year={1989}
}

@article{droit2001temporal,
  title={Temporal bisection in children},
  author={Droit-Volet, Sylvie and Wearden, John H},
  journal={Journal of experimental child psychology},
  volume={80},
  number={2},
  pages={142--159},
  year={2001},
  publisher={Elsevier}
}

@article{aufschnaiter2022investigating,
  title={Investigating time-based expectancy beyond binary timing scenarios: evidence from a paradigm employing three predictive pre-target intervals},
  author={Aufschnaiter, Stefanie and Zhao, Fang and Gaschler, Robert and Kiesel, Andrea and Thomaschke, Roland},
  journal={Psychological research},
  volume={86},
  number={6},
  pages={2007--2020},
  year={2022},
  publisher={Springer}
}

@article{kleinman2016two,
  title={A two-stage model of concurrent interval timing in monkeys},
  author={Kleinman, Matthew R and Sohn, Hansem and Lee, Daeyeol},
  journal={Journal of Neurophysiology},
  year={2016},
  publisher={American Physiological Society Bethesda, MD}
}

@article{meck1984simultaneous,
  title={Simultaneous temporal processing.},
  author={Meck, Warren H and Church, Russell M},
  journal={Journal of Experimental Psychology: Animal Behavior Processes},
  volume={10},
  number={1},
  pages={1},
  year={1984},
  publisher={American Psychological Association}
}

@article{leak1995simultaneous,
  title={Simultaneous timing of multiple intervals: implications of the scalar property.},
  author={Leak, Theresa M and Gibbon, John},
  journal={Journal of Experimental Psychology: Animal Behavior Processes},
  volume={21},
  number={1},
  pages={3},
  year={1995},
  publisher={American Psychological Association}
}

@article{ng2021ecological,
  title={The ecological significance of time sense in animals},
  author={Ng, Leslie and Garcia, Jair E and Dyer, Adrian G and Stuart-Fox, Devi},
  journal={Biological Reviews},
  volume={96},
  number={2},
  pages={526--540},
  year={2021},
  publisher={Wiley Online Library}
}

@article{artificial_yang_2021,
	title = {Artificial neural networks for neuroscientists: a primer},
	doi = {10.1016/J.NEURON.2021.01.022},
	author = {Yang, Guangyu Robert and Wang, Xiao-Jing},
	journal = {Neuron},
	year = {2021},
	journal = {Neuron},
	researchRabbitId = {1943f9e6-7c76-40e1-940c-09e2db83a498}
}

@article{interpretability_kar_2022,
	title = {Interpretability of artificial neural network models in artificial intelligence versus neuroscience},
	doi = {10.1038/S42256-022-00592-3},
	author = {Kar, Kohitij and Kornblith, Simon and Fedorenko, Evelina},
	journal = {Nat. Mac. Intell.},
	year = {2022},
	journal = {Nat. Mac. Intell.},
	researchRabbitId = {0dfad8bf-0f17-4215-85e0-0dba44e7aae0}
}

@article{convolutional_lindsay_2021,
	title = {Convolutional Neural Networks as a Model of the Visual System: Past, Present, and Future.},
	doi = {10.1162/JOCN_A_01544},
	author = {Lindsay, Grace W.},
	journal = {Journal of Cognitive Neuroscience},
	year = {2021},
	journal = {Journal of Cognitive Neuroscience},
	researchRabbitId = {871393a0-49be-4ff3-a42d-bdc5ff6056da}
}

@incollection{ludvig2011primer,
  title={A primer on reinforcement learning in the brain: Psychological, computational, and neural perspectives},
  author={Ludvig, Elliot A and Bellemare, Marc G and Pearson, Keir G},
  booktitle={Computational neuroscience for advancing artificial intelligence: Models, methods and applications},
  pages={111--144},
  year={2011},
  publisher={IGI Global Scientific Publishing}
}

@article{task_yang_2019,
	title = {Task representations in neural networks trained to perform many cognitive tasks},
	doi = {10.1038/S41593-018-0310-2},
	author = {Yang, Guangyu Robert and Joglekar, Madhura R. and Song, Hui and Newsome, William T. and Wang, Xiao‐Jing},
	journal = {Nature Neuroscience},
	year = {2019},
	journal = {Nature Neuroscience},
	researchRabbitId = {0e78a864-a517-477f-b992-6fa528c3940a}
}

@article{artificial_kanwisher_2023,
	title = {Using artificial neural networks to ask ‘why’ questions of minds and brains},
	doi = {10.1016/J.TINS.2022.12.008},
	author = {Kanwisher, Nancy and Khosla, Meenakshi and Dobs, Katharina},
	journal = {Trends in Neurosciences},
	year = {2023},
	journal = {Trends in Neurosciences},
	researchRabbitId = {6539abe7-405f-48ce-9b7c-1e4b42bd69aa}
}

@incollection{bausenhart2018assessing,
  title={Assessing duration discrimination: Psychophysical methods and psychometric function analysis},
  author={Bausenhart, Karin M and Di Luca, Massimiliano and Ulrich, Rolf},
  booktitle={Timing and time perception: Procedures, measures, \& applications},
  pages={52--78},
  year={2018},
  publisher={Brill}
}

@book{sutton1998reinforcement,
  title={Reinforcement learning: An introduction},
  author={Sutton, Richard S and Barto, Andrew G and Barto, Andrew},
  volume={1},
  number={1},
  year={1998},
  publisher={MIT press Cambridge}
}

@article{watson2026let,
  title={Let it Cook: Learning to Wait in Sequential Decision Making},
  author={Watson, Christopher and Krishna, Arjun and Jayaraman, Dinesh and Alur, Rajeev},
  journal={arXiv preprint arXiv:2608.11511},
  year={2026}
}

@article{jia2025time,
  title={Time as a Control Dimension in Robot Learning},
  author={Jia, Yinsen and Chen, Boyuan},
  journal={arXiv preprint arXiv:2511.07654},
  year={2025}
}

@article{shouval2014does,
  title={What does scalar timing tell us about neural dynamics?},
  author={Shouval, Harel Z and Hussain Shuler, Marshall G and Agarwal, Animesh and Gavornik, Jeffrey P},
  journal={Frontiers in Human Neuroscience},
  volume={8},
  pages={438},
  year={2014},
  publisher={Frontiers Media SA}
}

@article{marshall2015everywhere,
  title={Everywhere and everything: The power and ubiquity of time},
  author={Marshall, Andrew T and Kirkpatrick, Kimberly},
  journal={International journal of comparative psychology},
  volume={28},
  pages={http--escholarship},
  year={2015}
}

@article{guilhardi2010effect,
  title={The effect of stimulus discriminability on strategies for learning multiple temporal discriminations},
  author={Guilhardi, Paulo and Menez, Marina and Caetano, Marcelo S and Church, Russell M},
  journal={Behavioural processes},
  volume={84},
  number={1},
  pages={476--483},
  year={2010},
  publisher={Elsevier}
}

@article{rice2014pigeons,
  title={Pigeons learn signal-food intervals independently in a multiple peak procedure.},
  author={Rice, Nathaniel and Grace, Randolph C and Kyonka, Elizabeth GE},
  journal={Journal of Experimental Psychology: Animal Learning and Cognition},
  volume={40},
  number={2},
  pages={241},
  year={2014},
  publisher={American Psychological Association}
}

@article{nepomoceno2020temporal,
  title={Temporal regularity and stimulus control in multiple fixed interval schedules},
  author={Nepomoceno, Estela B and Cravo, Andr{\'e} M and Reyes, Marcelo B and Caetano, Marcelo S},
  journal={Behavioural processes},
  volume={171},
  pages={104019},
  year={2020},
  publisher={Elsevier}
}

@article{buhusi2009interval,
  title={Interval timing accuracy and scalar timing in C57BL/6 mice.},
  author={Buhusi, Catalin V and Aziz, Dyana and Winslow, David and Carter, Rickey E and Swearingen, Joshua E and Buhusi, Mona C},
  journal={Behavioral neuroscience},
  volume={123},
  number={5},
  pages={1102},
  year={2009},
  publisher={American Psychological Association}
}

@article{caetano2012stimulus,
  title={Stimulus control in multiple temporal discriminations},
  author={Caetano, Marcelo S and Guilhardi, Paulo and Church, Russell M},
  journal={Learning \& Behavior},
  volume={40},
  number={4},
  pages={520--529},
  year={2012},
  publisher={Springer}
}

@article{labliuk2015determinants,
  title={Determinants of temporal or stimulus control in humans and rats},
  author={Labliuk, Tatiane Panzarini and Guilhardi, Paulo and Cravo, Andr{\'e} Mascioli and Church, Russell M and Caetano, Marcelo Salvador},
  journal={International Journal of Comparative Psychology},
  volume={28},
  number={1},
  year={2015}
}

@article{aggadi2025rats,
  title={Rats and mice rapidly update timed behaviors},
  author={Aggadi, Nada and Krikawa, Sophia and Paine, Tracie A and Simen, Patrick and Howard, Christopher D},
  journal={Animal cognition},
  volume={28},
  number={1},
  pages={6},
  year={2025},
  publisher={Springer}
}

@inproceedings{bengio2009curriculum,
  title={Curriculum learning},
  author={Bengio, Yoshua and Louradour, J{\'e}r{\^o}me and Collobert, Ronan and Weston, Jason},
  booktitle={Proceedings of the 26th annual international conference on machine learning},
  pages={41--48},
  year={2009}
}

@article{wolf2018joint,
  title={Joint action coordination in expert-novice pairs: Can experts predict novices’ suboptimal timing?},
  author={Wolf, Thomas and Sebanz, Natalie and Knoblich, G{\"u}nther},
  journal={Cognition},
  volume={178},
  pages={103--108},
  year={2018},
  publisher={Elsevier}
}

@article{jung2011sharing,
  title={Sharing a bimanual task between two: evidence of temporal alignment in interpersonal coordination},
  author={Jung, Christina and Holl{\"a}nder, Antje and M{\"u}ller, Karsten and Prinz, Wolfgang},
  journal={Experimental brain research},
  volume={211},
  number={3},
  pages={471--482},
  year={2011},
  publisher={Springer}
}

@article{miyata2021vocal,
  title={Vocal interaction during rhythmic joint action stabilizes interpersonal coordination and individual movement timing.},
  author={Miyata, Kohei and Varlet, Manuel and Miura, Akito and Kudo, Kazutoshi and Keller, Peter E},
  journal={Journal of Experimental Psychology: General},
  volume={150},
  number={2},
  pages={385},
  year={2021},
  publisher={American Psychological Association}
}

@String{Springer = "Springer-Verlag" }


\appendix
\section{Appendix - Multi-Timer FI}
\label{multi-timer_appendix}
\subsection{Baseline results}
In this study, for the multi-timer FI, we introduce target durations for CL in ascending order. TDs are kept short to isolate the effect of flexible temporal representation from that of long term memory capacity. There are two TD configurations used for this analysis, 2-4-6-8-10 and 3-6-9-12-15, to go up to 5 timers in the last CL stage. Number of valid TDs (n) in a given CL stage define the n-timer setting. The first CL stage (CL stage 0) focusses on training the agent on the FI task to facilitate better convergence (see Appendix Additional information \ref{multi-timer_cl_training_additional_info} section for more information). Each CL stage progressively adds a TD from the corresponding TD configuration, to the existing task after a pre-defined performance is reached.  The thresholds were 0.95 for CL Stages 0 and CL stage 1 (FI and 1-timer, respectively) and 0.90 for CL Stages 2 to 5 (from 2-timer to 5-timer setting).

The RL agents are trained for two variants of the task, an exact-timer (buffer = 0) and a categorical timer, where the valid oven check is smaller than or equal to the TD and greater than the preceding smallest TD of the corresponding TD configuration, or 0 when TD is the smallest duration. For example, for anchor configuration 2-4-6-8-10, for TD =2, agent can check the oven any time between 0 and 2 time steps of the oven timer, while for TD=4, oven can be checked any time between 2 and 4 time steps.  Each variant is evaluated using 4 runs from each model with a maximum of 1750k training steps. The task performance in the multi-timer task is quantified using the average reward of the RL agent across different seeds and runs.


\subsubsection{Task difficulty} All models performed better in the categorical multi-timer variant in comparison to the exact-timer variant. All models in both recurrent architectures reached the 5-timer setting (last CL stage) for both anchor configurations in the categorical variant (see Figure \ref{fig:multitimer_taskperf_cat}). However, for the exact-timer variant, most runs failed already in the 1-timer and 2-timer setting (see Appendix Figure \ref{fig:multitimer_taskperf_exact_supp}). CNN on the other hand, could not go beyond 1-timer setting even in the categorical variant and could not go beyond CL stage 0 (FI setting) in the exact timer variant. Thus in general, accurate timing was found to be more difficult than a more relaxed categorical timing across all models. 

\begin{figure}[ht]
  \centering
  \includegraphics[width=\linewidth]{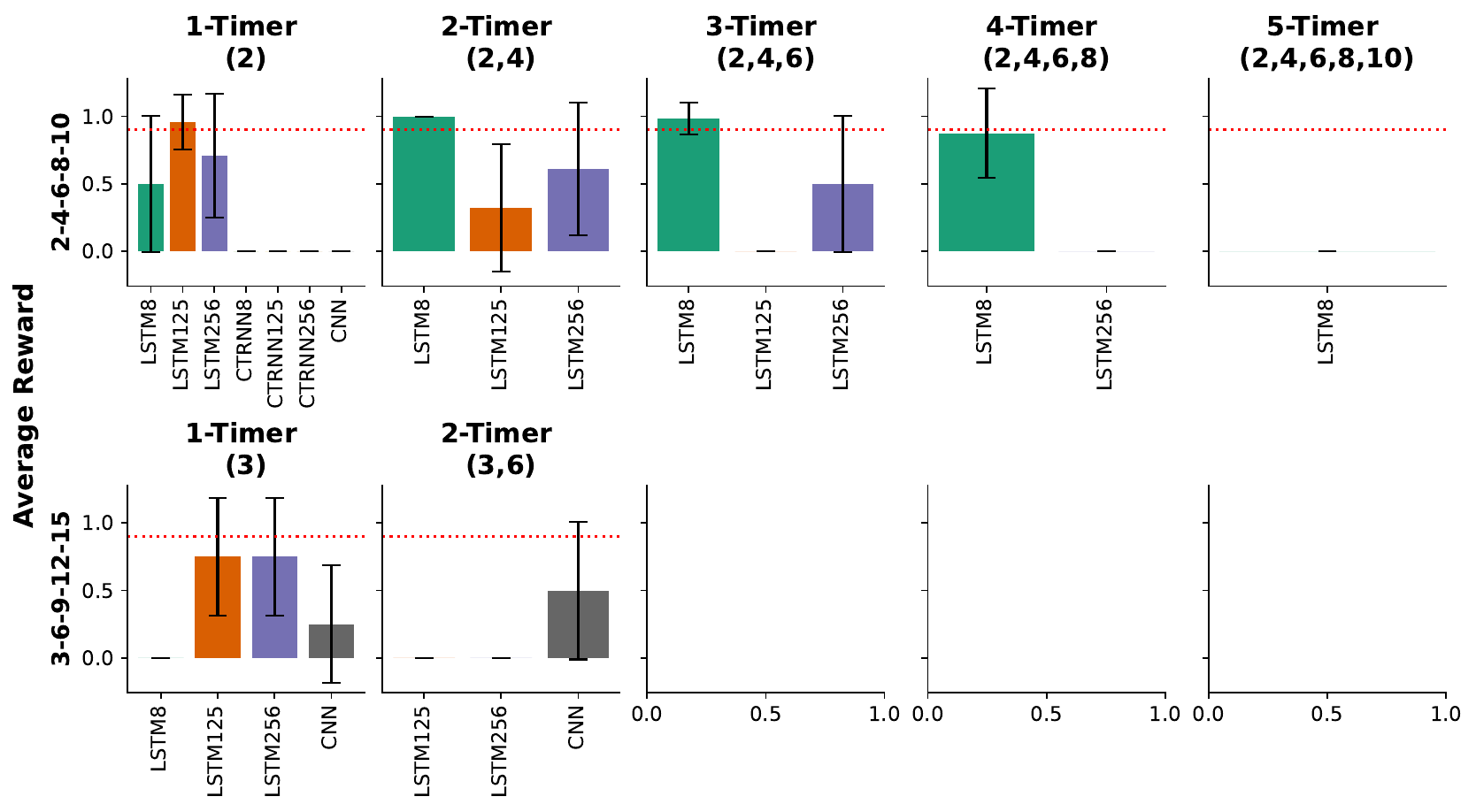}
  \caption{Multi-timer exact variant (task performance): The plots show the average reward of the different models (x-axis) in the different CL stages (columns) for the 2 TD configurations (rows) of the multi-timing exact-timer variant.}
  \label{fig:multitimer_taskperf_exact_supp}
  \Description{CTRNN models fail already in the 1-timer stage. The LSTM models perform better in anchor configuration with smaller distance between consecutive anchors. LSTM8 reached up to 5-timer stage surpassing the larger memory models. This could be attributed to a fix timing policy followed by this model.}
\end{figure}

\subsubsection{Task performance}
In the categorical variant, CTRNN models showed a better overall task performance than LSTM models for the 5-timer setting (see Figure \ref{fig:multitimer_taskperf_cat}). In addition to the low performance of $LSTM8$, the two higher memory LSTM models ($LSTM125$ and $LSTM256$) also showed a lower task performance compared to all CTRNN models. In contrast, for the exact-timer variant, all CTRNN models failed to converge even for 1-timer setting and hence LSTM models in general had a better overall task performance (see Appendix Figure \ref{fig:multitimer_taskperf_exact_supp}).

\begin{figure}[ht]
  \centering
  \includegraphics[width=\linewidth]{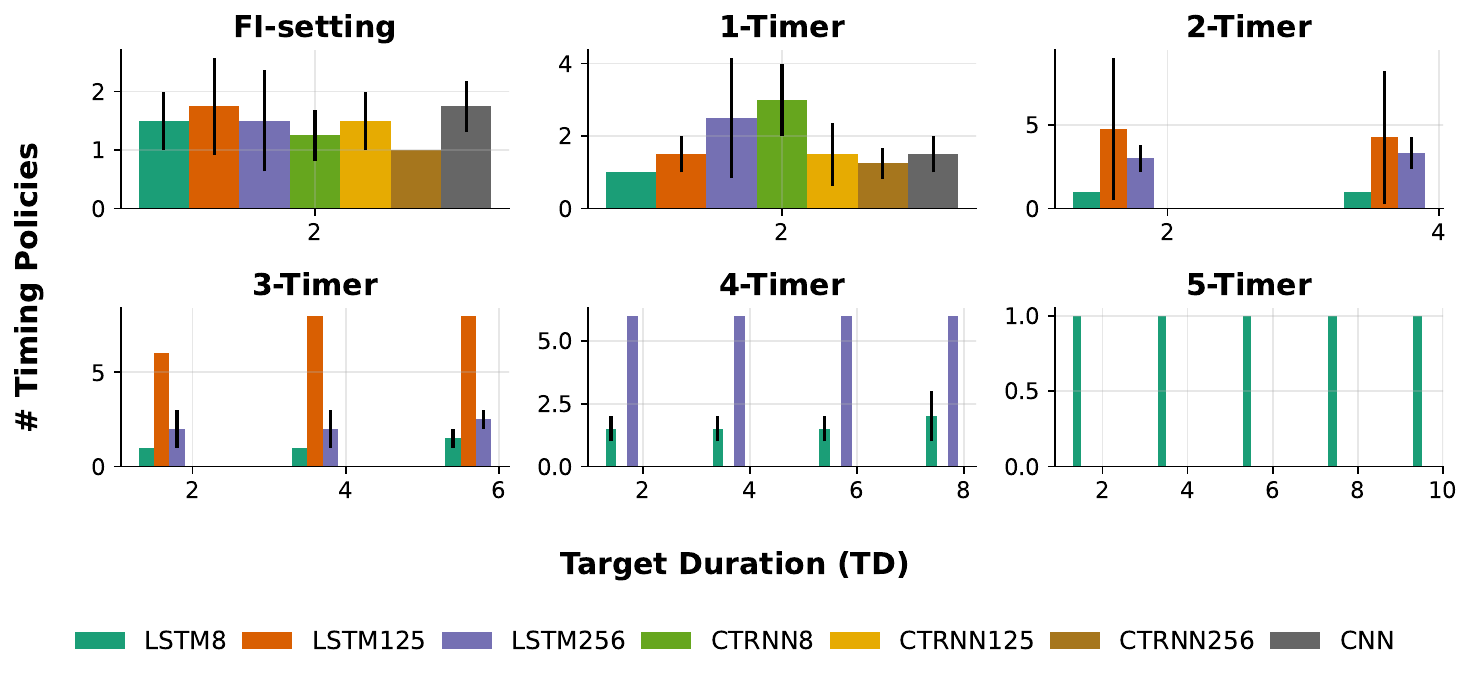}
  \caption{Multi-timer, exact-timer variant (\# timing policies): The plots show the number of unique timing policies in the different seeds, averaged across runs, for the different RL agents (shown in different colours) for the anchor configuration 2-4-6-8-10 in the exact-timer variant. The plot titles indicate the different CL stages. }
  \label{fig:multitimer_num_timing_actions_exact}
  \Description{As in the FI timing task, LSTM8 showed a lower number of timing action sequences compared to the two higher memory LSTM models (see Figure \ref{fig:multitimer_num_timing_actions_exact}), suggesting greater reliance on embodied timing (a fixed action sequence) rather than implicit time-keeping. This could be one of the possible reasons for LSTM8 model to successfully reach 5-timer setting in the more difficult exact-timer variant of the multi-timer task.}
\end{figure}

Interestingly, two runs from the smaller memory $LSTM8$ model successfully converged in the 4-timer setting and reached the 5-timer setting, surpassing the higher-memory LSTM models in the exact-timing task. As in the FI timing task, $LSTM8$ showed a lower number of timing action sequences compared to the two higher memory LSTM models (see Appendix Figure \ref{fig:multitimer_num_timing_actions_exact}), suggesting greater reliance on embodied timing (a fixed action sequence) rather than implicit time-keeping. This raises the possibility that exact timing, which requires precisely tracking a duration, may be more reliably solved through embodied strategies, while implicit time-keeping may struggle with the precision this task demands. However, this observation is based on only two successful runs and should be treated as preliminary. Additionally, the failure of all CTRNN models in the exact-timing task, paired with their comparatively high performance in the categorical variant, indicates an architectural shortcoming in CTRNN for exact-timing that needs further investigation. 

 \begin{figure}[ht]
  \centering
  \includegraphics[width=0.8\linewidth]{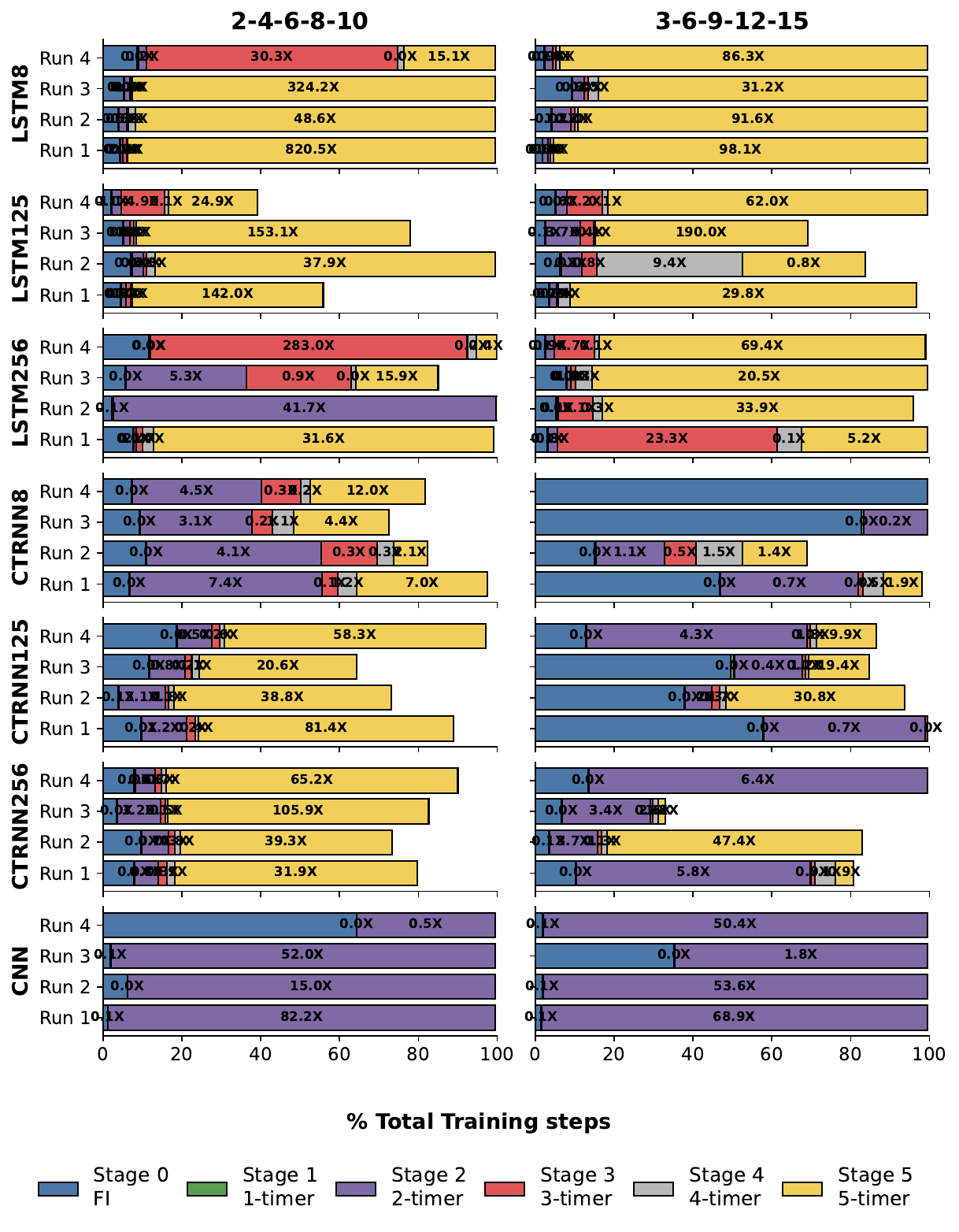}
  \caption{Multi-timer categorical variant (CL-efficiency): The training time steps required by the different runs of the the models (rows) for the TD configurations (columns), split by CL stages (shown in different colours) for categorical variant.  The numbers in the bars represent the corresponding CL-efficiency ratios.}
  \label{fig:multitimer_scalability_cat_supp}
  \Description{Except for CNN, at least one run from all models reach the last CL stage (5-timer) in the categorical timer variant}
\end{figure}

\subsubsection{N-timer capacity}
At an architectural level, CTRNN showed better capacity than LSTM in the categorical timer variant. However, this was reversed in the exact-timer variant where LSTM showed a better capacity than CTRNN. For the categorical variant, while models from both architectures reached the 5-timer setting (last CL stage), within a fixed training budget (1750 k training steps), 83\% of CTRNN runs converged in the 5-timer setting before reaching the maximum training budget, compared to only 42\% in LSTM (see Appendix figure \ref{fig:multitimer_scalability_cat_supp} for training steps data of categorical variant). On the other hand, for the exact-timer variant, 63\% of runs from LSTM models trained beyond the 1-timer setting (some runs successfully converging in 2, 3 and 4-timer settings), while none of the CTRNN models crossed the 1-timer setting (see Appendix figure \ref{fig:multitimer_scalability_exact_supp} for training steps data of exact-timer variant).

\subsubsection{CL efficiency} An architecture-level analysis of the CL-efficiency ratio is conducted for the categorical variant. This analysis however, comes with a caveat that in some cases either memory sizes within an architecture have a significant effect or the anchor configurations have an overall effect on some metrics. See Appendix Additional Information \ref{multi-timer_cl_significance_test_additional_info} section for details on memory effect within each architecture and overall anchor effect for the different CL-efficiency ratios. The CL-efficiency ratio for some of the important CL stages are discussed below:

\begin{itemize}

    \item \textbf{CL stage 0 (FI setting):}
    CTRNN (N=23, M=19.23, SD=20.84, Median=10.17) models took significantly longer to converge in the first CL stage (FI setting) compared to LSTM (N=24, M=5.04, SD=2.58, Median=4.74) (Mann-Whitney U=473.00, p<0.001).
    \item \textbf{CL stage 1 (1-timer):} This stage in the categorical variant is not very different than the previous stage except having a higher penalty for overcooked (0 reward instead of 0.1). Hence, both architectures showed ratios significantly below 1 (CTRNN: N=23, M=0.02, SD=0.01, Median=  0.02, Wilcoxon W=0.00, p<0.001; LSTM: N=24, M=0.05, SD=0.03, Median=0.04, Wilcoxon, W=0.00, p<0.001) with no difference between them (Mann-Whitney U=95.00, p=0.999). 
    \item \textbf{CL stage 2 (2-Timer):} In this stage, which is also the first introduction of multi-timing task, The CL-efficiency ratio of CTRNN (M = 2.39, SD = 2.06, median = 1.18, Wilcoxon W = 168.00, p = 0.035) was significantly more than 1 but for LSTM (M = 0.84, SD = 1.24, median = 0.51, Wilcoxon W = 52.00, p = 0.997) it was not.  The CL-efficiency ratio of CTRNN was also significantly greater than LSTM (Mann-Whitney U, stat=369.00, p=0.001).
    \item \textbf{CL stage 5 (5-Timer):} In the last stage, the CL-efficiency ratio of both CTRNN (n=20, M=28.98, SD=29.97, Median=20.00, Wilcoxon W=209.0, p<0.001) and LSTM (n=10, M=69.15, SD=67.18, Median=32.74, Wilcoxon W=54.00, p=0.002) was significantly more than 1. There was no significant difference between the two architectures (Mann-Whitney U=66.00, p=0.935).

\end{itemize}


Hence, architecture-level analysis of the CL-efficiency ratio in the categorical variant shows that, CTRNN required significantly more training steps than LSTM specifically at transitions introducing new task elements, namely, CL stage 0 (the initial FI stage with CTRNN M=19.23 vs. LSTM M=5.04, p<0.001) and CL stage 2 (the first introduction of the multi-timer setting with CTRNN M=2.39 vs. LSTM M=0.85, p=0.001). In these stages, CTRNN's ratio was significantly more than 1, unlike LSTM. For CL stages where no new task element was introduced like 1-timer setting or 3 or 4 timer setting, both architectures showed ratios significantly below 1 with no difference between them.  At the final CL stage, as task complexity accumulates, both architectures require comparable additional training with CL-ratio significantly above 1. This suggests that CTRNN's slower learning can be associated with stages requiring the acquisition of substantially new timing capabilities.  This observation combined with CTRNN's better task performance than LSTM in the final CL stage of the categorical variant suggests that, once CTRNN models successfully learn to incorporate a new timing requirement, they are able to reach a level of overall CL efficiency and task performance comparable to, or exceeding, the more powerful LSTM models.

\begin{figure}[ht]
  \centering
  \includegraphics[width=0.8\linewidth]{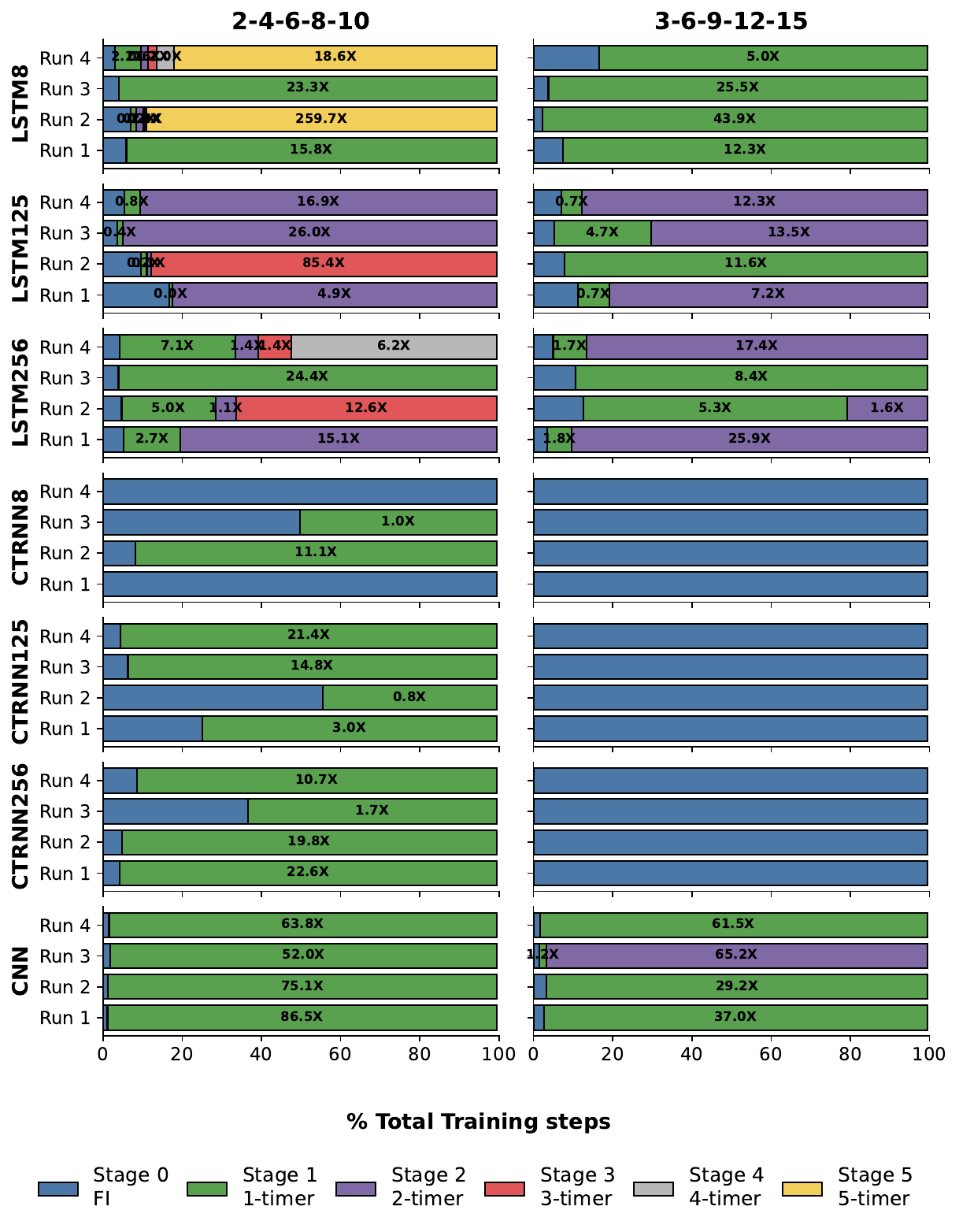}
  \caption{Multi-timer exact-timer variant (CL-efficiency): The training time steps required by the different runs of the the models (rows) for the TD configurations (columns), split by CL stages (shown in different colours) for exact-timer variant. The numbers in the bars represent the corresponding CL-efficiency ratios. }
  \label{fig:multitimer_scalability_exact_supp}
  \Description{All runs in the CTRNN models fail already in the 1-timer stage. The LSTM models perform better in anchor configuration with smaller distance between consecutive anchors. 2 runs from LSTM8 reached up to 5-timer stage surpassing the larger memory models. This could be attributed to a fix timing policy followed by this model}
\end{figure}

\subsection{Additional Information}
\subsubsection{CL training details}
\label{multi-timer_cl_training_additional_info}
In the first CL stage, the focus is on training the agent for the basic soup delivery task, because it was found that starting with a strict timing requirement resulted in very low convergence rate across the different models. Hence, in the first CL stage, the soup cannot be retrieved before TD and retrieving it after TD results in a reduced reward of 0.1 Thus, the optimal policy is to check the oven at or before TD.

In the next CL stage, the timing conditions become stricter, the optimal policy requires the agent to check the oven within TD $\pm$ buffer, which results in a reward of +1 after successful soup delivery (a successful timing and soup delivery task). Checking the oven before or after  TD $\pm$ buffer, results in no reward even after successful soup delivery (unsuccessful timing task but successful soup delivery task). 

In subsequent stages, after a predefined performance threshold is reached, additional TDs are introduced incrementally. Thus in the later stages, the RL agent must associate different oven states with each TD and produce the corresponding amount of time to receive a reward of +1 after successful soup delivery.

\subsubsection{CL significance tests}
\label{multi-timer_cl_significance_test_additional_info}
The significance test for the CL-efficiency ratio of the multi-timer categorical variant is given below: 

\begin{itemize}
    \item CL stage 0: At an architecture level, there was no significant difference in memory size (F(2,20)=0.44, p= 0.651) or anchor duration (F(1,20)=1.35 , p=0.259) for LSTM. In contrast, for CTRNN there was no significant difference in memory size (F(2,19)=3.26, p= 0.061) but there was a significant difference due to anchor duration (F(1,19)=10.55, p=0.004). The CTRNN models took longer time to converge in CL first stage with anchor configuration 3-6-9-12-15 (N=11, M=30.53, SD=25.88) than 2-4-6-8-10 (N=12, M=8.87, SD=3.95). The high variation is due to CTRNN256 (N=8, M=7.79, SD=3.38) which took considerably lower time than CTRNN125 (N=8, M=25.19, SD=20.30) and CTRNN8 (N=7,M=25.49, SD=28.84), both these models show higher convergence time for anchor 3-6-9-12-15. CTRNN (N=23, M=19.23, SD=20.84) was found to be significantly greater than LSTM (N=24, M=5.04, SD=2.58) (U=473.00, p<0.001) but with the caveat that anchor configuration has a significant effect for CTRNN models.

    \item CL stage1 : Within LSTM, no significant difference was found in memory size (F(2,20)=0.68, p=0.519) or anchor duration (F(1,20)=0.49, p=0.493). Memory sizes 8, 125, and 256 showed comparable means (M=0.041, SD=0.021, N=8; M=0.052, SD=0.026, N=8; M=0.060, SD=0.045, N=8, respectively), as did the two anchor conditions (2-4-6-8-10: M=0.046, SD=0.034, N=12; 3-6-9-12-15: M=0.055, SD=0.031, N=12). For CTRNN, given non-normal residuals (Shapiro-Wilk p<0.001) and a non-significant, small anchor effect (F(1,19)=1.03, p=0.322), memory size was tested using Kruskal-Wallis, ignoring anchor. Memory sizes 8, 125, and 256 showed means of M=0.021 (SD=0.011, N=7), M=0.018 (SD=0.018, N=8), and M=0.029 (SD=0.016, N=8), respectively. No significant effect of memory size was found (Kruskal-Wallis H=4.45, p=0.108, df=2). No significant difference was found between the two architectures ( Mann-Whitney U=95.000, p=0.9999). Both CTRNN (Wilcoxon, W=0.000, p<0.0001, n=23) and LSTM (W=0.000, p<0.0001, n=24) showed CL-efficiency ratios significantly less than 1 at this CL stage, indicating that training required fewer steps than the preceding CL stage for nearly all runs across both architectures.

    \item CL stage 2 (first introduction of multi-timer): CL-efficiency of CTRNN was significantly greater than LSTM (Mann-Whitney U, stat=369.00, p=0.001). Cl-efficiency ratio of CTRNN was also significantly more than 1 but for LSTM it was not (CTRNN: Wilcoxon W = 168.00, p = .035, M = 2.39, SD = 2.06, median = 1.18; LSTM: Wilcoxon W = 52.000, p = 0.997, M = 0.85, SD = 1.24, median = 0.51).

    \item 
    Last CL stage (5-timer): 
    The full model (architecture × memory-size + anchor) found no significant effect of architecture (F(1,24)=0.00, p=1.000), after accounting for memory size, anchor configuration, and their interaction. Memory size showed a significant effect (F(2,24)=4.03, p=0.031), while anchor (F(1,24)=1.85, p=0.186) and the architecture × memory size interaction (F(2,24)=2.71, p=0.087) did not reach significance. Within-architecture analyses showed that LSTM was not significantly affected by memory size (F(1,7)=1.36, p=0.281) or anchor (F(1,7)=0.04, p=0.848), whereas CTRNN showed significant effects of both memory size (F(2,16)=5.66, p=0.014) and anchor (F(1,16)=7.18, p=0.016). CL-efficiency in all models was significantly greater than 1 (CTRNN8: t(5) = 2.276, p = .036, M = 4.79, SD = 4.08, median = 3.27; CTRNN125: t(6) = 3.796, p = .005, M = 37.03, SD = 25.11, median = 30.77; CTRNN256: t(6) = 2.927, p = .013, M = 41.67, SD = 36.77, median = 39.29; LSTM125: t(5) = 2.708, p = .021, M = 90.12, SD = 80.62, median = 85.89; LSTM256: t(3) = 3.247, p = .024, M = 37.71, SD = 22.61, median = 32.74)

\end{itemize}

\newpage

\section{Appendix - Multi-Anchor Categorization}
\label{multianchor_appendix}
\subsection{Baseline Results}

The RL agents that reliably performed the temporal bisection task ($LSTM125$, $LSTM256$, $CTRNN125$, $CTRNN256$) were tested for multi-anchor categorization ability. The analysis is conducted for 4 types of main anchor configurations, namely, 4-8-12-18, 4-12-18-25, 3-10-20-25 and 5-10-15-20. The models were trained using CL starting from 2 anchors and progressively adding a new anchor after a performance threshold was reached. For example, the main anchor configuration, 4-8-12-18, starts with a 2-anchor setting with 4 and 8 as its anchors (CL stage 1). After a certain performance threshold, the training transitions to 3-anchor setting with 4,8 and 12 as the anchor durations  (CL stage 2). Similarly for 4-anchor setting, the anchor durations become 4,8,12 and 18  (CL stage 3). The performance threshold for 2-anchor setting and 3-anchor setting was 0.98 and 0.96 respectively, thus making sure the agent successfully performs the 3-anchor task before transitioning to the 4-anchor setting. The result analysis is restricted to 4 anchors, since the focus is on studying flexible temporal representations rather than testing model capacity, which is primarily determined by memory size and training duration. The analysis is conducted using 4 runs from each model with a maximum of 1750k training steps

\begin{figure}[ht]
  \centering
  \includegraphics[width=\linewidth]{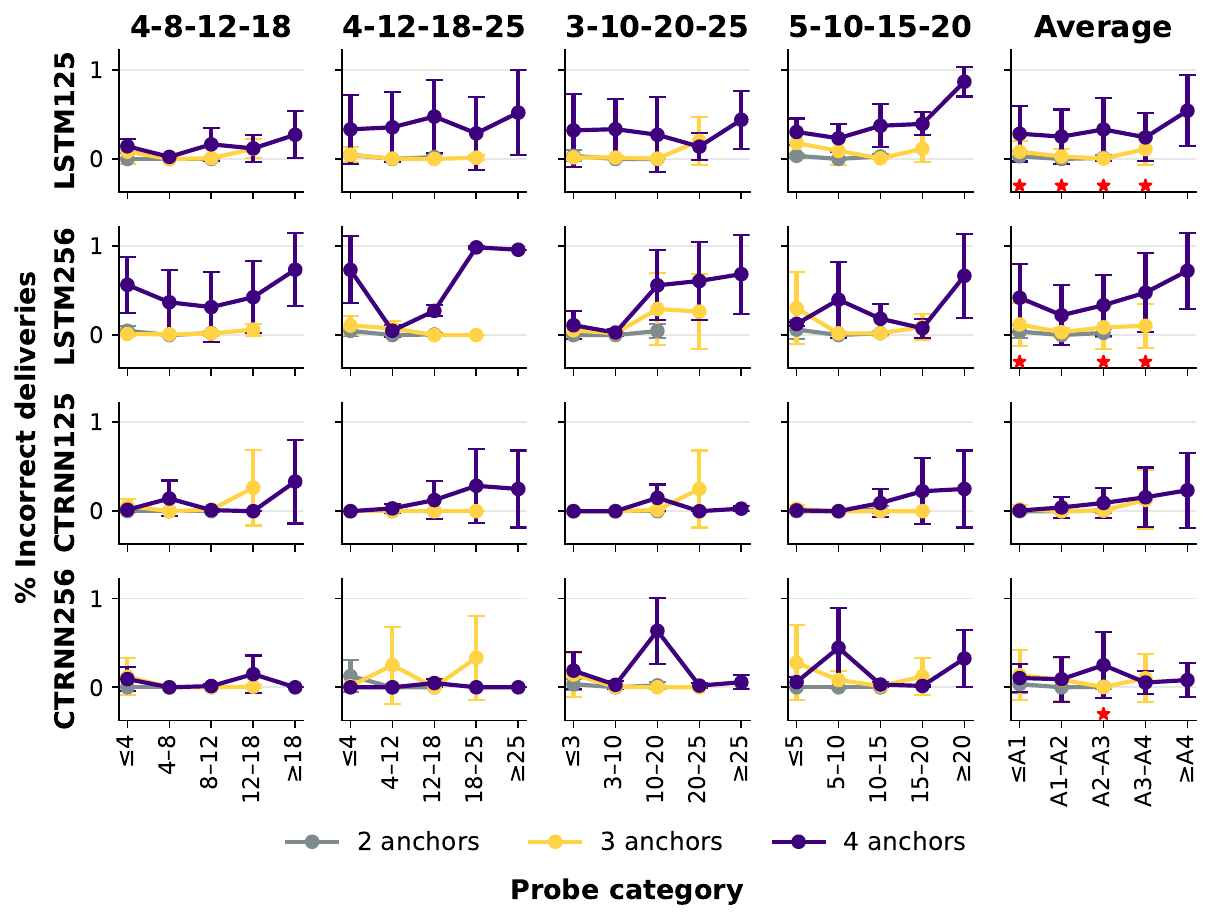}
  \caption{Multi-anchor categorization task performance: The plots show the  percentage of incorrect deliveries across runs and seeds for a given probe duration category, with error bars indicating +- standard deviation. The final column shows the mean and standard deviation across runs and anchor configurations. Lower values indicate better performance;  Models should be compared along each probe-duration interval (<=A1, A1–A2, etc.) to isolate the effect of probe durations. The red stars in the average column indicates the cases when anchor 4 performance was significantly worse than corresponding anchor 3 performance. At an architecture level, for 4 anchor setting, CTRNN yielded significantly fewer incorrect deliveries than LSTM in 3 of the 5 probe categories: Probe category 0 (F = 14.93, p < .001), Probe category 1 (F = 4.48, p = .041), and Probe category 4 (F = 9.27, p = .004) }
  \label{fig:multianchor_task_performance_supp}
  \Description{ At an architecture level, for 4 anchor setting, CTRNN yielded significantly fewer incorrect deliveries than LSTM in 3 of the 5 probe categories}
\end{figure}

The task performance of the different models was quantified using \% of incorrect deliveries (as defined in Section 8.1). Appendix Figure \ref{fig:multianchor_task_performance_supp} shows the task performance of each RL agent (rows) for the different CL stages or n-anchor settings (shown in a different colour) separately for each main anchor configuration (columns). For a model-level performance, runs are pooled across the different main anchor configurations to compare the \% of incorrect deliveries between the 4-anchor and 3-anchor setting for each probe duration category. The resulting p-values are corrected using the Holm–Bonferroni method. Additionally, the task performance and scalability are evaluated at an architecture-level (across all main anchor configurations and the two memory sizes). The different data pooling for model-level and architecture-level analysis is justified since no significant differences in task performance or scalability metrics were found between the main anchor configurations or the two memory sizes of either architecture (see Appendix Additional information sections \ref{multianchor_model_level_significance_test_additional_info}, \ref{multianchor_arch_level_significance_test_additional_info}, \ref{multianchor_cl_significance_tests_additional_info}  for detailed significance test results).

\subsubsection{Temporal categorization ability} All the four RL agents exhibited temporal categorization ability beyond binary decisions. Within a fixed training budget (1750k training steps), at least 1 run of every model reached the 4-anchor CL stage (see Appendix Figure \ref{fig:Scalability_multianchor_supp}). Figure \ref{fig:multianchor_task_visualization} shows the percentage of deliveries at the different delivery counters across probe durations for various 3-anchor and 4-anchor settings. Although there are some cases of non-convergence (shown in grey) in the 4-anchor setting, the figure shows a clear transition from delivery counter 1 to 2, then 3, and then 4, as probe durations move from one extreme to the other. Importantly, this includes OOD probe durations. Hence, the RL agents not only exhibit multi-anchor temporal categorization ability, but are also able to generalize beyond the trained anchor durations.

\begin{figure}[ht]
  \centering
  \includegraphics[width=\linewidth]{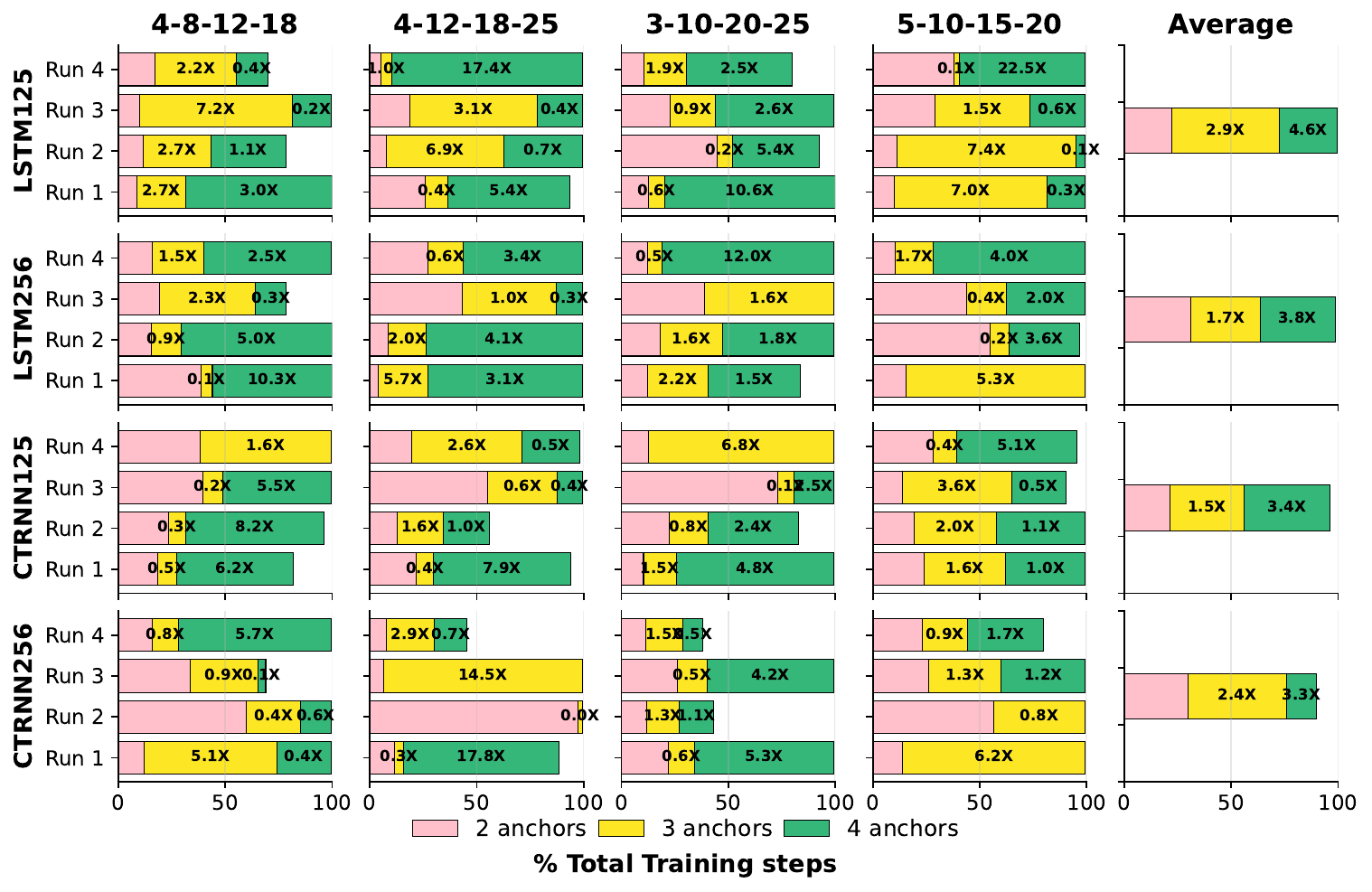}
  \caption{Multi-anchor (CL efficiency): Training time steps required by each run of the models (rows) for the different CL stages (shown in different colours) across all main anchor configurations (columns). The numbers in the bars represent the corresponding CL-efficiency ratios. The last column shows an average taken across the main anchor configurations for each model. }
  \label{fig:Scalability_multianchor_supp}
  \Description{Most runs from all models reached the last CL stage (4-anchor) across all anchor configurations.}
\end{figure}

\subsubsection{Model-level performance}
 Figure \ref{fig:multianchor_task_performance} (left plot) shows the overall task performance of each model for the different CL stages (shown in a different colour). For the LSTM models ($LSTM125$ and $LSTM256$), the 4-anchor performance (last CL stage) was significantly worse (significantly more incorrect deliveries) than the 3-anchor performance (shown using red stars) for most probe duration categories (see Appendix Table \ref{tab:multianchor_tak_perf_wilcoxon_results} for significance test results). In contrast, CTRNN models achieved comparable performance in the 4-anchor and corresponding 3-anchor settings, indicating that they reached the expected performance threshold in the final CL stage, unlike the LSTM models.
 

\subsubsection{Architecture-level performance}
A direct comparison between the 4-anchor performance of the two architectures reveals that, CTRNN models showed a better overall task performance for the 4-anchor setting in comparison to the LSTM models. In the 4-anchor setting, LSTM yielded significantly more incorrect deliveries than CTRNN across all five probe categories. The significance test results (shown as red stars in Figure \ref{fig:multianchor_task_performance} right plot) are as follows:
\begin{itemize}
\item \textbf{Probe category A1:} LSTM ($N=24$, $M = 0.38$, $SD = 0.37$) vs. CTRNN ($N=24$, $M = 0.05$, $SD = 0.12$), Mann--Whitney $U = 482.50$, $p_{\mathrm{Holm}} < .001$.

\item \textbf{Probe category A1--A2:} LSTM ($N=24$, $M = 0.23$, $SD = 0.31$) vs. CTRNN ($N=24$, $M = 0.06$, $SD = 0.20$), Mann--Whitney $U = 420.00$, $p_{\mathrm{Holm}} = .002$.

\item \textbf{Probe category A2--A3:} LSTM ($N=23$, $M = 0.32$, $SD = 0.35$) vs. CTRNN ($N=24$, $M = 0.16$, $SD = 0.30$), Mann--Whitney $U = 400.50$, $p_{\mathrm{Holm}} = .004$.

\item \textbf{Probe category A3--A4:} LSTM ($N=23$, $M = 0.35$, $SD = 0.38$) vs. CTRNN ($N=23$, $M = 0.11$, $SD = 0.28$), Mann--Whitney $U = 410.50$, $p_{\mathrm{Holm}} = .001$.

\item \textbf{Probe category A4:} LSTM ($N=22$, $M = 0.60$, $SD = 0.42$) vs. CTRNN ($N=23$, $M = 0.17$, $SD = 0.36$), Mann--Whitney $U = 423.00$, $p_{\mathrm{Holm}} < .001$.

\end{itemize}


Thus, this analysis shows further evidence that, within the same training budget, CTRNN architecture is able to perform the multi-anchor categorization task better than the more powerful LSTM architecture even for higher probe durations.

\subsubsection{N-anchor capacity} At an architectural level, while at least one run of models from both recurrent architectures reached the 4-anchor CL stage, the CTRNN architecture showcased a greater capacity to incorporate 4 anchors than LSTM. This is because, within a fixed training budget (1750k training steps), 44\% of CTRNN runs (50\% from CTRNN125 and 38\% from CTRNN256) converged in the 4-anchor setting (i.e. reached a performance threshold of 0.8) before the maximum training budget, compared to only 19\% of LSTM runs (25\% from LSTM125 and 13\% from LSTM256). See Appendix Figure \ref{fig:Scalability_multianchor_supp} for training time steps data. This is a notable result considering the greater parameter capacity of LSTM as compared to CTRNN. The latter's higher convergence rate suggests that it may support more scalable temporal strategies that facilitate multi-anchor categorization.

\subsubsection{CL efficiency}:  No evidence of a scalable time representation was found in either architectures. CL efficiency analysis at an architectural-level revealed the following observations (non-converging runs were excluded from each CL stage analysis). 

\begin{itemize}

    \item \textbf{CL stage 1 (2-anchor setting):}
    CTRNN training time was greater than LSTM training time
    (CTRNN: N=32, M=27.19, Median=21.93, SD=20.70; LSTM: N=32, M=20.76, Median=15.79, SD=13.59), but this difference was not significant (Mann-Whitney U=624.50, p=0.066).

    \item \textbf{CL stage 2 (3-anchor setting):}
    The CL-efficiency ratio of LSTM (N=30, M=2.22, SD=2.27, Median = 1.55 ) was significantly greater than 1 (Wilcoxon W=353.00, p=0.006). while for CTRNN (N=26, M=1.26, SD=1.17, Median =0.85 ) it was not (Wilcoxon W=183.00, p=0.430). However, the CL-efficiency ratio of LSTM was not significantly greater than that of CTRNN  (Mann-Whitney U=292.00, p=0.947). 

    \item \textbf{CL stage 3 (4-anchor setting):}
    The CL-efficiency ratio of both CTRNN (N=14, M=3.96, Median=1.60, SD=4.93) and LSTM (N=6, M=3.33, Median=3.42, SD=2.16) in CL stage 3, was significantly more than 1 (CTRNN: Wilcoxon W=75.00, p=0.020; LSTM: Wilcoxon W=2.643, p=0.023). Additionally, the CL-efficiency ratio of LSTM was not significantly greater than that of CTRNN  (Mann-Whitney U=37.00, p=0.670). 

\end{itemize}

Thus, the CL-efficiency ratio for the LSTM models was significantly more than 1, whereas for the CTRNN models it was significantly more than 1 only in the 4-anchor setting. Hence, neither architecture showed conclusive evidence of a flexible timing strategy. Further investigation into the changes in neural dynamics for the different n-anchor settings can help uncover flexible temporal representations in the two architectures. Nevertheless, the CTRNN architecture exhibited better overall performance than LSTM, with higher task performance and higher convergence rate in the final CL stage (4-anchor setting).

\subsection{Additional Information}

\subsubsection{Model-level performance significance tests}
\label{multianchor_model_level_significance_test_additional_info}
Table \ref{tab:multianchor_tak_perf_wilcoxon_results} shows the Wilcoxon signed-rank test results comparing 4-anchor and 3-anchor performance of each model across anchor configurations by probe duration categories. The test checks if 4-anchor setting \% of incorrect deliveries significantly greater than 3-anchor setting. Appendix Figure \ref{fig:multianchor_task_performance_supp} shows the \% of incorrect deliveries for each n-anchor setting (shown in a different colour), for each model (rows) and main anchor configuration (columns) separately. 

The main anchor configuration did not show a significant effect on task performance across the different probe duration categories (<=A1: F(3, 44) = 0.48, p = 0.699; A1-A2:  F(3, 45) = 0.46, p = 0.709; A2-A3: F(3, 44) = 1.81, p = 0.159; A3-A4: F(3, 43) = 0.89, p = 0.455; >=A4:F(3, 42) = 0.73, p = 0.541) , after accounting for architecture, memory size, and their interaction. Hence, data was pooled across the main anchor configuration for a model-level task performance.

\begin{table}[htbp]
\centering
\caption{Multi-anchor (Task performance): Wilcoxon signed-rank comparisons between the 3-anchor and 4-anchor conditions. $p$-values were corrected using the Holm method.}
\label{tab:multianchor_tak_perf_wilcoxon_results}
\begin{tabular}{llccccccl}
\hline
\textbf{Model} & \shortstack{\textbf{Probe}\\\textbf{category}} & \textbf{$N$} & \shortstack{\textbf{4-anchor}\\\textbf{Median [IQR]}} &
\shortstack{\textbf{3-anchor}\\\textbf{Median [IQR]}} & \textbf{$W$} & \textbf{$p$} & \textbf{$p_{\mathrm{Holm}}$} & \shortstack{\textbf{Reject}\\\textbf{$H_0$}} \\
\hline
LSTM125 & <=A1 & 15 & 0.19 [0.23] & 0.00 [0.16] & 76.00 & 0.016 & 0.033 & Yes \\
LSTM125 & A1-A2 & 15 & 0.11 [0.42] & 0.00 [0.01] & 75.00 & 0.002 & 0.007 & Yes \\
LSTM125 & A2-A3 & 15 & 0.15 [0.58] & 0.00 [0.00] & 102.00 & < .001 & 0.004 & Yes \\
LSTM125 & A3-A4 & 15 & 0.09 [0.33] & 0.00 [0.10] & 82.00 & 0.032 & 0.033 & Yes \\
LSTM256 & <=A1 & 12 & 0.27 [0.69] & 0.05 [0.10] & 43.00 & 0.006 & 0.023 & Yes \\
LSTM256 & A1-A2 & 13 & 0.08 [0.19] & 0.02 [0.04] & 49.00 & 0.087 & 0.087 & No \\
LSTM256 & A2-A3 & 12 & 0.18 [0.40] & 0.00 [0.01] & 60.00 & 0.007 & 0.023 & Yes \\
LSTM256 & A3-A4 & 12 & 0.53 [0.87] & 0.00 [0.06] & 56.00 & 0.021 & 0.041 & Yes \\
CTRNN125 & <=A1 & 13 & 0.00 [0.00] & 0.00 [0.00] & 1.00 & 0.938 & 0.938 & No \\
CTRNN125 & A1-A2 & 13 & 0.00 [0.00] & 0.00 [0.00] & 3.00 & 0.250 & 0.500 & No \\
CTRNN125 & A2-A3 & 13 & 0.00 [0.04] & 0.00 [0.00] & 9.00 & 0.125 & 0.375 & No \\
CTRNN125 & A3-A4 & 13 & 0.00 [0.03] & 0.00 [0.00] & 13.00 & 0.094 & 0.375 & No \\
CTRNN256 & <=A1 & 11 & 0.00 [0.18] & 0.00 [0.08] & 15.00 & 0.469 & 0.500 & No \\
CTRNN256 & A1-A2 & 11 & 0.00 [0.00] & 0.00 [0.00] & 3.00 & 0.250 & 0.500 & No \\
CTRNN256 & A2-A3 & 11 & 0.06 [0.27] & 0.00 [0.00] & 28.00 & 0.008 & 0.031 & Yes \\
CTRNN256 & A3-A4 & 10 & 0.00 [0.02] & 0.00 [0.00] & 6.00 & 0.125 & 0.375 & No \\
\hline
\end{tabular}
\end{table}

\subsubsection{Architecture-level performance significance tests}
\label{multianchor_arch_level_significance_test_additional_info}
For task performance at an architecture-level, data in the 4-anchor setting was combined across memory sizes and main anchor configurations. In addition to no significant effect of main anchor configuration as seen in model-level analysis above, there was also no significant difference between the task performance of the two memory sizes within a architecture.  Table \ref{tab:multianchor_arch_compariosn_whitney} shows the Mann-Whitney U test comparing the percentage of incorrect deliveries in 4-anchor setting between the two memory sizes (125 and 256) of each model architecture. This justifies the data pooling for an architecture-level comparison. The actual task performance comparison between LSTM and CTRNN for the 4-anchor setting at a probe duration category level is discussed in the main text.

\begin{table}[htbp]
\centering
\caption{Multi-anchor (task performance): Mann--Whitney U test comparing the percentage incorrect deliveries in 4-anchor setting between the two memory sizes (125 and 256) of a specific model architecture.}
\label{tab:multianchor_arch_compariosn_whitney}
\begin{tabular}{llccccccl}
\hline
\textbf{Model} & \shortstack{\textbf{Probe}\\\textbf{category}}  & \shortstack{\textbf{$N$}\\\textbf{125 / 256}}  & \shortstack{\textbf{Memory=125}\\\textbf{Median [IQR]}} & \shortstack{\textbf{Memory=256}\\\textbf{Median [IQR]}} & \textbf{Statistic} & \textbf{$p$-value} & \textbf{$p_{\mathrm{Holm}}$} & \shortstack{\textbf{Reject}\\\textbf{$H_0$}} \\
\hline
LSTM & <=A1 & 15 / 12 & 0.19 [0.23] & 0.27 [0.69] & 76.50 & 0.524 & 1.000 & No \\
LSTM & A1-A2 & 15 / 13 & 0.11 [0.42] & 0.08 [0.19] & 106.50 & 0.689 & 1.000 & No \\
LSTM & A2-A3 & 15 / 12 & 0.15 [0.58] & 0.18 [0.40] & 84.00 & 0.788 & 1.000 & No \\
LSTM & A3-A4 & 15 / 12 & 0.09 [0.33] & 0.53 [0.87] & 79.00 & 0.606 & 1.000 & No \\
LSTM & >=A4 & 15 / 11 & 0.59 [0.86] & 1.00 [0.51] & 67.00 & 0.418 & 1.000 & No \\
CTRNN & <=A1 & 13 / 11 & 0.00 [0.00] & 0.00 [0.18] & 52.50 & 0.159 & 0.648 & No \\
CTRNN & A1-A2 & 13 / 11 & 0.00 [0.00] & 0.00 [0.00] & 69.50 & 0.894 & 1.000 & No \\
CTRNN & A2-A3 & 13 / 11 & 0.00 [0.04] & 0.06 [0.27] & 47.00 & 0.130 & 0.648 & No \\
CTRNN & A3-A4 & 13 / 10 & 0.00 [0.03] & 0.00 [0.02] & 67.50 & 0.879 & 1.000 & No \\
CTRNN & >=A4 & 13 / 10 & 0.00 [0.06] & 0.00 [0.00] & 74.00 & 0.494 & 1.000 & No \\
\hline
\end{tabular}
\end{table}

\subsubsection{CL-efficiency significance tests}
\label{multianchor_cl_significance_tests_additional_info}

Figure \ref{fig:Scalability_multianchor_supp} shows the training time steps for each model run for the different main anchor configurations separately.

For CL efficiency analysis, data was combined across memory sizes and anchors for a architecture-level analysis. Anchor configuration did not have a significant effect on any of the CL metric (p<0.05), after accounting for architecture and memory size (2-anchor: F(3,57)=0.11, p=0.954; 3-anchor: F(3,49)=1.20, p=0.318; 4-anchor(3,13): F=0.41 , p=0.751). 

\begin{itemize}
    \item For CL stage 1 (2-anchor setting) training time steps, the two memory sizes within each architecture  (LSTM: Mann-Whitney U=95.00, p=0.221; CTRNN: Mann-Whitney U=148.00, p=0.462), did not show a significant difference.

    \item  For CL-efficiency ratio in CL stage 2 (3-anchor setting),  the two memory sizes within each architecture (LSTM: Mann-Whitney U=143.00, p=0.205; CTRNN: Mann-Whitney U=73.00, p=0.589), did not show a significant difference. 

    \item For CL-efficiency ratio in CL stage 3 (4-anchor setting),  the two memory sizes within each architecture  (LSTM: Mann-Whitney U=6.00, p=0.533; CTRNN: Mann-Whitney U=29.00, p=0.573), did not show a significant difference. 
    
\end{itemize}
Hence, an architecture level analysis of CL-efficiency (or scalability in general) was possible for the multi-anchor categorization task.

\newpage
\section{Appendix - Asynchronous Multi-Oven}
\label{async_multioven_appendix}
\subsection{Baseline Results}
The RL agents that reliably performed the temporal bisection task ($LSTM125$, $LSTM256$, $CTRNN125$, $CTRNN256$) were tested for their asynchronous multi-oven bisection capabilities. The analysis is conducted for 3 types of anchor configurations ($S$-$L$), namely, 4-8, 4-12 and 4-18. The RL agents were trained using CL with first CL stage corresponding to the basic temporal bisection task (1-oven setting), second CL stage corresponding to the first introduction of concurrent asynchronous (or synchronous) bisection task (2-oven setting) and third CL stage with 3 concurrent asynchronous (or synchronous) bisection tasks (3-oven setting). The performance thresholds for 1-oven setting and 2-oven setting was 0.95 and 0.90 respectively. The analysis is conducted using 4 runs from each model with a maximum of 1750k training steps.

\begin{figure}[ht]
    \centering

    \begin{subfigure}[t]{0.8\linewidth}
        \centering
        \includegraphics[width=\linewidth]{figures/multioven_asyn_performance.pdf}
        \caption{Asynchronous Multi-Oven Bisection}
        \label{fig:multioven_async_supp}
    \end{subfigure}

    \begin{subfigure}[t]{0.8\linewidth}
        \centering
        \includegraphics[width=\linewidth]{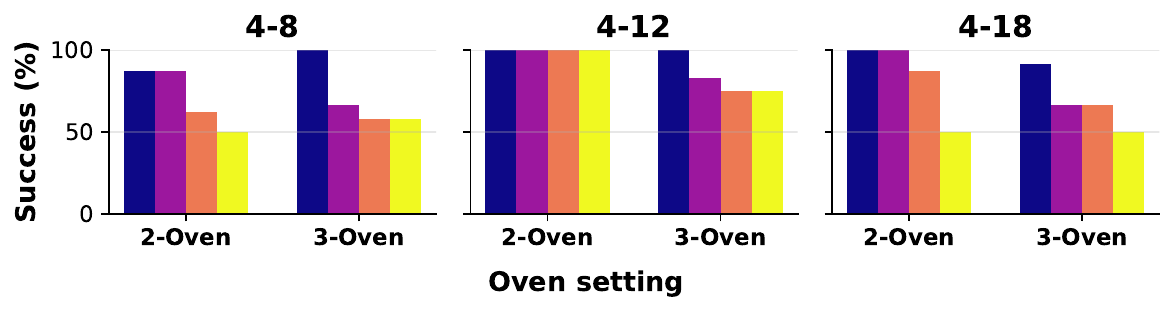}
        \caption{Synchronous Multi-Oven Bisection}
        \label{fig:multioven_sync_supp}
    \end{subfigure}

    \caption{Multi-oven bisection (task performance): The top plot depicts the asynchronous task performance and bottom plot shows the synchronous task performance. The plots show the success rate (successful bisection tasks) across all ovens, probe durations (including OOD durations) and runs, for each model (shown in different colours) and anchor configuration (columns), in 2-oven and 3-oven setting (x-axis). While the asynchronous multi-oven task was in general difficult, performance improved as distance between $S$ and $L$ anchor increased.}
    \label{fig:multioven_task_performance}
    \Description{All the RL agents achieved good performance in the synchronous variant of the task. In the asynchronous variant, performance improved as distance between $S$ and $L$ increased. In the 3-oven setting, models achieved at most a 50\% success rate for the 4–18 anchor configuration (maximum distance between $S$ and $L$), while performance was lower for all other anchor configurations. }
\end{figure}

\begin{figure}[ht]
  \centering
  \includegraphics[width=0.6\linewidth]{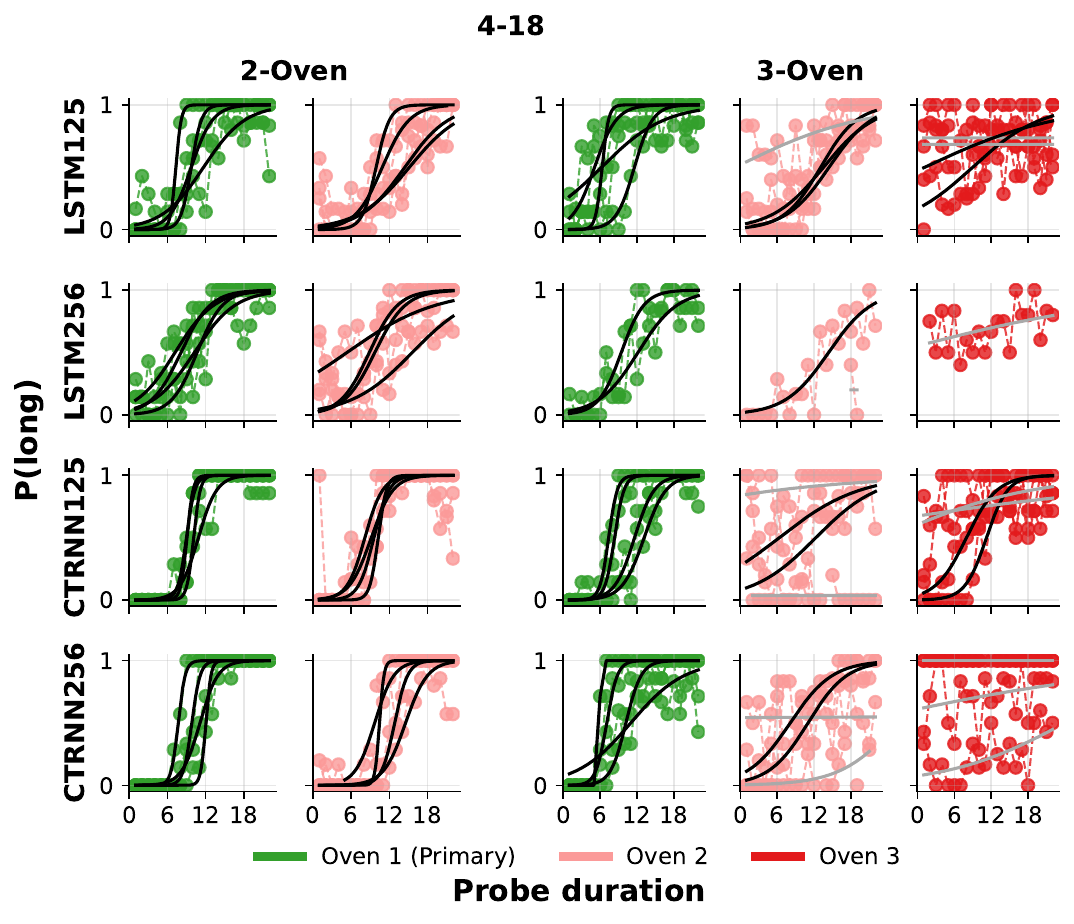}
  \caption{Asynchronous multi-oven bisection (psychometric curves): The plot shows the psychometric curve of each model (rows) for the different ovens in the 2-oven (left) and 3-oven setting (right), for the anchor configuration 4-18 which was the best performing anchor (also the one with the largest distance between the anchors). The primary oven (oven 1) shows the best performance among the different ovens, especially in the 3-oven setting. }
  \label{fig:multioven_asyn_psychometric}
  \Description{The primary oven (oven 1) shows the best performance among the different ovens, especially in the 3-oven setting.}
\end{figure}

The task performance is evaluated using separate psychometric curves for each oven. This enables comparison of the primary oven (initiated by the agent) against other ovens (initiated by the environment). Appendix Figure \ref{fig:multioven_asyn_psychometric} shows the psychometric curves for anchor configuration 4-18 of the asynchronous variant for 2-oven and 3-oven setting. Task performance is quantified using the number of successful bisection tasks across the different ovens. A successful bisection task is determined using the same techniques from temporal bisection (section 7.2). It is important to note that the evaluation also includes OOD probe durations. Appendix Figure \ref{fig:multioven_task_performance} shows the success rate of each model (shown in a different colour) for the asynchronous (top) and synchronous (bottom) variant, for the three anchor configurations (4-8, 4-12, 4-18).

\subsubsection{Asynchronous vs Synchronous timing} 

\begin{figure}[ht]
  \centering
  \includegraphics[width=0.9\linewidth]{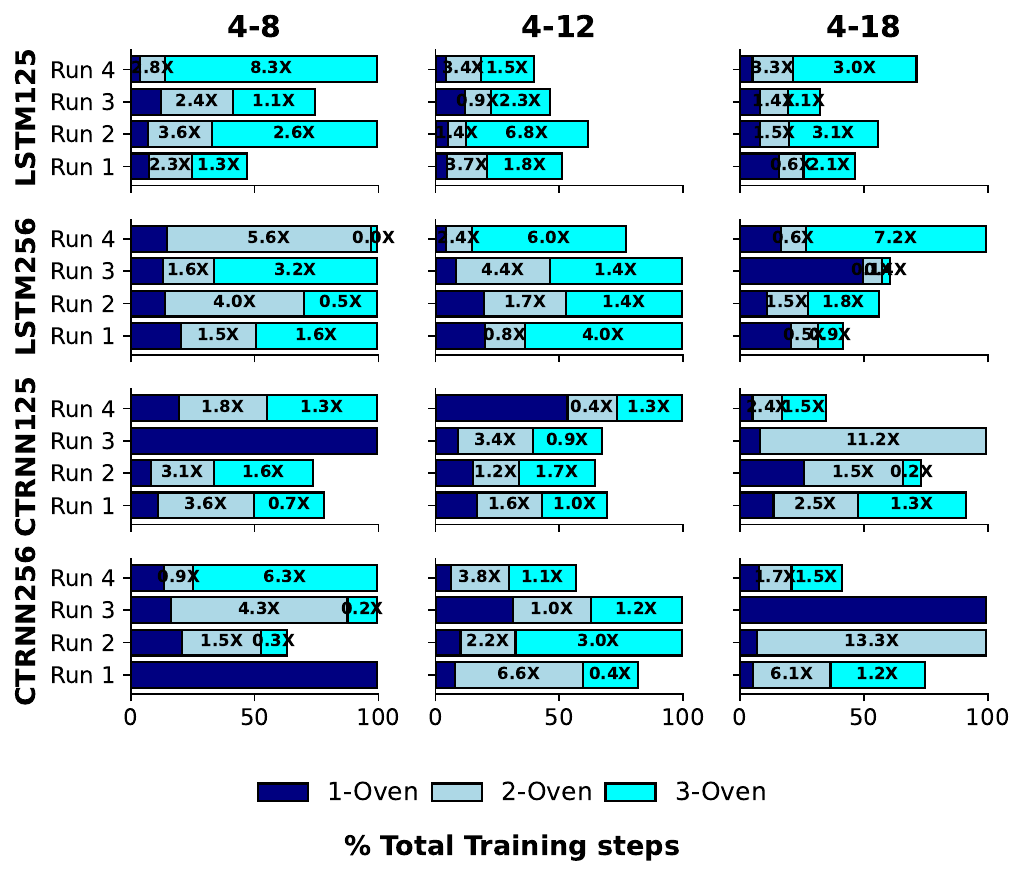}
  \caption{Synchronous multi-oven bisection (scalability): The plots show the training time steps required for each CL stage (n-oven setting) shown in different colours for each model (rows) and anchor configuration (columns) for the synchronous variant. The numbers in the bars depict CL-efficiency ratios. At an architecture level, the CL-efficiency ratio of both architectures was significantly more than 1.}
  \label{fig:scalability_multioven_sync_cl_supp}
  \Description{In the synchronous variant most model runs reached the last CL stage (3-oven).}
\end{figure}

The asynchronous task was more difficult than the synchronous variant in which all ovens started at the same time as the primary oven (oven 1). Under the same training budget of 1750k steps, for the synchronous variant, 56\% of runs across all models converged in the 3-oven setting (last CL stage) before the maximum training budget was reached, while none of the runs from the asynchronous variant converged. For the 2-oven setting (first introduction of concurrent synchronous or asynchronous bisection task), 90\% of runs from synchronous variant converged, compared to only 44\% from the asynchronous variant (see Figures \ref{fig:scalability_multioven_async_cl_supp} and Appendix Figure \ref{fig:scalability_multioven_sync_cl_supp} for training step information of asynchronous and synchronous variant respectively). Additionally, taken across all models and anchor configurations (with the caveat that significant difference due to architecture, memory size and anchor configurations were observed for this metric, see Appendix Additional information section \ref{async_multioven_cl_significance_test_additional_info} for details), the CL-efficiency ratio in the the 2-oven setting of the synchronous variant was 2.4, while for asynchronous variant it was 5.9 for the converged runs. Hence, the RL agents require more training steps to incorporate a second asynchronous oven compared to a second synchronous oven. Additionally, from Figure \ref{fig:multioven_task_performance} we can deduce that the overall 3-oven performance was much better in the synchronous variant than in the asynchronous variant, further providing evidence of a higher rate of non-convergence in the later. 

\begin{figure}[ht]
    \centering

    \begin{subfigure}[t]{0.8\linewidth}
        \centering
        \includegraphics[width=0.6\linewidth]{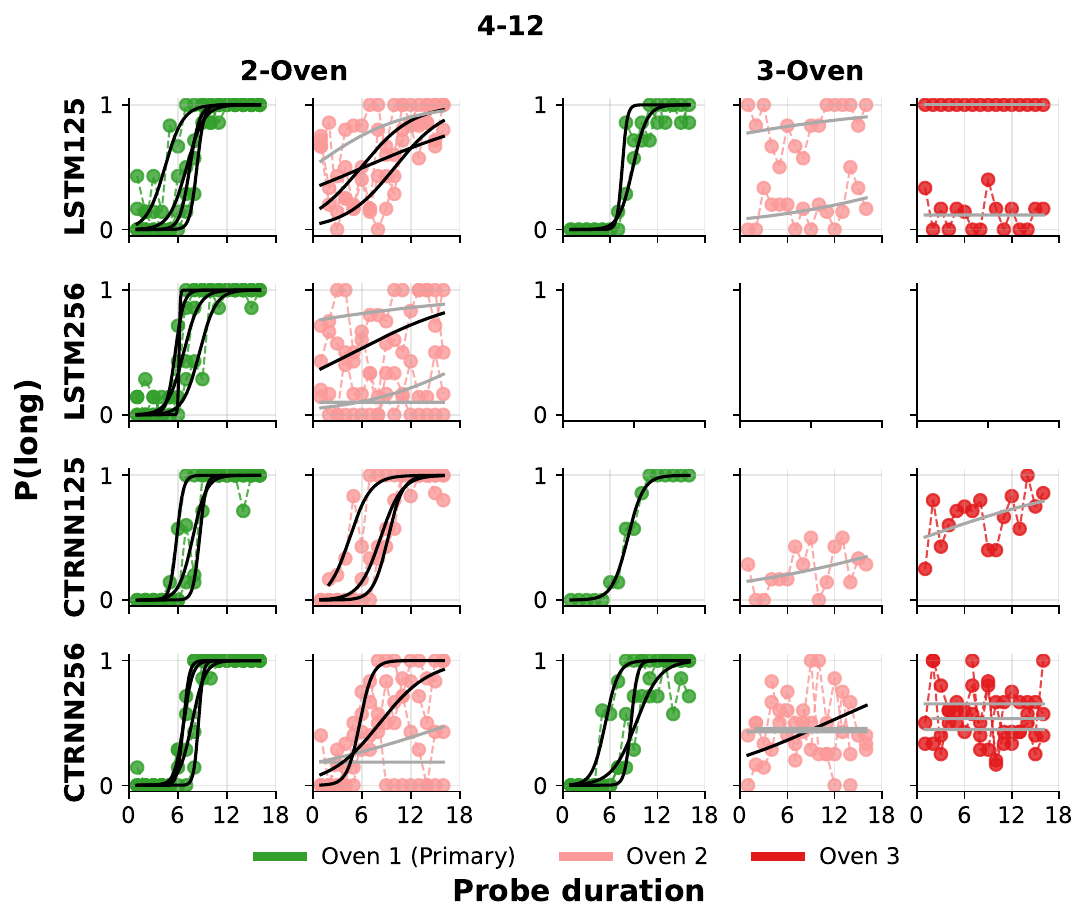}
        \caption{Anchor configuration: 4-12}
        \label{fig:multioven_psychometric_4_12}
    \end{subfigure}

    \begin{subfigure}[t]{0.6\linewidth}
        \centering
        \includegraphics[width=0.8\linewidth]{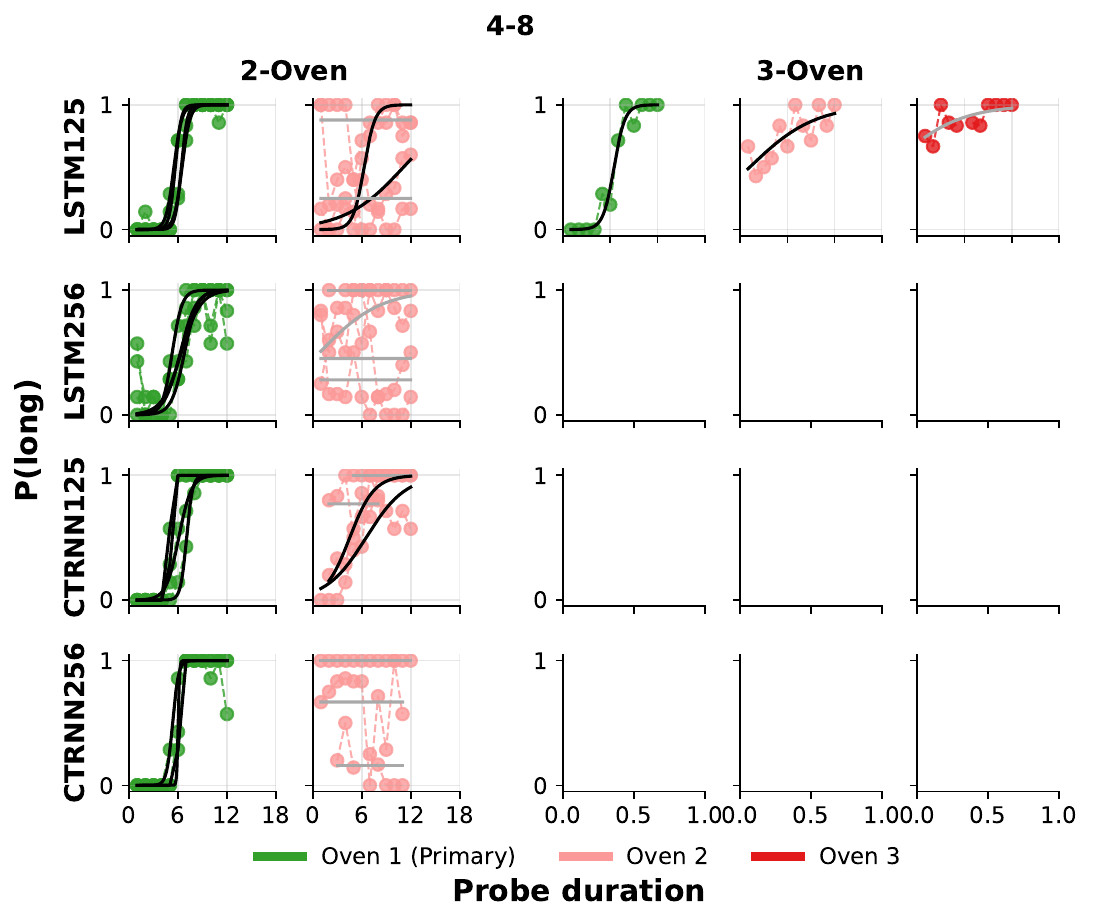}
        \caption{Anchor configuration: 4-8}
        \label{fig:multioven_psychometric_4_8}
    \end{subfigure}

    \caption{Asynchronous Multi-oven bisection (psychometric curve): The plot shows the psychometric curve for the different
ovens for 2-oven and 3-oven setting for anchor 4-12 (top) and 4-8(bottom). All model performance degrades in the 3-oven setting. The primary oven (oven 1) shows the best performance among
the different ovens.}
    \label{fig:multioven_psychometric_supp}
    \Description{The performance in the 3-oven stage for the asynchronous timing task is better in the 4-12 anchor configuration rather than the 4-8 configuration. All model performance degrades in the 3-oven setting. The primary oven (oven 1) shows the best performance among
the different ovens.}
\end{figure}
\subsubsection{Self-triggered vs environment-triggered} The psychometric curves of individual ovens in the asynchronous variant (Figure \ref{fig:multioven_asyn_psychometric} and Appendix Figures \ref{fig:multioven_psychometric_4_12} and \ref{fig:multioven_psychometric_4_8} ) reveal that the primary oven (oven 1), whose start time is controlled by the agent, showed the best performance compared to the other ovens in the 3-oven setting.  This observation combined with the comparison against the synchronous variant indicates that, events triggered by the RL agent's own actions are easier for it to keep track of, than events triggered by the environment. However, there are a few runs across all models that successfully perform the bisection task even for the asynchronous ovens (Oven 2 and 3), especially for the easier anchor configuration (4-18). This shows evidence that RL agents are capable of tracking environment-triggered events, although with limited capacity. Further investigation (for example, through neural analysis) is needed to determine whether the two types of timing events rely on different underlying mechanisms. 


\subsubsection{Task difficulty}
The task performance for the asynchronous variant improves as distance between $L$ and $S$ anchor increases (see Appendix Figure \ref{fig:multioven_task_performance} top plot). For the smallest distance anchor configuration (4-8), only one run of model ($LSTM125$) advances to the 3-oven CL stage (Figure \ref{fig:scalability_multioven_async_cl_supp}). Additionally, for the 2-oven CL stage, all models show task performance below the CL threshold (0.90) (see Appendix Figure \ref{fig:multioven_task_performance} top plot), indicating that they could not generalize to OOD probe durations for the 4–8 anchor configuration. The performance of all models improves for anchor configuration 4-12. For the anchor configuration 4-18 (with the largest distance between the two anchors), all models show good generalization for the 2-oven setting, and at least 2 runs from each model advance to the 3-oven CL stage, within the same training budget (1750k training steps). Moreover, the task performance across models for the 4-18 anchor configurations is 52\% which is quite close to the corresponding performance in synchronous variant (56\%) (last column in Appendix Figure \ref{fig:multioven_task_performance}). This shows that RL agents are able to perform asynchronous bisection tasks, and that the distance between the two anchors plays a role in task difficulty. 

\subsubsection{Model-level performance}
$LSTM256$ exhibited very poor performance in the asynchronous multi-oven bisection task. For anchor configuration 4-18, where all other models exhibited task performance of at least 50\% success rate, $LSTM256$ had a success rate of only 25\% (see Appendix Figure \ref{fig:multioven_task_performance} top plot). In terms of capacity as well, all 4 runs from other models successfully converged for the 2-oven setting while for $LSTM256$ only 2 runs converged. Similarly for anchor configuration 4-12, where all models had at least one run converging in the 2-oven setting, none of the the runs of $LSTM256$ converged (see Figure \ref{fig:scalability_multioven_async_cl_supp}). The psychometric curves of this model from Appendix Figure \ref{fig:multioven_asyn_psychometric} also reveal that the models performance was not as good as other models even for the primary oven in both 2-oven and 3-oven setting. $LSTM256$ is the most powerful model (highest number of parameters) of the 4 with respect to number of parameters. Further investigation using more hyperparameter tuning especially including higher memory sizes in both architectures may reveal some useful insights and help to identify the cause of this observation.

\subsubsection{CL efficiency} An architecture-level analysis of the CL-efficiency ratio is conducted for the categorical variant.  This analysis however, comes with a caveat that in some cases either memory sizes and anchor configurations within an architecture have a significant effect on some CL-efficiency ratios (See Appendix Additional information section \ref{async_multioven_cl_significance_test_additional_info} for details). 

\begin{itemize}

    \item \textbf{CL stage 1 (1-oven setting):}
    In the asynchronous variant, CTRNN training time in CL stage 1 was significantly greater than LSTM training time (CTRNN: N=24, M=15.37, Median=12.13, SD=10.40; LSTM: N=24, M=11.87, Median=7.95, SD=13.86) (Mann-Whitney U=415.00, p=0.004). However, this comparison should be interpreted with some caution, since LSTM's two memory sizes differed significantly from each other (Mann-Whitney U=6.00, p<0.001), while CTRNN's did not (Mann-Whitney U=84.00, p=0.507). The small sample sizes per model (n=12) limits further disaggregation.

    \item \textbf{CL stage 2 (2-oven setting):}
    For the second CL stage in asynchronous variant, anchor 4-8 was excluded from analysis since it had only 1 data point from $LSTM125$ (see Figure \ref{fig:scalability_multioven_async_cl_supp}). Descriptively, CTRNN (M=4.56, SD=3.24, Median=3.86, N=12) showed lower mean CL-efficiency ratio for stage 2 than LSTM (M=6.51, SD=2.08, Median=6.37, N=8). But LSTM was not significantly greater than CTRNN (Mann-Whitney U=21.00, p=0.984).  Both architectures exhibited CL-efficiency ratio for CL stage 2 that was significantly greater than 1 (CTRNN: N=12, Wilcoxon, stat=78.00, p<0.001; LSTM: N=8, t=7.50, p<0.001). Hence, for the first introduction of an asynchronous bisection task, there was no significant difference between the two architectures and both needed significantly more than 1x the training steps required in the first stage. Interestingly in synchronous variant as well the CL-efficiency ratio of both architectures was significantly more than 1 for the 2-oven setting (CTRNN: t=4.16, p<0.001; LSTM: t=4.06, p<0.001).

\end{itemize}

Since CL-efficiency ratio was significantly more than 1, no evidence of an efficient temporal representation was found. However, although the CTRNN models needed significantly more training steps than LSTM in the first CL stage, the task performance of these models was comparable to LSTM models. This shows that the CTRNN models are able to catch up with the more powerful LSTM models in the later CL stage, which could hint towards the emergence of a flexible timing strategy not caught by the simple definition of CL efficiency established in this study. As mentioned before a neural analysis could help better understand timing mechanism of the different models and architectures for the different cases (asynchronous, synchronous, self-triggered and environment-trigged) brought forward by this task.  

\subsection{Additional Information}

\subsubsection{CL-efficiency significance tests}
\label{async_multioven_cl_significance_test_additional_info}
For the first CL stage in the asynchronous variant, the full model (performance ~ architecture × memory-size + env-config) showed no significant effect of anchor configuration after accounting for architecture and memory size (F(2,42)=0.412, p=0.665). Comparing memory sizes within each architecture using Mann–Whitney U tests, LSTM showed a significant difference between memory sizes (U=6.0,p<0.001), while CTRNN did not (U=84.0,p=0.507). Given this, CTRNN and LSTM data were compared directly as an approximate architecture-level comparison, treating memory sizes as pooled within each architecture. This showed that CTRNN models had significantly greater CL stage 1 training steps than LSTM (Mann-Whitney U=415.00, p=0.004). However, this comparison should be interpreted with some caution, since LSTM's two memory sizes differed significantly from each other, while CTRNN's did not, meaning the pooled LSTM distribution may not fully represent a single homogeneous group, and the modest sample sizes per memory size (n=12) limits further disaggregation.

For the second CL stage in asynchronous variant, the anchor 4-8 was removed since it had only 1 sample. The overall model (performance ~ C(architecture) × C(memory-size) + C(anchor)) found no significant effect of architecture (F(1,15)=2.74, p=0.119), memory size (F(1,15)=0.004, p=0.949), anchor configuration (F(1,15)=2.71, p=0.121), or the architecture × memory size interaction (F(1,15)=0.63, p=0.440). Within-architecture analyses similarly found no significant effects. For LSTM, neither memory size (F(1,5)=1.67, p=0.253) nor anchor configuration (F(1,5)=0.12, p=0.745) had a significant effect. Same for CTRNN, neither memory size (F(1,9)=0.09, p=0.772) nor anchor configuration (F(1,9)=4.64, p=0.060) reached significance. Descriptively, CTRNN (M=4.56, SD=3.24, Median=3.86, n=12) showed lower mean than LSTM (M=6.51, SD=2.08, Median=6.37, n=8), and anchor configuration 4-12 (M=7.01, SD=3.54, n=6) showed a higher mean than 4-18 (M=4.62, SD=2.43, n=14), though neither difference reached statistical significance. 

Similarly for the second CL stage in synchronous variant, overall model (performance ~ C(architecture) × C(memory-size) + C(anchor)), no significant effect of architecture was found (F(1,37)=0.790, p=0.380). Similarly, no significant effects of memory size (F(1,37)=0.379, p=0.542) or anchor configuration (F(2,37)=1.165, p=0.323) were observed. The interaction between architecture and memory size was also not significant (F(1,37)=1.384, p=0.247). Within the LSTM architecture, no significant effect of memory size was found after accounting for anchor configuration (F(1,20)=0.147, p=0.705). However, a significant effect of anchor configuration was observed (F(2,20)=3.933, p=0.036). Within the CTRNN architecture, memory size likewise had no significant effect after accounting for anchor configuration (F(1,15)=1.503, p=0.239), and no significant effect of anchor configuration was found (F(2,15)=0.102, p=0.904). The CL-efficient ratio in CL stage 2 for CTRNN (N=19, M=2.62, SD=1.70) was not significantly greater than LSTM (N=24, M=2.17, SD=1.42) ( t=0.94, p=0.177). 

\newpage

\section{Appendix - Miscellaneous}
\label{misc_appendix}

\subsection{Baseline models}
\label{baseline_models_appendix}
The recurrent agents use a CNN-LSTM-MLP or CNN-CTRNN-MLP actor–critic architecture trained with Recurrent PPO from stable-baselines3. A custom CNN processes the six-channel observation (grid layout, agent position, oven state, carrying onion, carrying soup, one extra channel was added for compatibility to another task not included in this study because the idea was to keep the observation space same for all tasks) and produces a 64-dimensional feature representation. This representation is passed to a single-layer LSTM or CTRNN with hidden size H (8/125/256). The LSTM or CTRNN layer is shared between the actor and critic, followed by separate 64-unit MLP (Multilayer Perceptron) heads for policy and value estimation. The CNN agents have the same configuration except the recurrent layer is removed and it is trained with PPO from stable-baselines3. 

The models were trained for 1750k training steps for all core task extensions, 960k training steps for the bisection task and 900k training steps for the FI task. Model checkpoints were saved every 50,000 training steps. The models used 1000 steps per rollout and 20 optimisation epochs per update. The discount factor was 0.99 and entropy coefficient was 0. All other parameters used the default values from Stable-Baselines3 (2.9.0) and Stable-Baselines3 Contrib (2.9.0).

All experiments were conducted in a Python 3.10 environment using PyTorch 2.14.0, Gymnasium 1.3.0, Stable-Baselines3 2.9.0, and Stable-Baselines3 Contrib 2.9.0. Training jobs were run in a Docker environment based on the pytorch/pytorch:2.0.1-cuda11.7-cudnn8-runtime image, using Ubuntu 20.04 LTS (Focal Fossa). The jobs were allocated 2 CPUs and 16 GB of system memory.

\subsection{FI timing}
\label{fi_timing_appendix}
Table \ref{tab:FI_FOC_significance_supp} shows the results of significance test comparing FOC with the corresponding TD for each model. 

\begin{table}
\caption{FI timing: Significance test comparing FOC with the corresponding TD for each model}
\label{tab:FI_FOC_significance_supp}
\begin{tabular}{lrrrrrr}

\toprule
TD & Model & N & M & SD & t & p \\
\midrule
4 & LSTM8 & 36 & 3.39 & 0.87 & -4.21 & 0.000 \\
4 & LSTM125 & 48 & 3.98 & 0.53 & -0.27 & 0.785 \\
4 & LSTM256 & 48 & 3.96 & 0.68 & -0.42 & 0.674 \\
4 & CTRNN8 & 24 & 0.00 & 0.00 & -inf & 0.000 \\
4 & CTRNN125 & 42 & 2.52 & 1.44 & -6.66 & 0.000 \\
4 & CTRNN256 & 36 & 1.97 & 1.63 & -7.46 & 0.000 \\
4 & CNN & 48 & 0.06 & 0.43 & -63.00 & 0.000 \\
6 & LSTM8 & 24 & 3.92 & 1.91 & -5.35 & 0.000 \\
6 & LSTM125 & 48 & 5.62 & 1.23 & -2.11 & 0.040 \\
6 & LSTM256 & 48 & 5.90 & 0.72 & -1.00 & 0.322 \\
6 & CTRNN8 & 24 & 0.00 & 0.00 & -inf & 0.000 \\
6 & CTRNN125 & 24 & 3.71 & 1.88 & -5.98 & 0.000 \\
6 & CTRNN256 & 48 & 4.56 & 1.07 & -9.31 & 0.000 \\
6 & CNN & 48 & 0.06 & 0.43 & -95.00 & 0.000 \\
9 & LSTM8 & 17 & 5.59 & 3.86 & -3.65 & 0.002 \\
9 & LSTM125 & 48 & 9.02 & 0.96 & 0.15 & 0.881 \\
9 & LSTM256 & 48 & 8.67 & 0.97 & -2.37 & 0.022 \\
9 & CTRNN8 & 12 & 7.58 & 1.08 & -4.53 & 0.001 \\
9 & CTRNN125 & 36 & 2.86 & 3.21 & -11.48 & 0.000 \\
9 & CTRNN256 & 36 & 6.78 & 2.15 & -6.19 & 0.000 \\
9 & CNN & 48 & 0.81 & 1.35 & -42.10 & 0.000 \\
12 & LSTM8 & 18 & 12.50 & 2.73 & 0.78 & 0.447 \\
12 & LSTM125 & 48 & 11.46 & 2.13 & -1.76 & 0.085 \\
12 & LSTM256 & 48 & 11.96 & 1.38 & -0.21 & 0.836 \\
12 & CTRNN8 & 24 & 6.17 & 6.39 & -4.47 & 0.000 \\
12 & CTRNN125 & 36 & 7.14 & 3.61 & -8.08 & 0.000 \\
12 & CTRNN256 & 36 & 8.83 & 3.31 & -5.74 & 0.000 \\
12 & CNN & 48 & 0.00 & 0.00 & -inf & 0.000 \\
18 & LSTM8 & 0 & nan & nan & nan & nan \\
18 & LSTM125 & 48 & 18.19 & 2.43 & 0.53 & 0.595 \\
18 & LSTM256 & 48 & 17.12 & 1.57 & -3.87 & 0.000 \\
18 & CTRNN8 & 9 & 5.44 & 2.13 & -17.70 & 0.000 \\
18 & CTRNN125 & 32 & 13.03 & 3.70 & -7.59 & 0.000 \\
18 & CTRNN256 & 36 & 12.53 & 2.85 & -11.51 & 0.000 \\
18 & CNN & 48 & 0.75 & 1.31 & -91.04 & 0.000 \\
24 & LSTM8 & 13 & 10.85 & 4.49 & -10.57 & 0.000 \\
24 & LSTM125 & 48 & 22.75 & 2.72 & -3.18 & 0.003 \\
24 & LSTM256 & 48 & 23.73 & 2.09 & -0.90 & 0.374 \\
24 & CTRNN8 & 0 & nan & nan & nan & nan \\
24 & CTRNN125 & 36 & 18.92 & 4.25 & -7.18 & 0.000 \\
24 & CTRNN256 & 36 & 16.92 & 4.09 & -10.40 & 0.000 \\
24 & CNN & 48 & 0.81 & 1.35 & -119.24 & 0.000 \\
\bottomrule

\end{tabular}
\end{table}

\subsection{Bisection task}
\label{bisection_appendix}
\begin{figure}[ht]
  \centering
  \includegraphics[width=\linewidth]{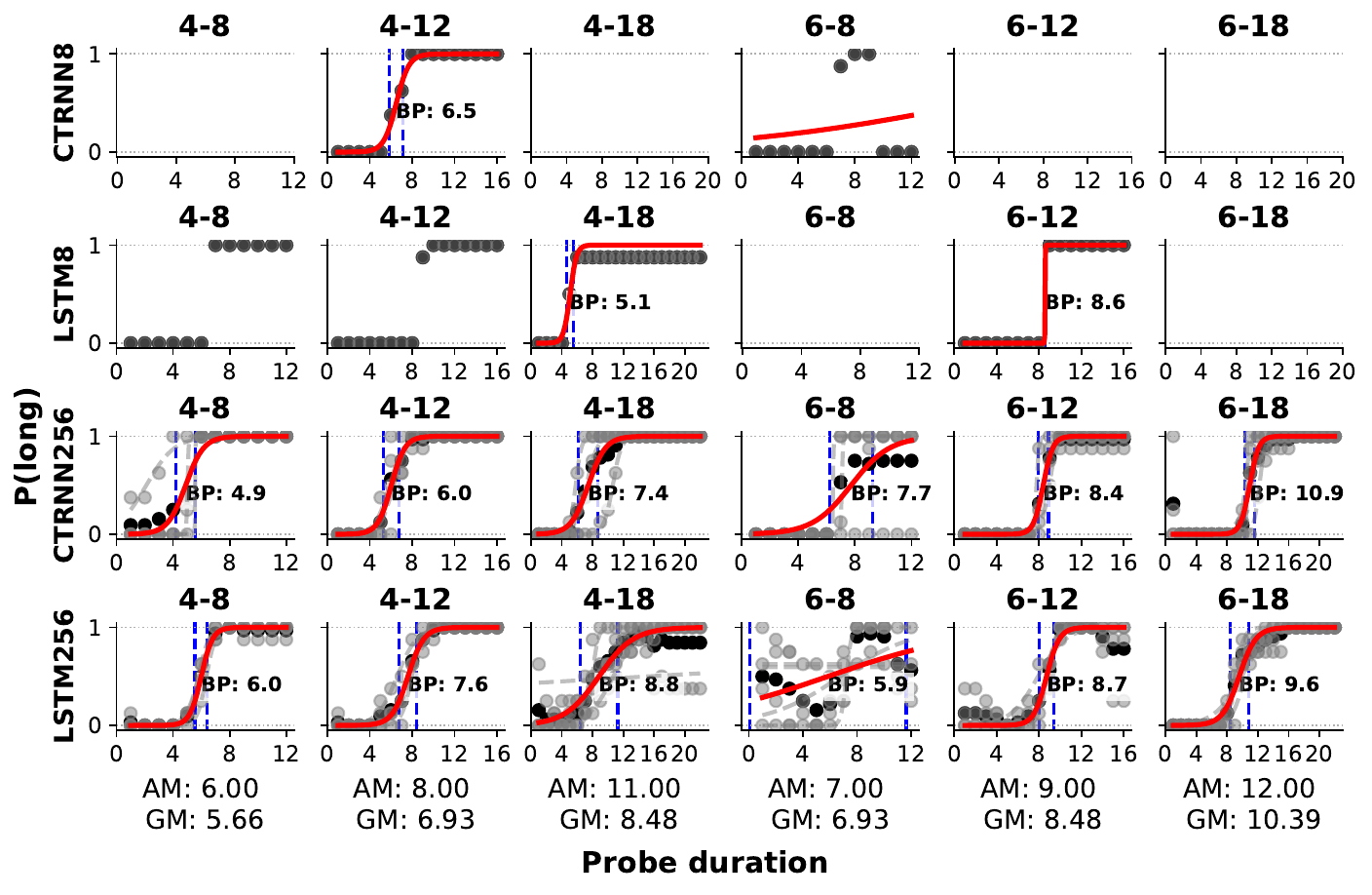}
  \caption{Psychometric curves of the models (rows) for the different anchors (columns). Y-axis shows percentage of long - P(long) and x-axis shows probe (test) durations along the the arithmetic mean (AM) and geometric mean (GM) of each anchor. The grey dots represent p(long) of each run (with 11 seeds each). The black dots is the average P(long) across all runs. The red line shows the sigmoid fit on the average P(long). The bisection points (BP) are shown in the respective plots.  }
  \label{fig:psychometric_curves_supp}
  \Description{The smaller memory sizes in the recurrent models do not perform as good as the corresponding high memory models}
\end{figure}

Figure \ref{fig:psychometric_curves_supp} shows the psychometric curves for CTRNN8, LSTM8, CTRNN256 and LSTM256. 

\subsubsection{BP and Weber ratio comparison}
There was no strong positive correlation between the distance of BP from AM and GM and the spread for any of the artificial agent models as found in the human data \cite{kopec2010human} (First column in Figure \ref{fig:bp_plots}). LSTM showed a weak negative correlation (r=-0.23, p=0.027) and no evidence for correlation was found for CTRNN (r=-0.08, p>0.05). Hence, unlike human data where BP transitions from being sub-geometric to sub-arithmetic with increasing spread (strong positive correlation with spread r=0.58, p<0.01 \cite{kopec2010human}), no evidence of a comparable relationship was found in the RL agents data.

BP when normalized by AM shows a negative correlation to spread (Second column in Figure \ref{fig:bp_plots}). For LSTM, at smaller spreads, BP was approximately equal to AM, whereas it decreased substantially as a proportion of AM with increasing spread.  For CTRNN, while BP is sub-arithmatic for smaller spread, it decreases substantially as spread increases. According to human data in \cite{kopec2010human}, BP was found to be sub-arithmetic for smaller spreads and it does not decrease much as spread increases, it reached approximately 90\% of AM for larger spreads. Hence, there was no similarity between this characteristic of BP between humans and RL agents. 

There was no evidence of BP normalized by GM going from sub-geometric to being supra-geometric as spread increased. (third column in Appendix Figure \ref{fig:bp_plots}). Although both architectures showed positive correlation (LSTM: r=0.31, p<0.05, CTRNN: r=0.25, p<0.05) it was not as strong as that reported in human studies.(r=0.88, p<0.01; \cite{kopec2010human}). Hence BP appears to track GM instead of AM in RL agents' data. 

BP behaviour with respect to GM for RL agents can be considered analogous to BP behaviour with respect to AM in human data. In human data BP tracks AM and remains at or below 90\% of AM across all spreads while for artificial agents BP tracks GM and remains at or below 1.2 times GM for all spreads. 

Nevertheless, the significant negative correlation between $\frac{BP}{AM}$ and the spread in the artificial agents suggests that the agents' responses are systematically influenced by the spread. Further investigation of the neural activity of the models may help uncover the mechanisms underlying this behaviour.

The weber ratio from human data shows that weber ratio increases with the spread (Figure \ref{fig:wf_plots}). Which is indicative of decreasing discrimination ability as spread increases. This is similar to the behaviour exhibited in artificial agents. However, the slope is steeper for RL agents (LSTM:r=0.67, p<0.05, CTRNN: r=0.56, p<0.05) in comparison to humans (r=0.32, p=0.04) which means that the discrimination ability decreases more with spread as compared to humans.



\newpage

\end{document}